\documentclass[10pt,twocolumn,letterpaper]{article}

\usepackage{wacv}              % To produce the CAMERA-READY version
\usepackage{graphicx}
\usepackage{amsmath}
\usepackage{amssymb}
\usepackage{booktabs}
\usepackage{enumitem}
\usepackage{multirow}
\usepackage{algorithm}  
\usepackage{algpseudocode}
\usepackage{mathabx}
\newlength{\resLen}            % <--- define it once
\usepackage{pifont}
\newcommand{\cmark}{\ding{51}}%
\newcommand{\xmark}{\ding{55}}%
\usepackage{xcolor}
\usepackage{placeins} % for \FloatBarrier
\usepackage{caption}
\usepackage[pagebackref,breaklinks,colorlinks]{hyperref}

\usepackage[capitalize]{cleveref}
\crefname{section}{Sec.}{Secs.}
\Crefname{section}{Section}{Sections}
\Crefname{table}{Table}{Tables}
\crefname{table}{Tab.}{Tabs.}

\def\wacvPaperID{2993} % *** Enter the WACV Paper ID here
\def\confName{WACV}
\def\confYear{2026}

\begin{document}

\title{Neural Geometry Image-Based Representations with Optimal Transport (OT)}

\author{Xiang Gao*\\
Futurewei Technologies\\
Stony Brook University\\
{\tt\small gao2@cs.stonybrook.edu}
\and
Yuanpeng Liu*\\
Futurewei Technologies\\
Stony Brook University\\
{\tt\small yuanpliu@cs.stonybrook.edu}
\and
Xinmu Wang\\
Futurewei Technologies\\
Stony Brook University\\
{\tt\small xinmuwang@cs.stonybrook.edu}
\and
Jiazhi Li*\\
Futurewei Technologies\\
University of Southern California\\
{\tt\small jiazhil@usc.edu}
\and
Minghao Guo\\
Massachusetts Institute of Technology\\
{\tt\small guomh2014@gmail.com}
\and
Yu Guo\\
Futurewei Technologies\\
George Mason University\\
{\tt\small tflsguoyu@gmail.com}
\and
Xiyun Song\\
Futurewei Technologies\\
{\tt\small xsong@futurewei.com}
\and
Heather Yu\\
Futurewei Technologies\\
{\tt\small hyu@futurewei.com}
\and
Zhiqiang Lao\\
Futurewei Technologies\\
{\tt\small zlao@futurewei.com}
\and
Xianfeng David Gu\\
Stony Brook University\\
{\tt\small gu@cs.stonybrook.edu}
}

\twocolumn[{%
\renewcommand\twocolumn[1][]{#1}%
\maketitle

% *= equal contribution
\begin{center}
  % --- your teaser content (no figure/figure*) ---
  
    \centering
    \addtolength{\tabcolsep}{-5pt}

    \begin{tabular}{c@{\hskip -5pt}c@{\hskip 12pt}c@{\hskip 12pt}c@{\hskip 12pt}c}
        % ---------- Ground Truth + LoD 0–3 ----------
        \multirow{2}{*}{
            \begin{minipage}{0.15\textwidth}
                \centering
                \includegraphics[trim={300 0 100 0},clip,width=\linewidth]{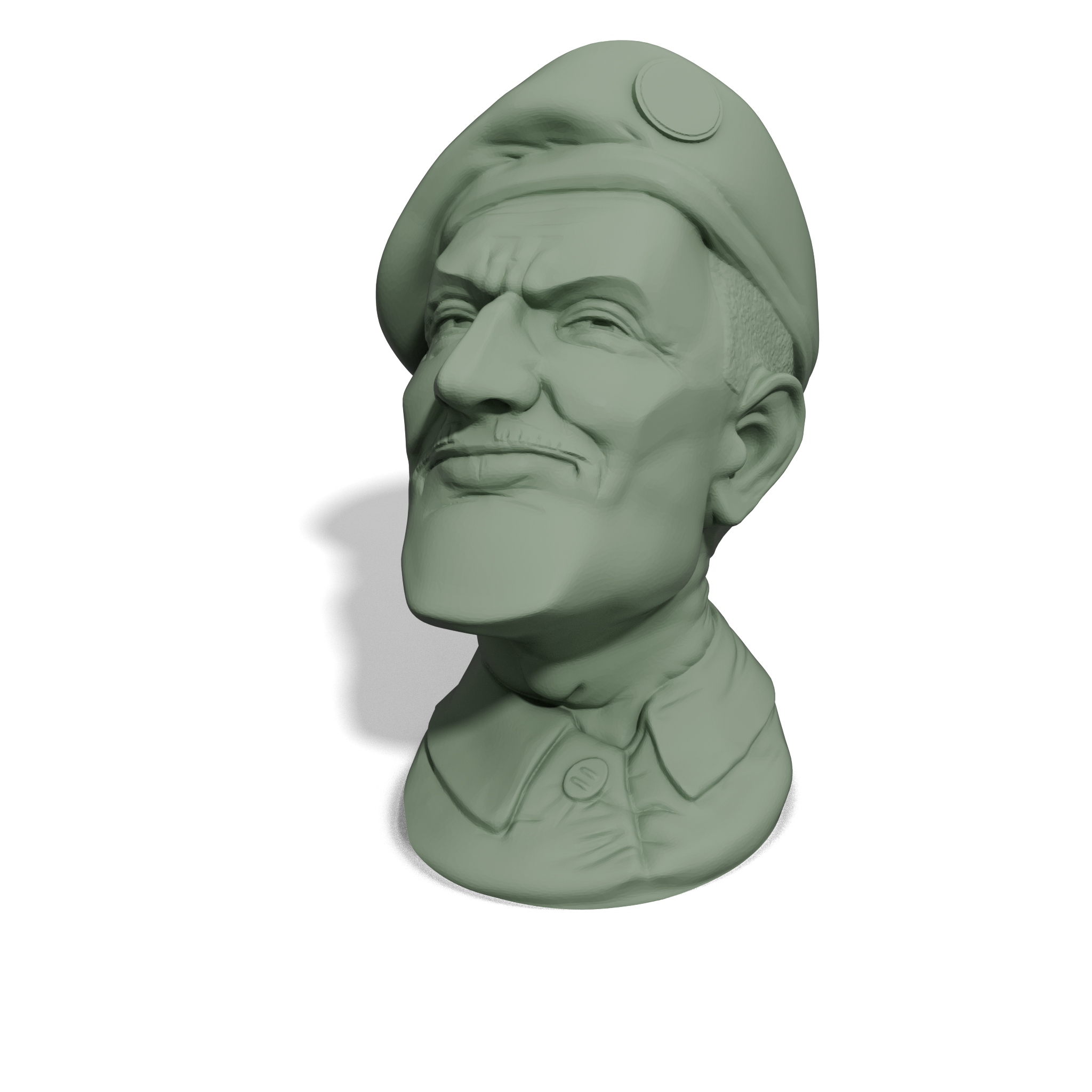} \\[-15pt]
                \small CR / CD ($\times 10^{-4}$) / HD ($\times 10^{-2}$) \\
                \normalsize Ground Truth
            \end{minipage}
        }
        &
        \begin{minipage}{0.19\textwidth}
            \centering
            \begin{tabular}{c@{\hskip -10pt}c}
                \begin{minipage}{0.65\textwidth}
                    \includegraphics[trim={300 0 250 0},clip,width=\linewidth]{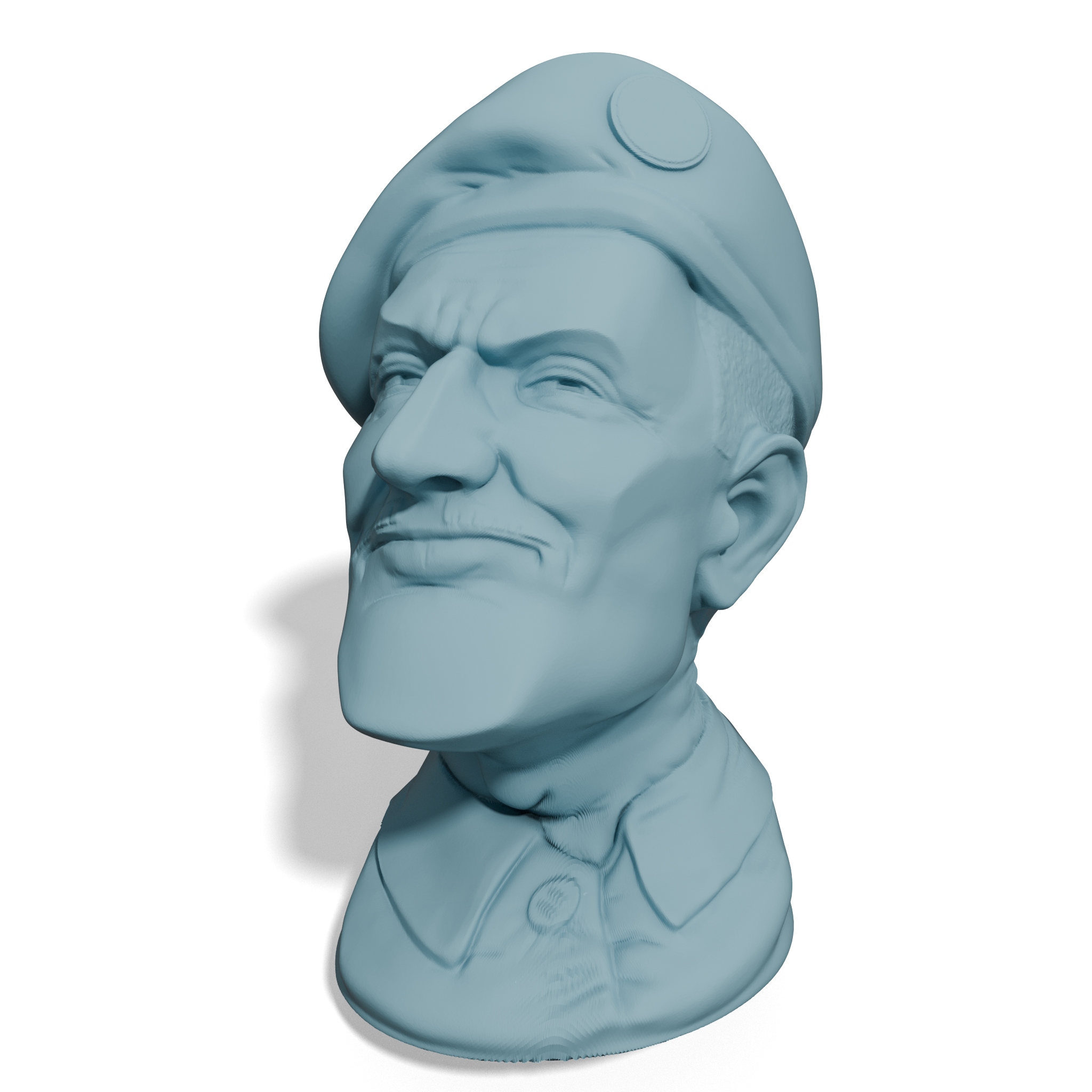}
                \end{minipage}
                \begin{minipage}{0.3\textwidth}
                    \includegraphics[width=\linewidth]{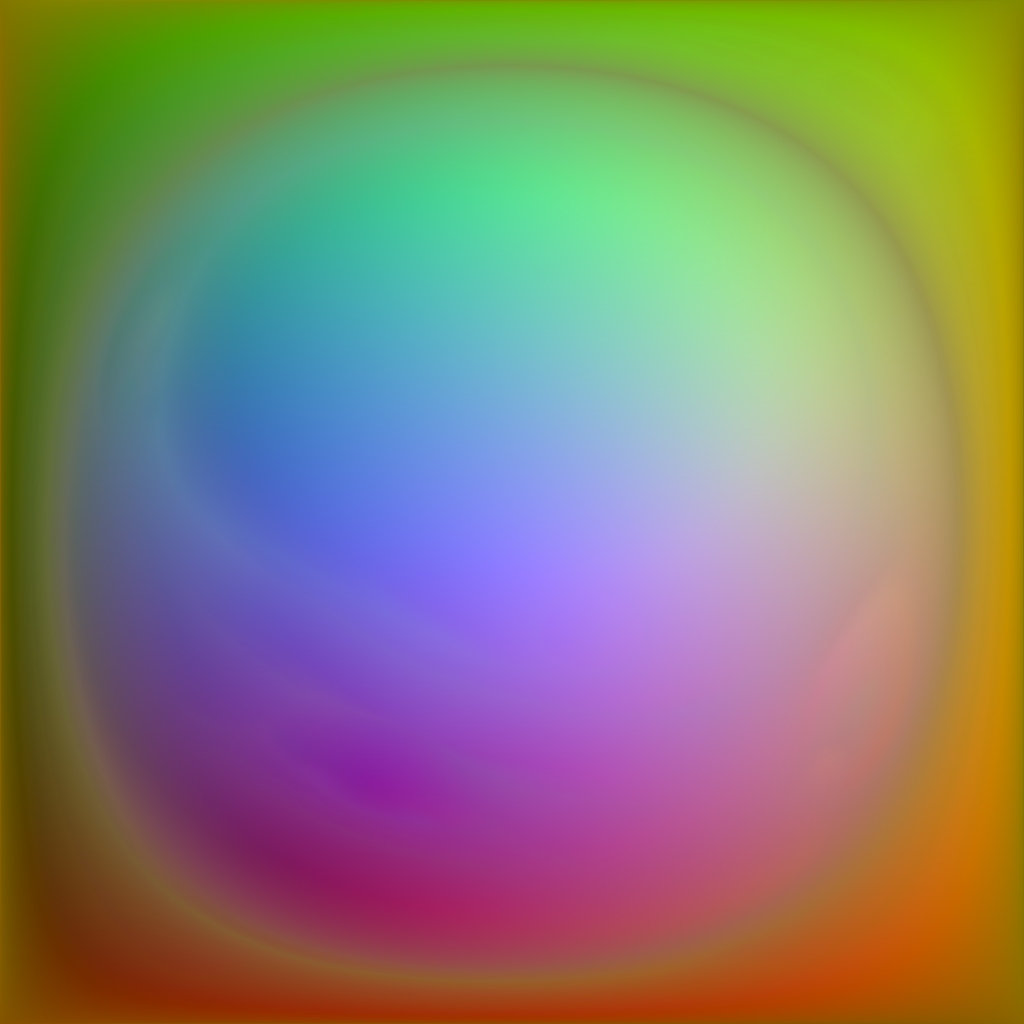}\\
                    \includegraphics[width=\linewidth]{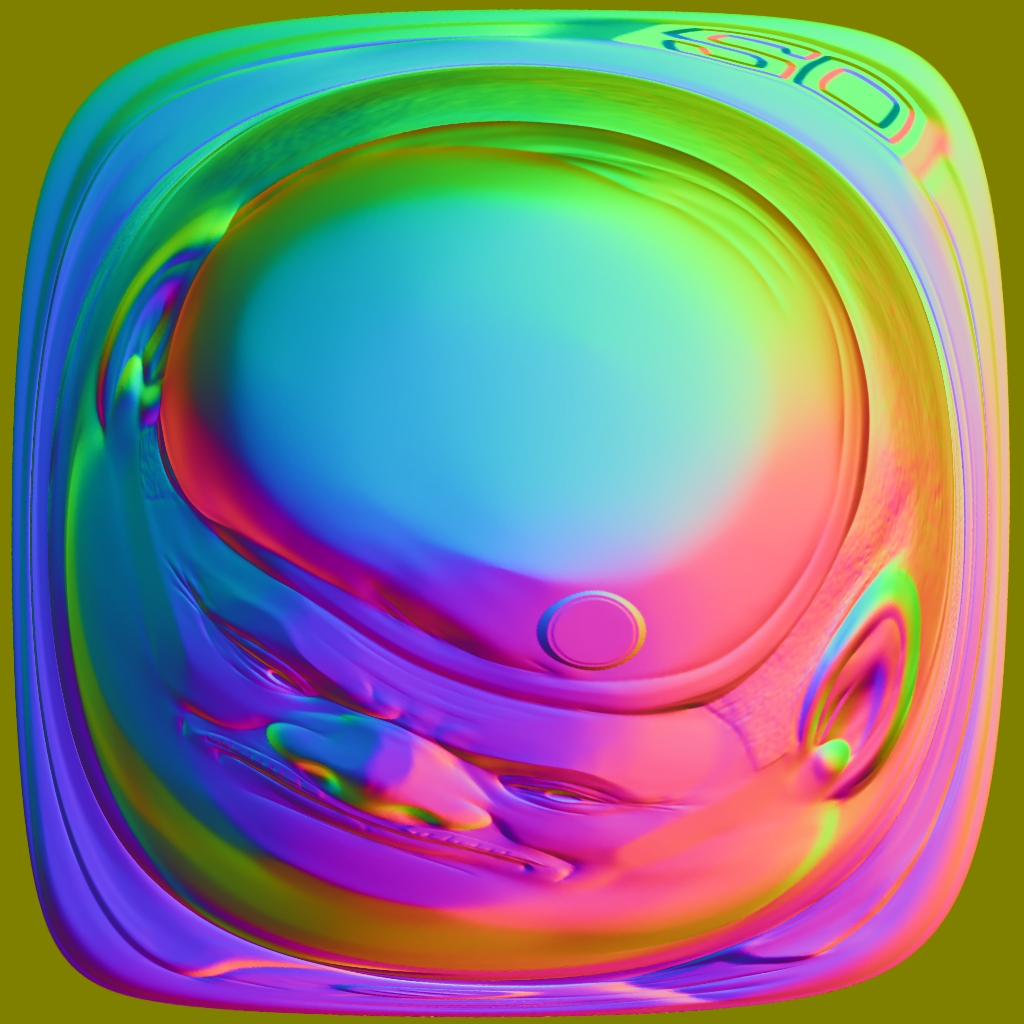}
                \end{minipage}
            \end{tabular}\\[2pt]
            1 / 1.3432 / 2.7919 \\ 
            Level 0 ($1024 \times 1024$)
        \end{minipage}
        &
        \begin{minipage}{0.19\textwidth}
            \centering
            \begin{tabular}{c@{\hskip -10pt}c}
                \begin{minipage}{0.65\textwidth}
                    \includegraphics[trim={300 0 250 0},clip,width=\linewidth]{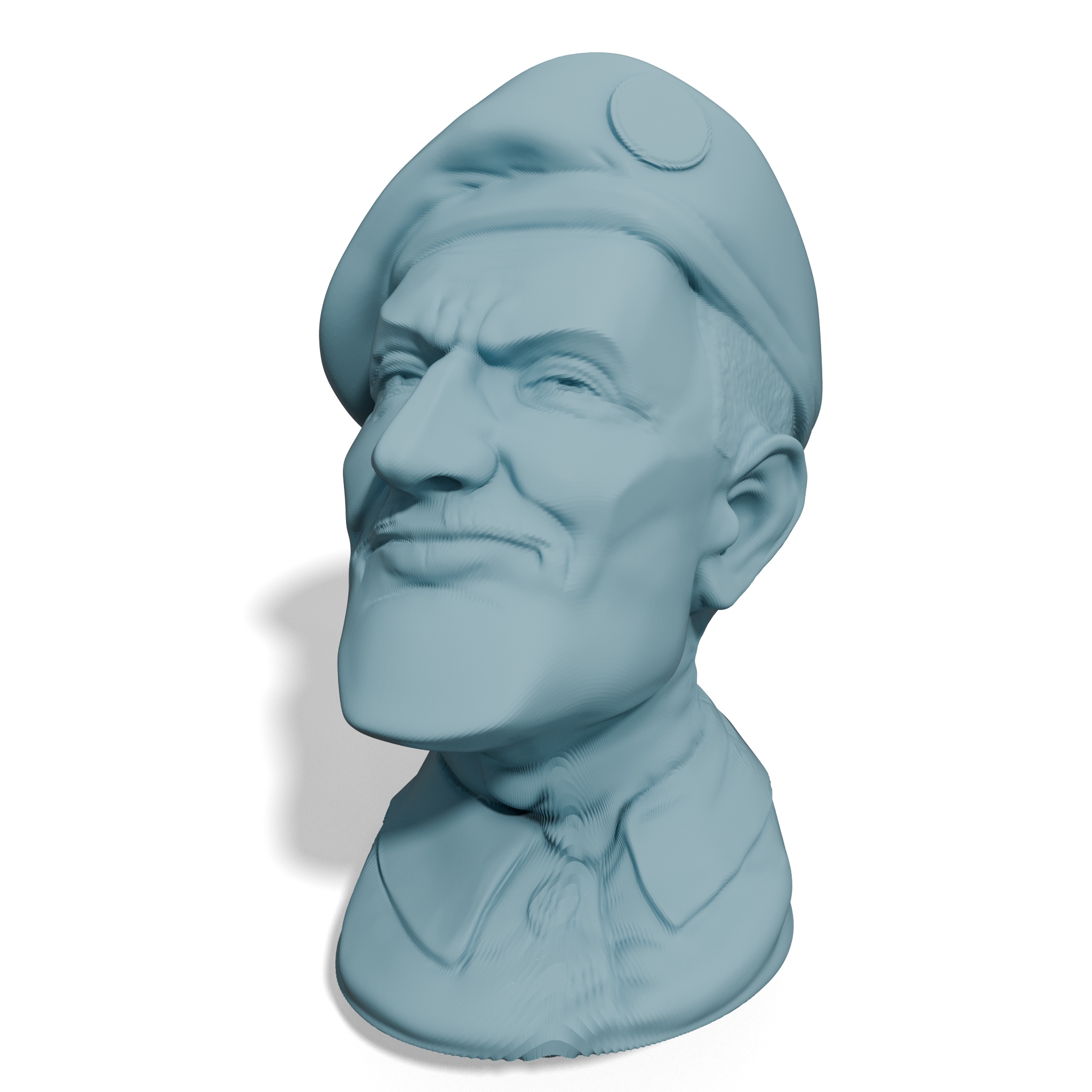}
                \end{minipage}
                \begin{minipage}{0.3\textwidth}
                    \includegraphics[width=\linewidth]{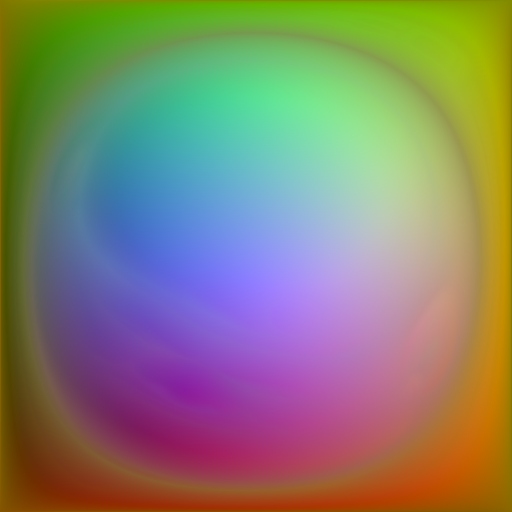}\\
                    \includegraphics[width=\linewidth]{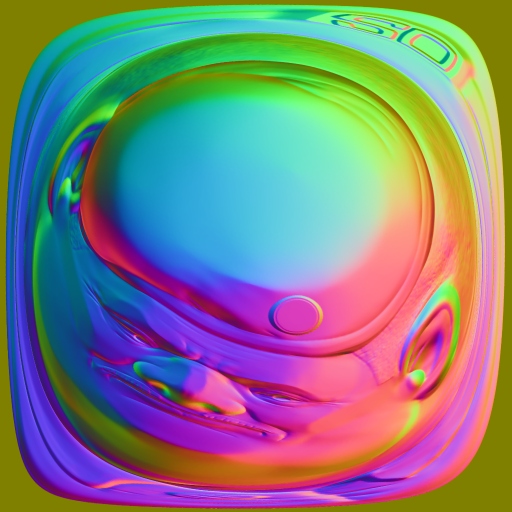}
                \end{minipage}
            \end{tabular}\\[2pt]
            4 / 1.3461 / 2.8221 \\
            Level 1 ($512 \times 512$)
        \end{minipage}
        &
        \begin{minipage}{0.19\textwidth}
            \centering
            \begin{tabular}{c@{\hskip -10pt}c}
                \begin{minipage}{0.65\textwidth}
                    \includegraphics[trim={300 0 250 0},clip,width=\linewidth]{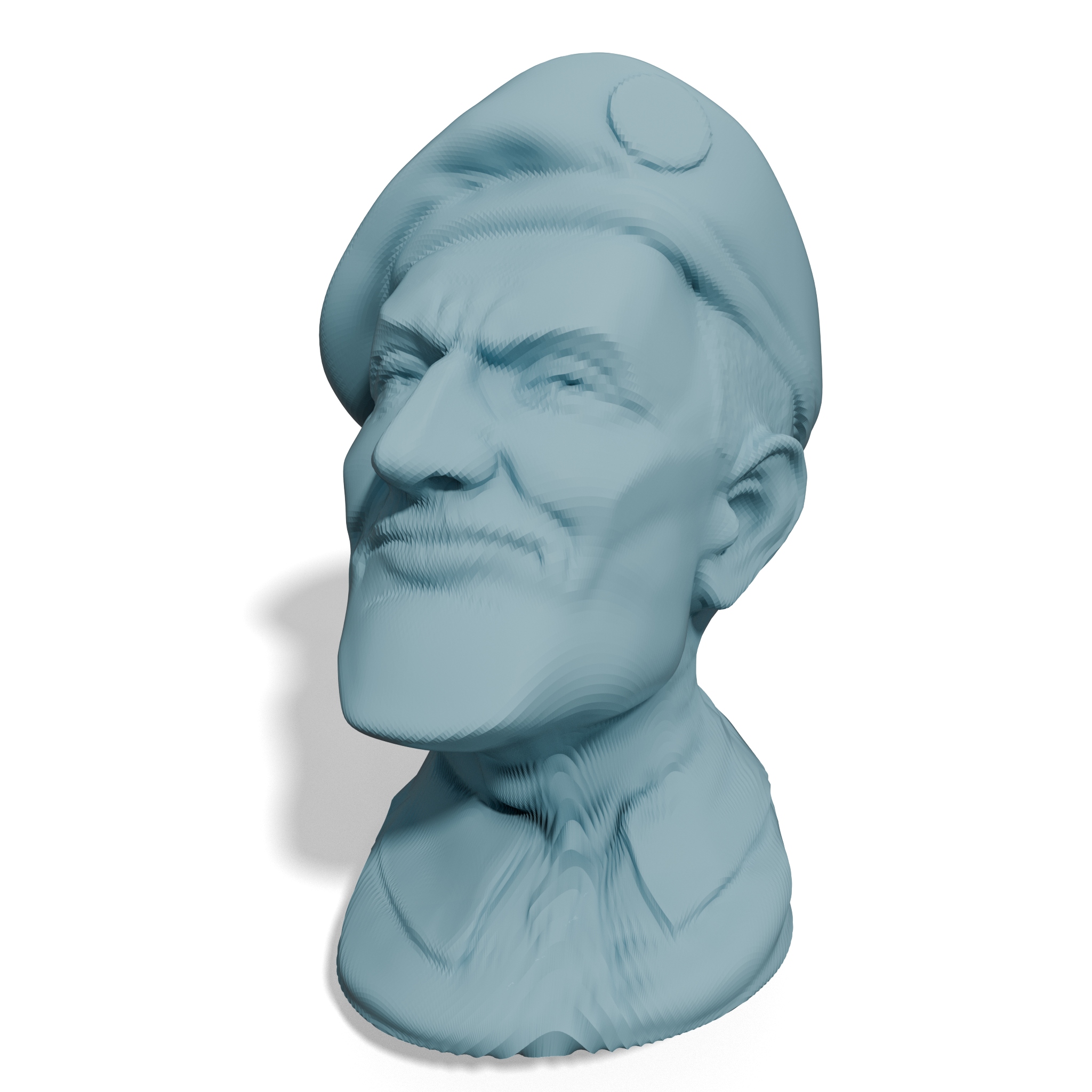}
                \end{minipage}
                \begin{minipage}{0.3\textwidth}
                    \includegraphics[width=\linewidth]{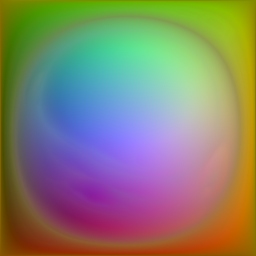}\\
                    \includegraphics[width=\linewidth]{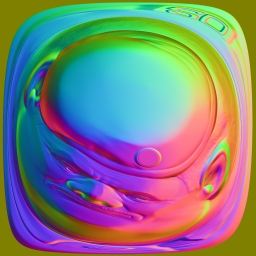}
                \end{minipage}
            \end{tabular}\\[2pt]
            16 / 1.3704 / 2.8340 \\
            Level 2 ($256 \times 256$)
        \end{minipage}
        &
        \begin{minipage}{0.19\textwidth}
            \centering
            \begin{tabular}{c@{\hskip -10pt}c}
                \begin{minipage}{0.65\textwidth}
                    \includegraphics[trim={300 0 250 0},clip,width=\linewidth]{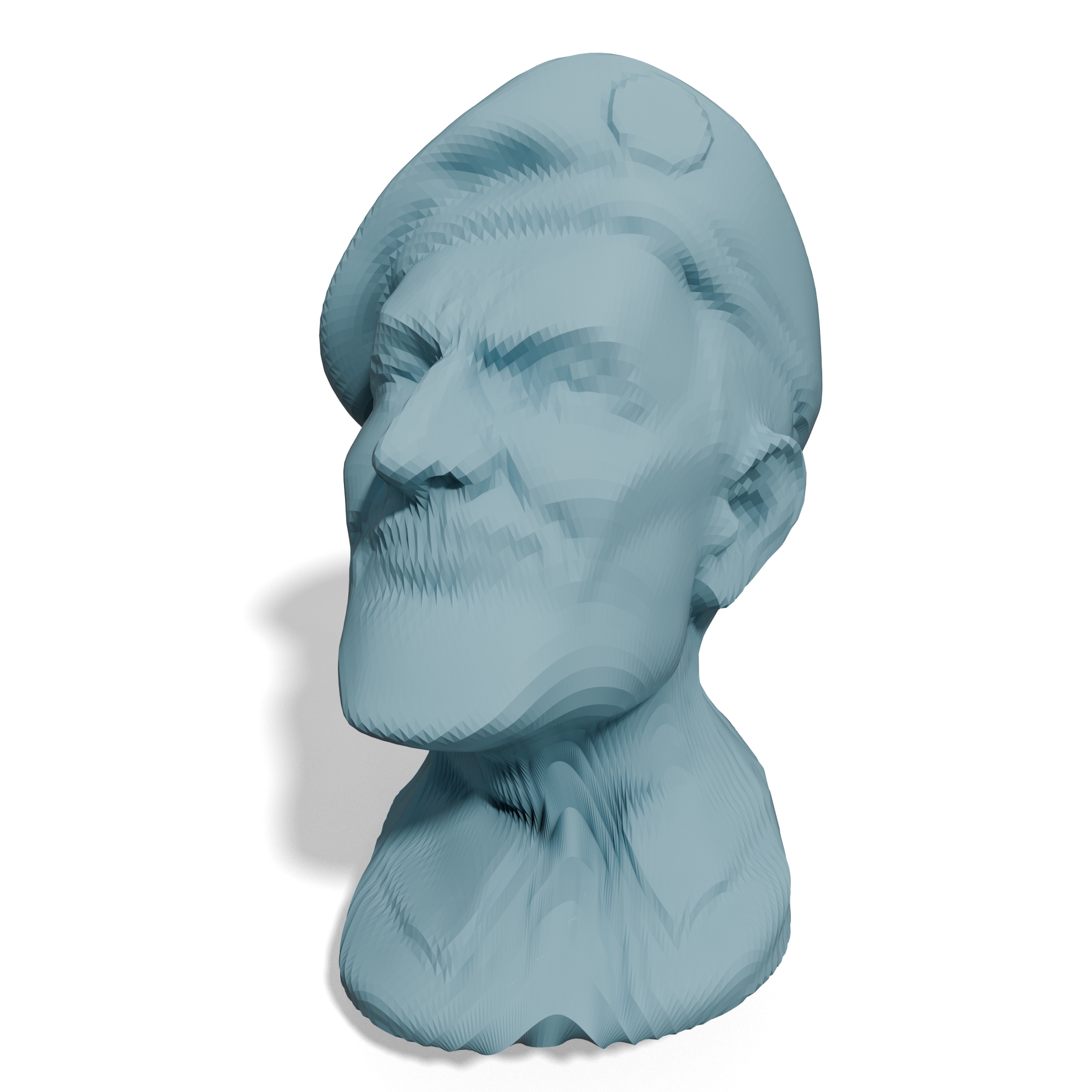}
                \end{minipage}
                \begin{minipage}{0.3\textwidth}
                    \includegraphics[width=\linewidth]{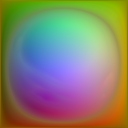}\\
                    \includegraphics[width=\linewidth]{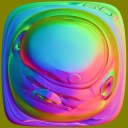}
                \end{minipage}
            \end{tabular}\\[2pt]
            64 / 1.5413 / 4.2496 \\
            Level 3 ($128 \times 128$)
        \end{minipage}
        \\
        % ---------- LoD 4–7 ----------
        &
        \begin{minipage}{0.19\textwidth}
            \centering
            \begin{tabular}{c@{\hskip -10pt}c}
                \begin{minipage}{0.65\textwidth}
                    \includegraphics[trim={300 0 250 0},clip,width=\linewidth]{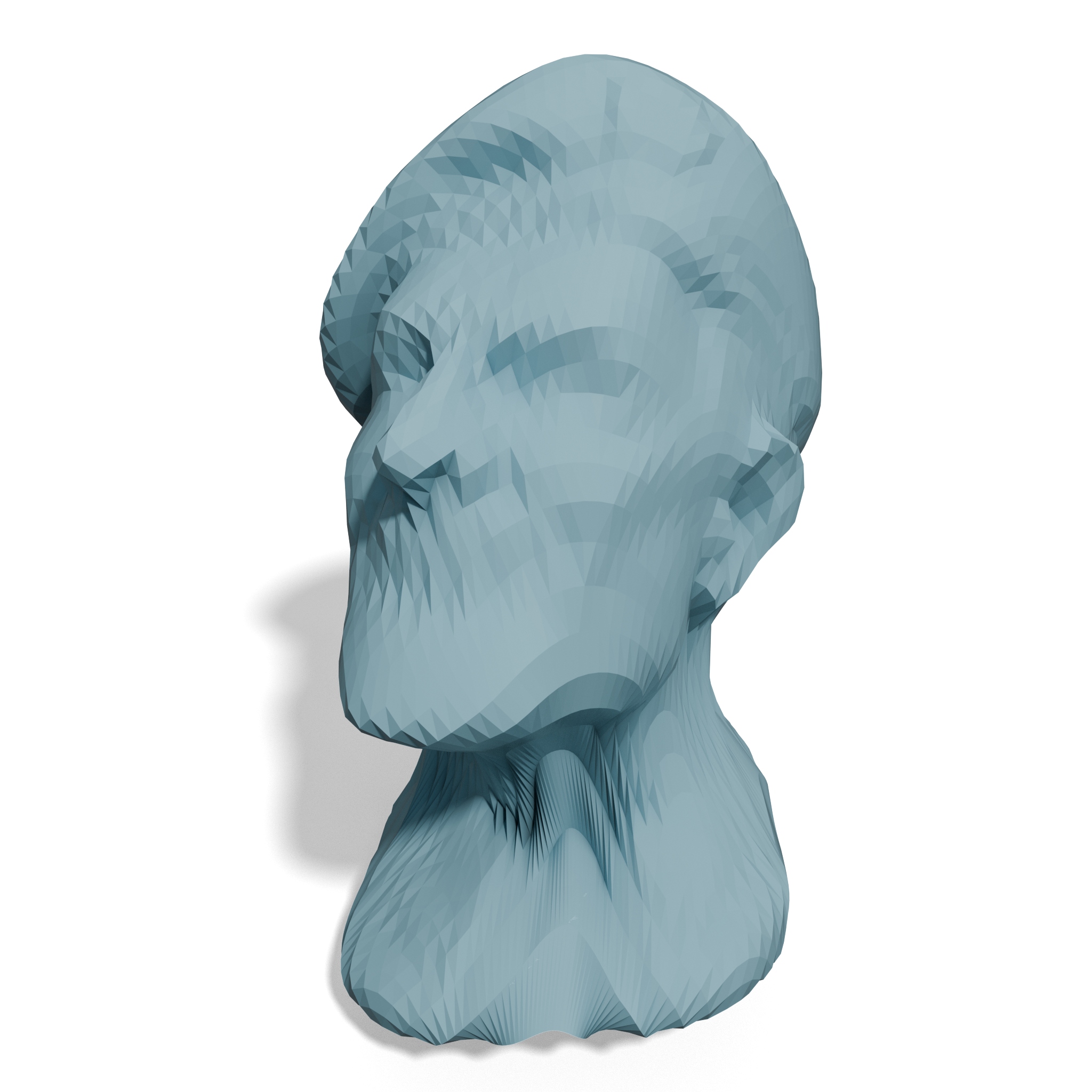}
                \end{minipage}
                \begin{minipage}{0.3\textwidth}
                    \includegraphics[width=\linewidth]{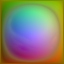}\\
                    \includegraphics[width=\linewidth]{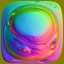}
                \end{minipage}
            \end{tabular}\\[2pt]
            256 / 2.4289 / 7.1198 \\
            Level 4 ($64 \times 64$)
        \end{minipage}
        &
        \begin{minipage}{0.19\textwidth}
            \centering
            \begin{tabular}{c@{\hskip -10pt}c}
                \begin{minipage}{0.65\textwidth}
                    \includegraphics[trim={300 0 250 0},clip,width=\linewidth]{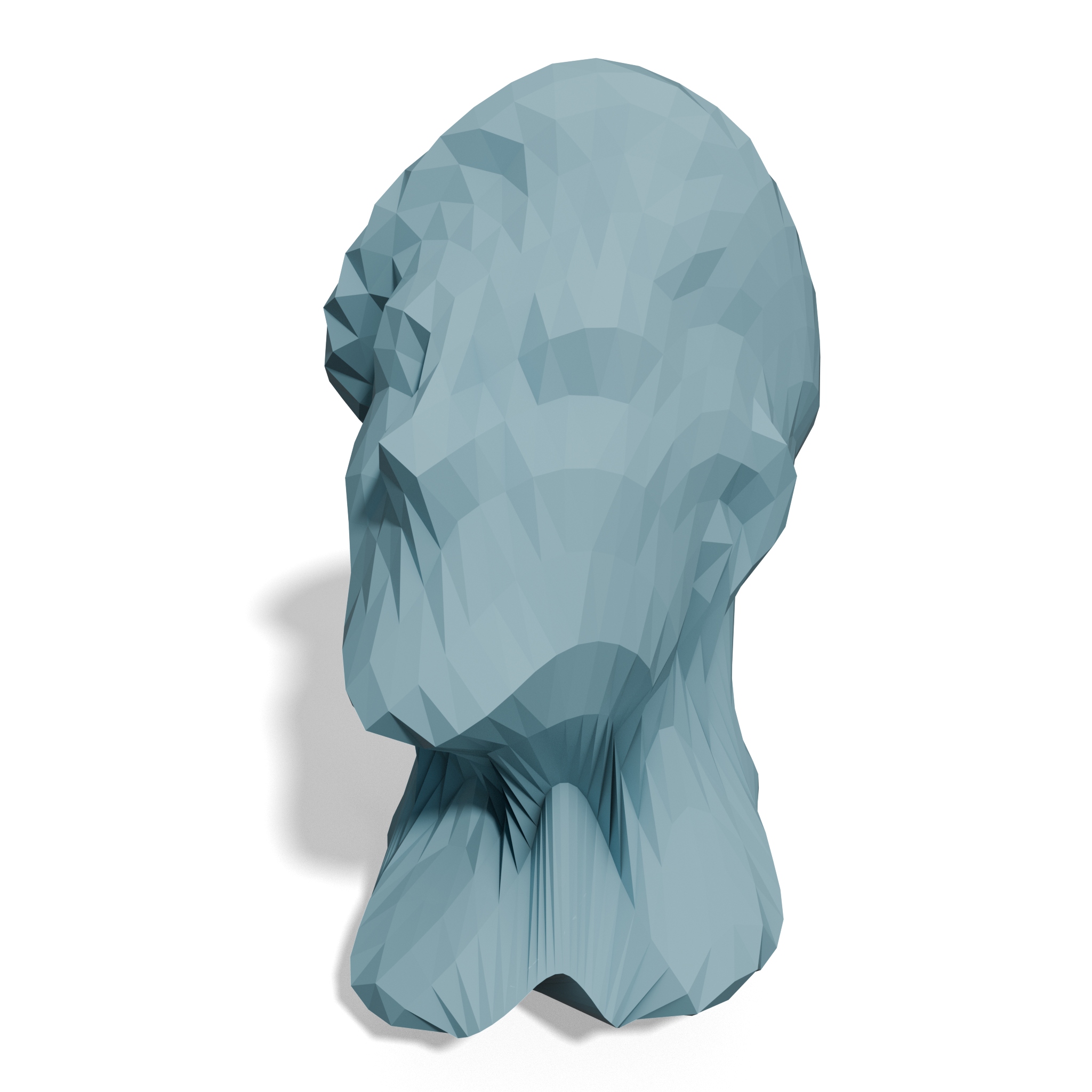}
                \end{minipage}
                \begin{minipage}{0.3\textwidth}
                    \includegraphics[width=\linewidth]{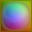}\\
                    \includegraphics[width=\linewidth]{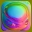}
                \end{minipage}
            \end{tabular}\\[2pt]
            1024 / 8.6698 / 13.3645 \\
            Level 5 ($32 \times 32$)
        \end{minipage}
        &
        \begin{minipage}{0.19\textwidth}
            \centering
            \begin{tabular}{c@{\hskip -10pt}c}
                \begin{minipage}{0.65\textwidth}
                    \includegraphics[trim={300 0 250 0},clip,width=\linewidth]{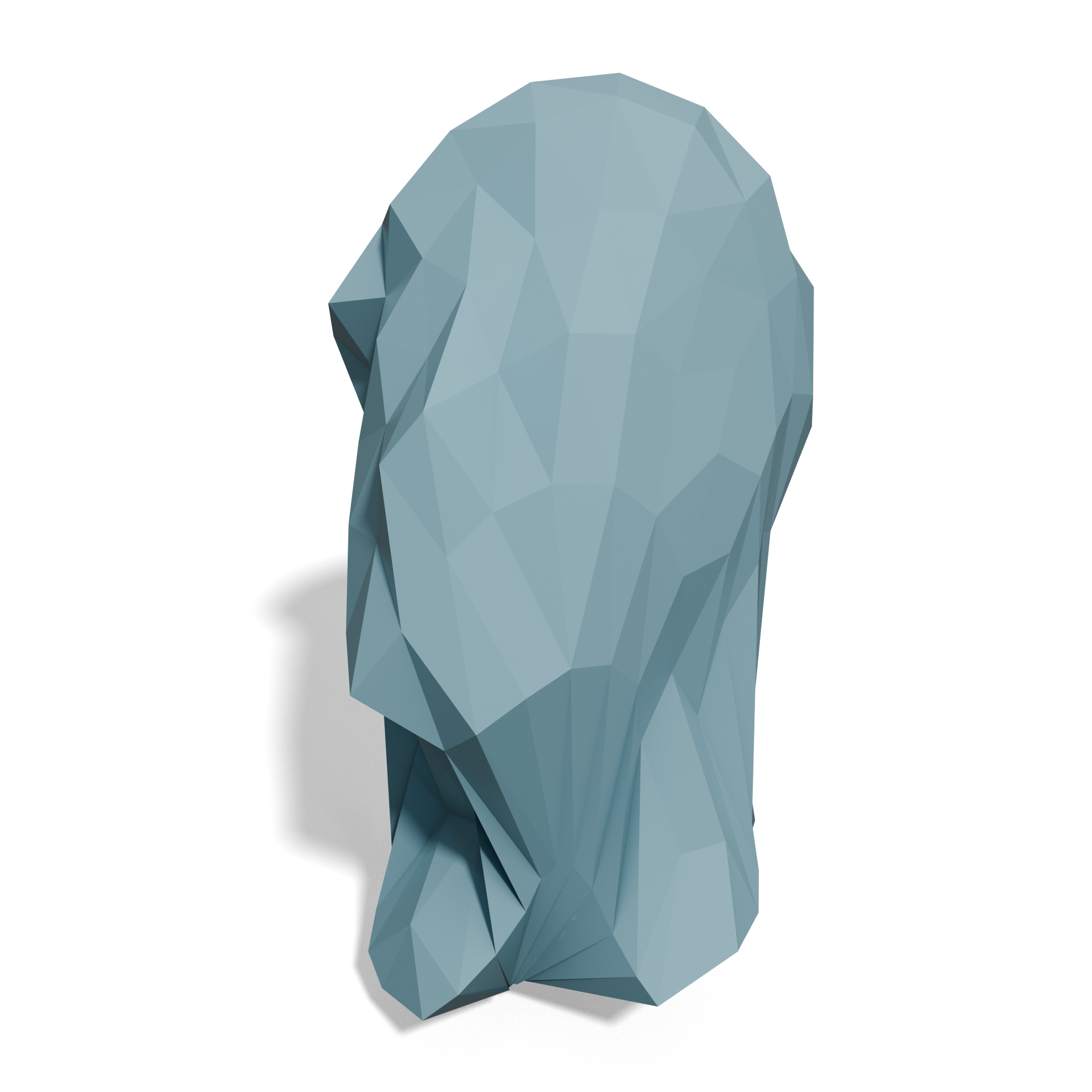}
                \end{minipage}
                \begin{minipage}{0.3\textwidth}
                    \includegraphics[width=\linewidth]{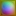}\\
                    \includegraphics[width=\linewidth]{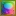}
                \end{minipage}
            \end{tabular}\\[2pt]
            4096 / 34.1197 / 19.4329 \\
            Level 6 ($16 \times 16$)
        \end{minipage}
        &
        \begin{minipage}{0.19\textwidth}
            \centering
            \begin{tabular}{c@{\hskip -10pt}c}
                \begin{minipage}{0.65\textwidth}
                    \includegraphics[trim={300 0 250 0},clip,width=\linewidth]{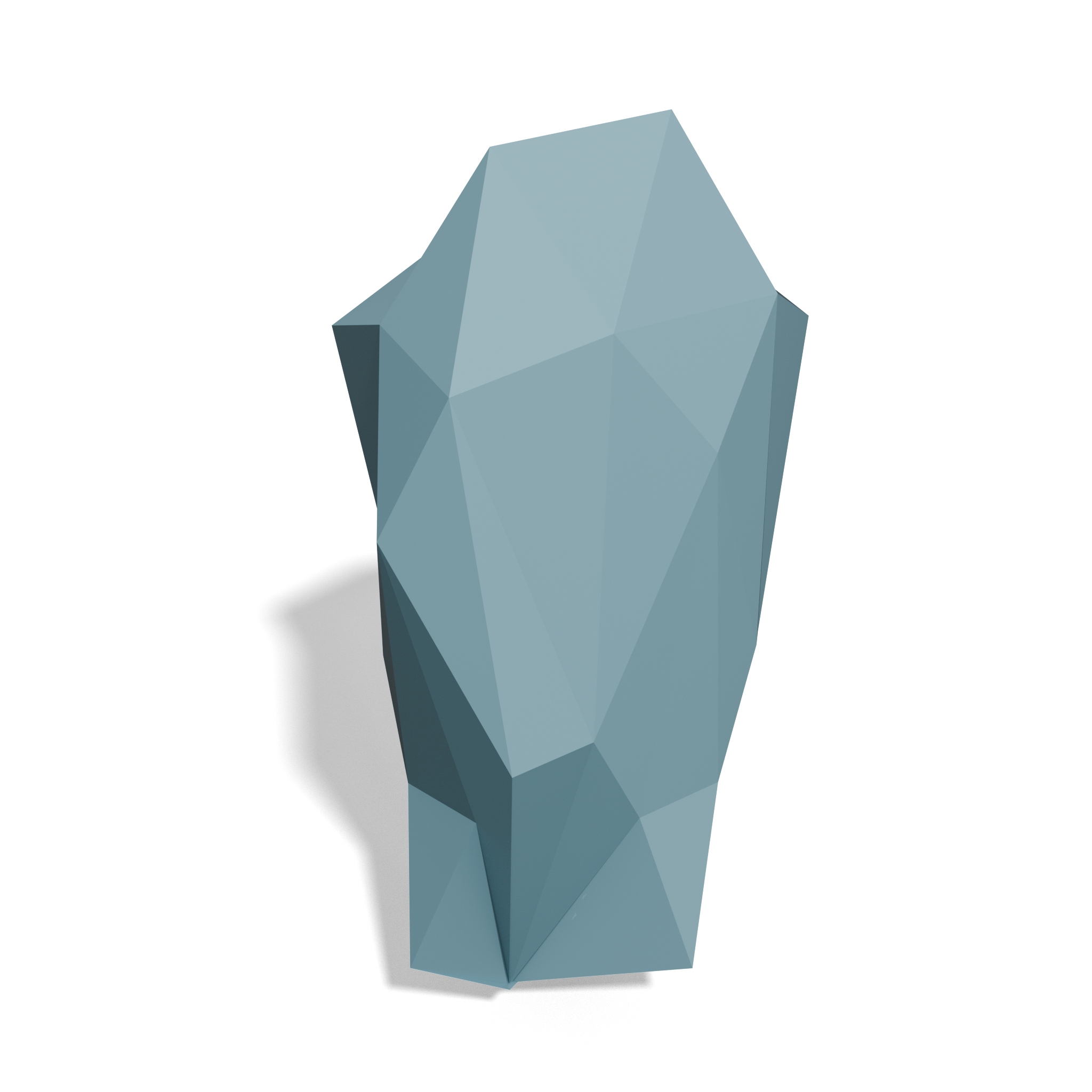}
                \end{minipage}
                \begin{minipage}{0.3\textwidth}
                    \includegraphics[width=\linewidth]{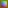}\\
                    \includegraphics[width=\linewidth]{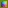}
                \end{minipage}
            \end{tabular}\\[2pt]
            16384 / 137.0457 / 39.4697 \\
            Level 7 ($8 \times 8$)
        \end{minipage}
    \end{tabular}

 % only the tabular/minipage code you pasted
\captionof{figure}{\textbf{Continuous Levels of Detail.}
Our neural geometry image–based representation with Optimal Transport (OT) restores full-resolution geometry images at Level~0 from a low-resolution input (Level~7) in a single forward pass. By leveraging OT, we avoid oversampling in flat regions and undersampling of facial details, a property we call \textit{uniform sampling}. Both rows show 3D meshes reconstructed from geometry image mipmaps across different levels. This demonstrates continuous levels of detail \textit{without the need for decoders}, exploiting GPU-optimized image mipmapping and achieving state-of-the-art storage efficiency measured by Compression Ratio (CR) and reconstruction accuracy measured by Chamfer Distance (CD) and Hausdorff Distance (HD).}
\label{fig:teaser}

\end{center}
}]

\begin{abstract}
Neural representations for 3D meshes are emerging as an effective solution for compact storage and efficient processing. Existing methods often rely on neural overfitting, where a coarse mesh is stored and progressively refined through multiple decoder networks. While this can restore high-quality surfaces, it is computationally expensive due to successive decoding passes and the irregular structure of mesh data. In contrast, images have a regular structure that enables powerful super-resolution and restoration frameworks, but applying these advantages to meshes is difficult because their irregular connectivity demands complex encoder–decoder architectures. Our key insight is that a \emph{geometry image–based representation} transforms irregular meshes into a regular image grid, making efficient image-based neural processing directly applicable. Building on this idea, we introduce \emph{our neural geometry image–based representation}, which is decoder-free, storage-efficient, and naturally suited for neural processing. It stores a low-resolution \emph{geometry-image mipmap} of the surface, from which high-quality meshes are restored in a single forward pass. To construct geometry images, we leverage Optimal Transport (OT), which resolves oversampling in flat regions and undersampling in feature-rich regions, and enables continuous levels of detail (LoD) through geometry-image mipmapping. Experimental results demonstrate state-of-the-art storage efficiency and restoration accuracy, measured by compression ratio (CR), Chamfer distance (CD), and Hausdorff distance (HD).

\begingroup
\renewcommand\thefootnote{}\footnotetext{*Authors contributed equally to this research.}
\endgroup

\end{abstract}
\section{Introduction}
Storing dense 3D surface meshes is a fundamental challenge in many real world applications such as VR/AR, gaming, CAD, robotics, and medical imaging, where models may contain millions of vertices and triangles. Recent advances in 3D reconstruction~\cite{mildenhall2020nerf, Shi_2025_CVPR, Neus, 2dgs24, guo2024tetsphere, Gao2025GraphCutUnwrapping}, generative modeling~\cite{dreamFusion22, liu2023zero1to3, liu2023syncdreamer, long2023wonder3d, MvDream23, Wang2025OTTALK}, and inverse rendering~\cite{Nicolet2021Large, mehta2022level, Jung2023} have produced highly detailed 3D shapes represented as point clouds~\cite{nichol2022pointe, pointflow, mandikal20183dlmnet, Groueix_2018_CVPR}, voxels~\cite{voxels, VoxNet15, ocnn17, SparseConvNet17}, and signed distance fields (SDFs)~\cite{neuralflow, deepSDF19, jiang2020sdfdiff, WangSDF2024}.
However, \textit{mesh-based} representations remain the preferred choice for their geometric precision, topological consistency, and flexibility. They are widely adopted in graphics since modern rendering pipelines and GPUs are optimized for mesh processing, and they provide a mathematically grounded discretization of continuous surfaces. This makes them indispensable in applications such as finite element analysis (FEA)~\cite{Sifakis2012FEM}, shape analysis, and physics simulation~\cite{BrennerScott1994}, as well as in discrete differential geometry~\cite{CCG08, Crane:2013:DGP, botsch2010polygon}, where preserving geometric detail and topological correctness is crucial.

Recently, the rapid growth of large-scale 3D datasets~\cite{chang2015shapenet, Thingi10K, objaverse, objaverseXL}, many of which contain extremely dense surface meshes, has intensified the demand for compact storage and efficient level-of-detail (LoD) representations. Such dense meshes not only incur significant storage overhead but also hinder real-time rendering, streaming, and interaction, particularly in game engines and AR/VR systems where high-resolution meshes are standard. Therefore, efficient storage techniques are critical not only for reducing memory usage but also for enabling deployment on devices with limited memory and computational capacity, such as mobile phones, AR/VR headsets, and embedded systems.

To address this need, state-of-the-art methods such as Neural Geometric Level of Detail (NGLOD)~\cite{takikawa2021nglod}, Neural Surface Meshing (NCS)~\cite{Morreale2022NCS}, Neural Subdivision (NS)~\cite{LIU2020NeuralSubdivision}, and Neural Progressive Meshes (NPM)~\cite{CHEN2023NeuralCompression} employ Encoder–Decoder–based neural architectures to learn \textbf{compact, low-resolution representations of 3D surfaces} that can be progressively refined into high-quality geometry. However, these approaches face two fundamental limitations. First, they require multiple forward passes through decoder networks, which makes the recovery of a dense 3D surface mesh from a sparse one computationally expensive. Second, their level-of-detail is realized through successive decoder stages, where each level demands a separate decoding pass. This design significantly increases computational overhead and makes real-time LoD control impractical. Furthermore, since many downstream tasks such as relighting, rendering, and physics simulation require mesh-based 3D surface representations, these methods often incur additional overhead from converting their learned representations back into mesh format. As a result, they struggle to jointly achieve storage efficiency, accurate surface restoration, and efficient LoD control.

To overcome these challenges, we propose a novel \textit{neural geometry image-based representation} that stores a \textbf{compact, low-resolution, area-preserving geometry-image mipmap} of the 3D surface mesh and restores a high-quality 3D surface mesh in a single forward pass. In contrast to existing approaches that require specially designed mesh encoder–decoder architectures to process the irregular structure of 3D surface mesh connectivity and rely on successive decoding passes, our method exploits the regular structure of the image, making it naturally compatible with efficient image-based neural processing. By leveraging Optimal Transport (OT) to construct area-preserving geometry images, we resolve sampling imbalance by avoiding oversampling in flat regions and undersampling in feature-rich regions. As a direct consequence of this construction, the representation supports continuous levels of detail at negligible computational cost on GPUs. This results in a decoder-free, storage-efficient, and computationally efficient \textit{neural geometry-image–based representation} that achieves accurate 3D surface mesh restoration while enabling continuous LoD representations. \\
\noindent
To summarize, we offer \textbf{three principal contributions:} 
\begin{itemize}[noitemsep, topsep=0pt, leftmargin=*]
    \item We propose a novel neural geometry-image–based representation that efficiently stores 3D surface meshes as compact, low-resolution geometry-image mipmaps and restores high-quality meshes in a single forward pass, avoiding multiple decoder stages.
    \item We apply Optimal Transport (OT) to construct area-preserving geometry images, which resolve sampling imbalance, achieving continuous level-of-detail representations at negligible cost on modern GPU.
    \item We demonstrate state-of-the-art efficiency and accuracy, achieving lower memory cost and superior mesh reconstruction quality in terms of Chamfer Distance (CD) and Hausdorff Distance (HD) compared to existing methods.
\end{itemize}

\section{Related Works}
\subsection{Geometry Image}
Geometry images, introduced by Gu et al.~\cite{Gu2002GeometryImages}, provide a structured 2D image-based representation of irregular 3D surface meshes by parameterizing them onto a square domain. This format enables efficient storage, processing, and rendering using standard image-based techniques. For example, Carr et al.~\cite{Carr2006Fast} use geometry images to accelerate GPU-based ray tracing and achieve efficient level-of-detail rendering at no additional cost through mipmapping. Similarly, Sander et al.~\cite{Kraevoy2003Matchmaker} decompose meshes into multiple charts and construct multi-chart geometry images to support continuous level-of-detail texture mapping, also at negligible cost through mipmapping. While such strategies are effective for rendering and texture mapping, extending geometry image mipmapping to 3D surface meshes remains difficult due to sampling imbalance, where flat regions are oversampled and feature-rich regions are undersampled.

\begin{figure}[t]
    \centering
    \includegraphics[width=1.05\linewidth]{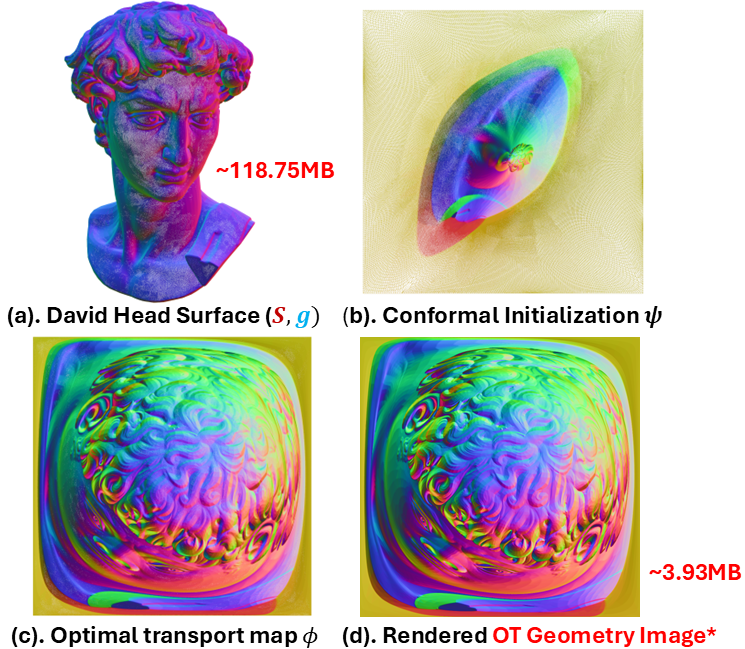}
\caption{\textbf{Overview of OT for geometry images.} 
A 3D surface mesh (a) is flattened into the 2D domain by conformal initialization (b), preserving angles. 
Optimal Transport refinement (c) applies a measure-preserving map that redistributes area for balanced sampling. 
The final OT-based geometry image (d) provides a compact, area-preserving representation of the 3D surface mesh for efficient storage and neural processing.}
    \label{fig:ot_pipeline}
\end{figure}

\begin{figure*}[t]

  \includegraphics[width=\textwidth]{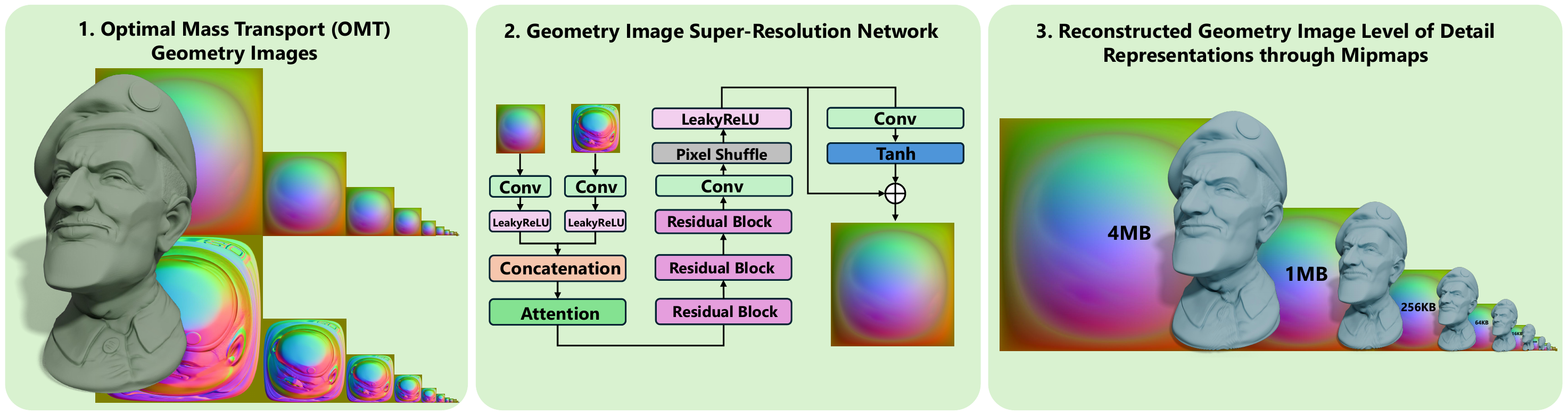}\\[-5pt] \vspace{-5mm}
    \caption{\textbf{Neural Geometry Image-Based Representation with Optimal Transport (OT).} 
    Our pipeline is \textit{decoder-free}, memory-efficient, and enables continuous geometric Level-of-Detail (LoD) entirely on the GPU without progressive refinement. 
    (1) Geometry images are first constructed using Optimal Mass Transport (OMT) to preserve surface area and regularize sampling across irregular meshes. 
    (2) A geometry image super-resolution network directly upsamples a low-resolution mipmapped geometry image (position and normal maps) in a single forward pass, avoiding the need for multiple decoders. 
    (3) The reconstructed geometry image is hierarchically mipmapped, yielding a pyramid of geometry images that allows continuous LoD control. At runtime, the GPU can seamlessly select and render any desired resolution, balancing storage and rendering constraints while preserving both global shape and local detail.}
  \label{fig:pipeline}

\end{figure*}

\subsection{Surface Parameterization}
Surface parameterization is a core technique in geometry processing that maps irregular 3D surface meshes onto a 2D domain, enabling tasks such as texture mapping, remeshing, and geometry image generation. One of the earliest methods is Tutte’s embedding~\cite{Tutte1963HowTD}, which guarantees a valid planar mapping by solving a linear system with uniform edge weights under convex boundary constraints. Although bijective, it neither preserves angles nor areas. Harmonic maps~\cite{Eck1995Harmonic} improve angular fidelity using cotangent weights but still lack strict conformality. Conformal methods such as Least Squares Conformal Maps (LSCM)~\cite{Levy2002LSCM}, holomorphic differentials~\cite{GU2003GlobalConformalParameterization}, and discrete Ricci flow~\cite{4483509} explicitly preserve angles but often introduce large area distortion. Conversely, authalic parameterizations~\cite{Desbrun2002Intrinsic} aim to preserve local area but cannot guarantee global consistency. Hybrid approaches attempt to balance angle and area distortion, but all the above-mentioned methods face the challenge of sampling imbalance when applied to generate a single geometry image from a 3D surface mesh: flat regions are typically oversampled while feature-rich regions are undersampled, degrading level-of-detail representations across mipmap levels.

\subsection{Neural Shape Representations}
A recent line of work explores \textbf{neural overfitting} as a strategy for the compact storage and reconstruction of 3D surface meshes. The key idea is to replace large, dense meshes with compact latent codes or low-resolution representations, and rely on decoder networks to progressively refine them into high-quality geometry. Neural Geometric Level of Detail (NGLOD)~\cite{takikawa2021nglod}, Neural Surface Meshing (NCS)~\cite{Morreale2022NCS}, Neural Subdivision (NS)~\cite{LIU2020NeuralSubdivision}, and Neural Progressive Meshes (NPM)~\cite{CHEN2023NeuralCompression} are representative examples. While these methods provide substantial storage savings, they suffer from two major limitations. First, dense 3D surface meshes must be reconstructed through multiple forward passes of decoder networks, making inference time-consuming. Second, their level-of-detail is realized through successive decoder stages, where each LoD requires a separate decoding pass. This design increases computational cost and makes real-time LoD control impractical. To overcome these limitations, a representation is needed that achieves compact storage while supporting single-pass reconstruction and efficient continuous LoD.

\begin{algorithm}[t] 
  \caption{Area Preserving Parameterization \label{algorithm:ot}}
  \begin{algorithmic}[1]
    \State \textbf{Input:} A planar rectangle with measure $(\Omega, \mu)$; 
           a point set with measure $(P, \nu)$ obtained from $\phi$; 
           a threshold $\epsilon$.
    \State \textbf{Output:} An area preserving map $f$.
    \State $\mathbf{h} \gets (|p_1|, |p_2|, \dots, |p_k|)^T$, $p_i \in P$
    \State Compute the power diagram $D(\mathbf{h})$
    \State Compute the dual power Delaunay triangulation $T(\mathbf{h})$
    \State Compute the cell areas $\mathbf{w}(\mathbf{h}) = (w_1(\mathbf{h}), \dots, w_k(\mathbf{h}))^T$
    \State Compute $\nabla E(\mathbf{h})$ using \cref{ot:grad}
    \While{$|\nabla E| > \epsilon$}
      \State Compute the Hessian matrix using \cref{ot:hessian}
      \State $\lambda \gets 1$
      \State $\mathbf{h} \gets \mathbf{h} - \lambda H^{-1} \nabla E(\mathbf{h})$
      \State Compute $D(\mathbf{h}), T(\mathbf{h}), \mathbf{w}(\mathbf{h})$
      \While{$\exists w_i(\mathbf{h}) = 0$}
        \State $\mathbf{h} \gets \mathbf{h} + \lambda H^{-1} \nabla E(\mathbf{h})$
        \State $\lambda \gets \lambda / 2$
        \State $\mathbf{h} \gets \mathbf{h} - \lambda H^{-1} \nabla E(\mathbf{h})$
        \State Compute $D(\mathbf{h}), T(\mathbf{h}), \mathbf{w}(\mathbf{h})$
      \EndWhile
      \State Compute $\nabla E(\mathbf{h})$ using \cref{ot:grad}
    \EndWhile
    \State Construct $\psi : \Omega \to P, W_i(\mathbf{h}) \mapsto p_i, i=1,\dots,k$
    \State \Return $\psi^{-1} \circ \phi$
  \end{algorithmic}
\end{algorithm}

\section{Neural Geometry Images with OT}
We introduce a neural geometry image-based representation that transforms irregular 3D meshes into structured, low-resolution geometry image mipmaps for efficient storage and neural processing. Conformal initialization preserves local angles, while optimal transport refinement corrects area distortion to guarantee balanced sampling. These compact mipmaps are stored and serve as inputs to a CNN-based network, which restores full-resolution geometry images in a single forward pass. Unlike prior approaches, this design circumvents decoder architectures and naturally supports continuous level-of-detail representations on GPUs. The overall pipeline is shown in Figure~\ref{fig:pipeline}.

\begin{figure}[t]
    \centering
    \includegraphics[width=\linewidth]{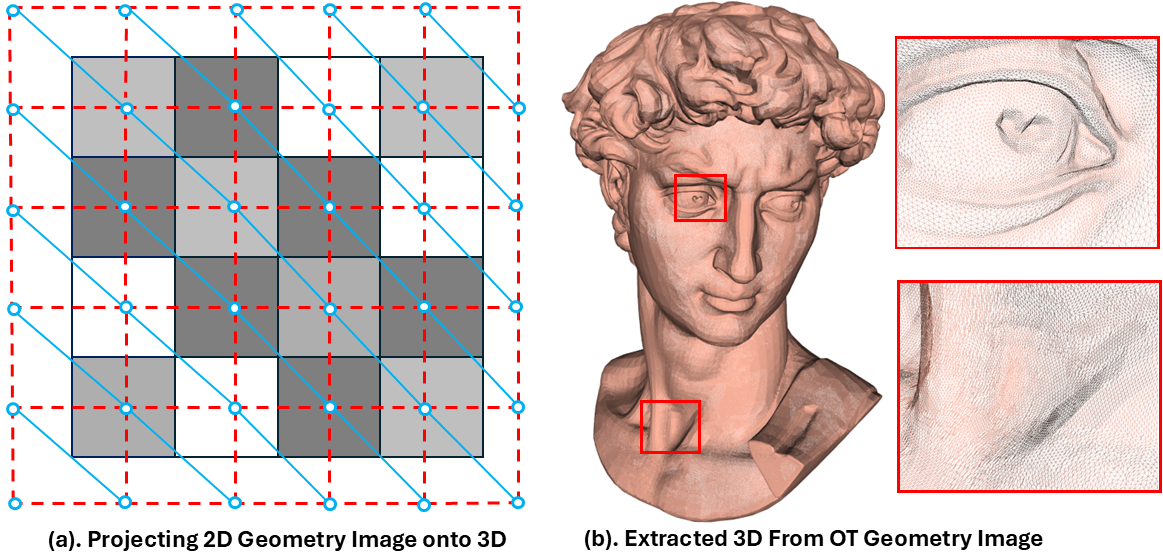}
    \caption{\textbf{Projection from 2D image space to 3D surface.} 
    The 2D OT-based geometry image is first triangulated in image space and then projected back to reconstruct the 3D surface mesh.}
    \label{fig:restore}
\end{figure}

\subsection{Optimal Transport for Geometry Images}
We construct geometry images in two stages. First, a conformal parameterization via Ricci flow maps the 3D surface onto a 2D domain. Next, optimal transport (OT) refinement corrects area distortion to enforce balanced, area-preserving sampling. The resulting bijective map is stored as a 16-bit RGB image encoding $(x,y,z)$ coordinates. Figure~\ref{fig:ot_pipeline} shows the construction process, while Figure~\ref{fig:restore} illustrates restoration through image-space triangulation and reprojection to 3D.

\vspace{-4mm}
\paragraph{Conformal Initialization.}  
We first obtain a conformal parameterization using the Ricci flow algorithm~\cite{4483509}. For a genus-zero mesh with a single boundary, four boundary vertices are fixed to the corners of a square and assigned target curvatures $\tfrac{\pi}{4}$, with zero elsewhere. The conformal factor $\mathbf{u}$ is obtained by minimizing the Ricci energy
\begin{equation}\label{formula:re}
    E(\mathbf{u}) = \int_{\mathbf{0}}^{\mathbf{u}} \sum_i \bigl(\Bar{K}_i - K_i\bigr)\, \mathrm{d}u_i ,
\end{equation}
where $K_i$ and $\Bar{K}_i$ denote the current and target Gaussian curvatures. Newton’s method is applied with gradient $(\Bar{K}_i-K_i)^T$ and Hessian
\begin{equation}\label{formula:hessian}
   \begin{cases}
    \tfrac{\partial K_i}{\partial u_i} = -\sum_j w_{ij}, & \text{on the diagonal}, \\
    \tfrac{\partial K_i}{\partial u_j} = \tfrac{\partial K_j}{\partial u_i} = w_{ij}, & \text{elsewhere}.
   \end{cases}
\end{equation}
After convergence, the mesh is flattened conformally to a square domain. Conformal maps preserve angles but not area, leading to oversampling in flat regions and undersampling in curved regions.

\begin{figure*}[ht]
  \centering
  \setlength{\resLen}{0.18\textwidth}

  \begin{tabular}{c@{\hskip -2pt}cc@{\hskip -2pt}cc@{\hskip -2pt}cc@{\hskip -2pt}cc@{\hskip -2pt}c}
    \multicolumn{2}{c}{\includegraphics[trim={300 0 150 0},clip,width=\resLen]{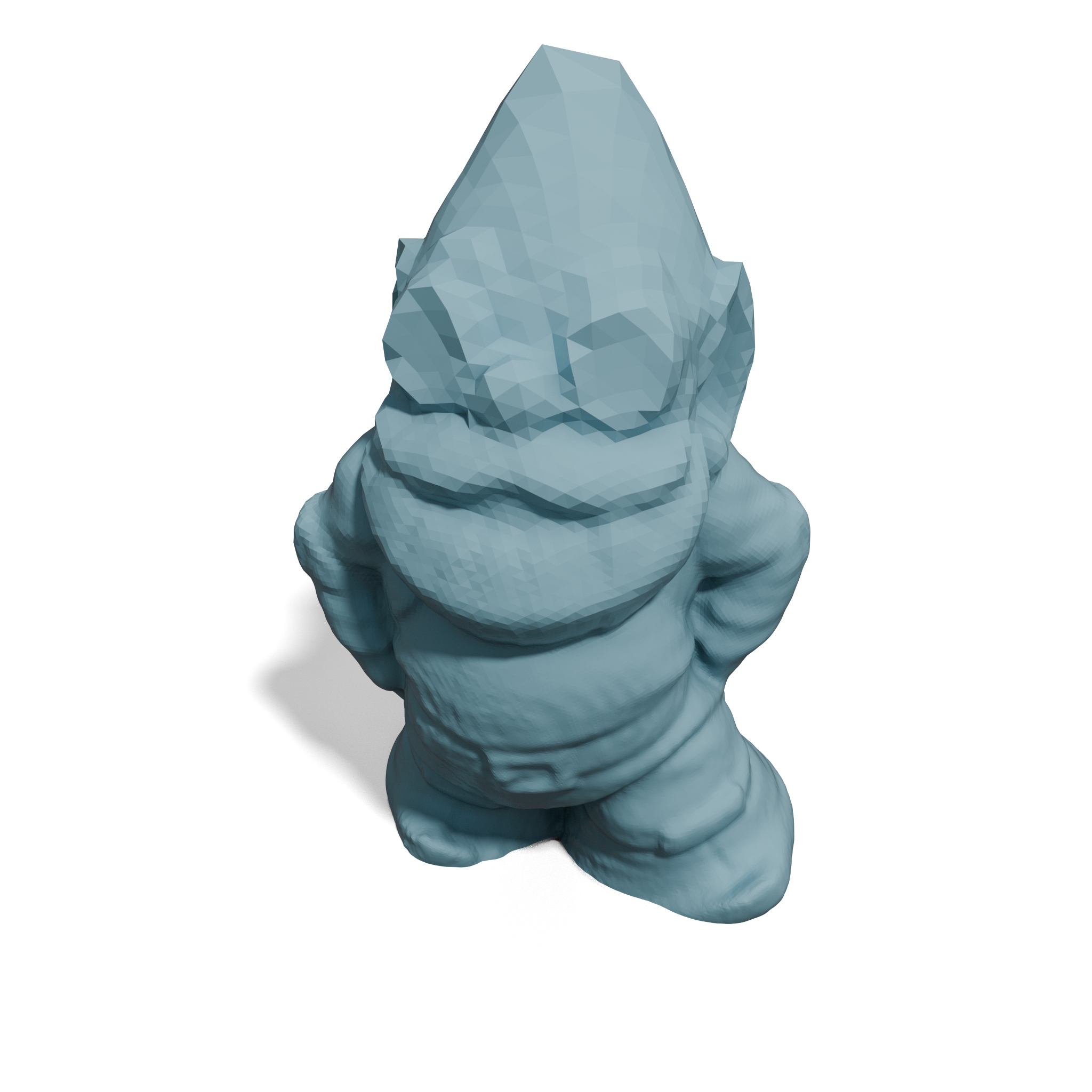}} &
    \multicolumn{2}{c}{\includegraphics[trim={300 0 150 0},clip,width=\resLen]{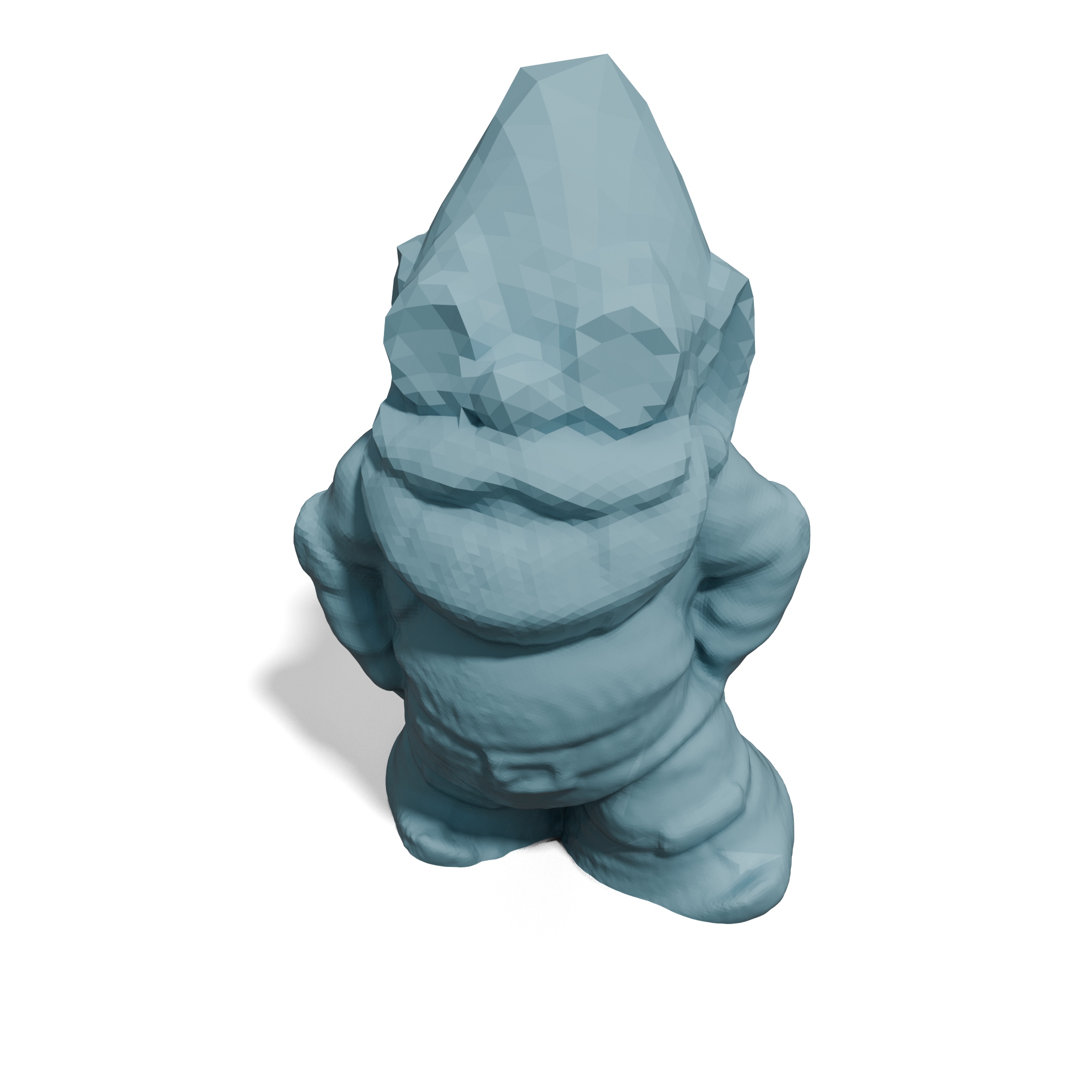}} &
    \multicolumn{2}{c}{\includegraphics[trim={300 0 150 0},clip,width=\resLen]{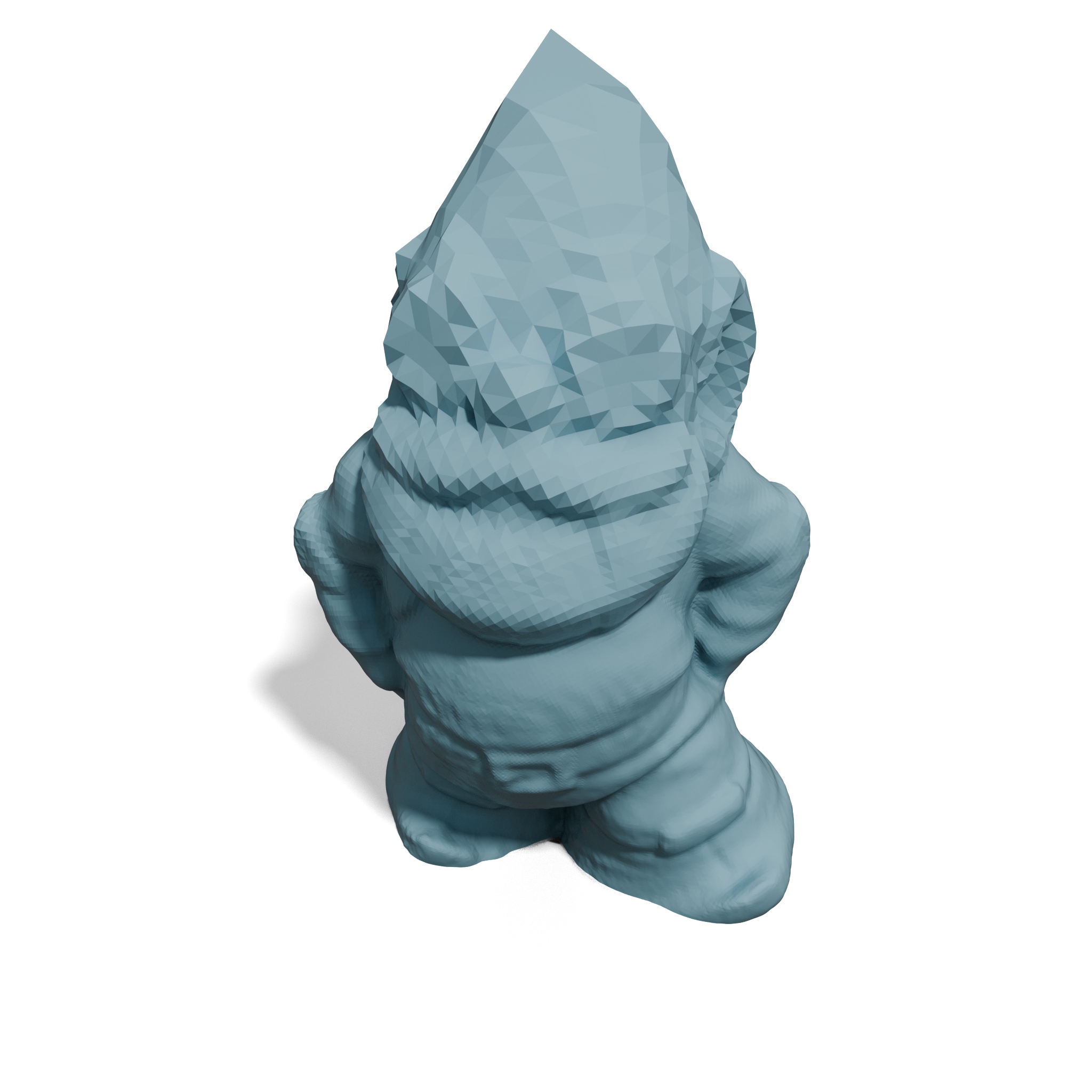}} &
    \multicolumn{2}{c}{\includegraphics[trim={300 0 150 0},clip,width=\resLen]{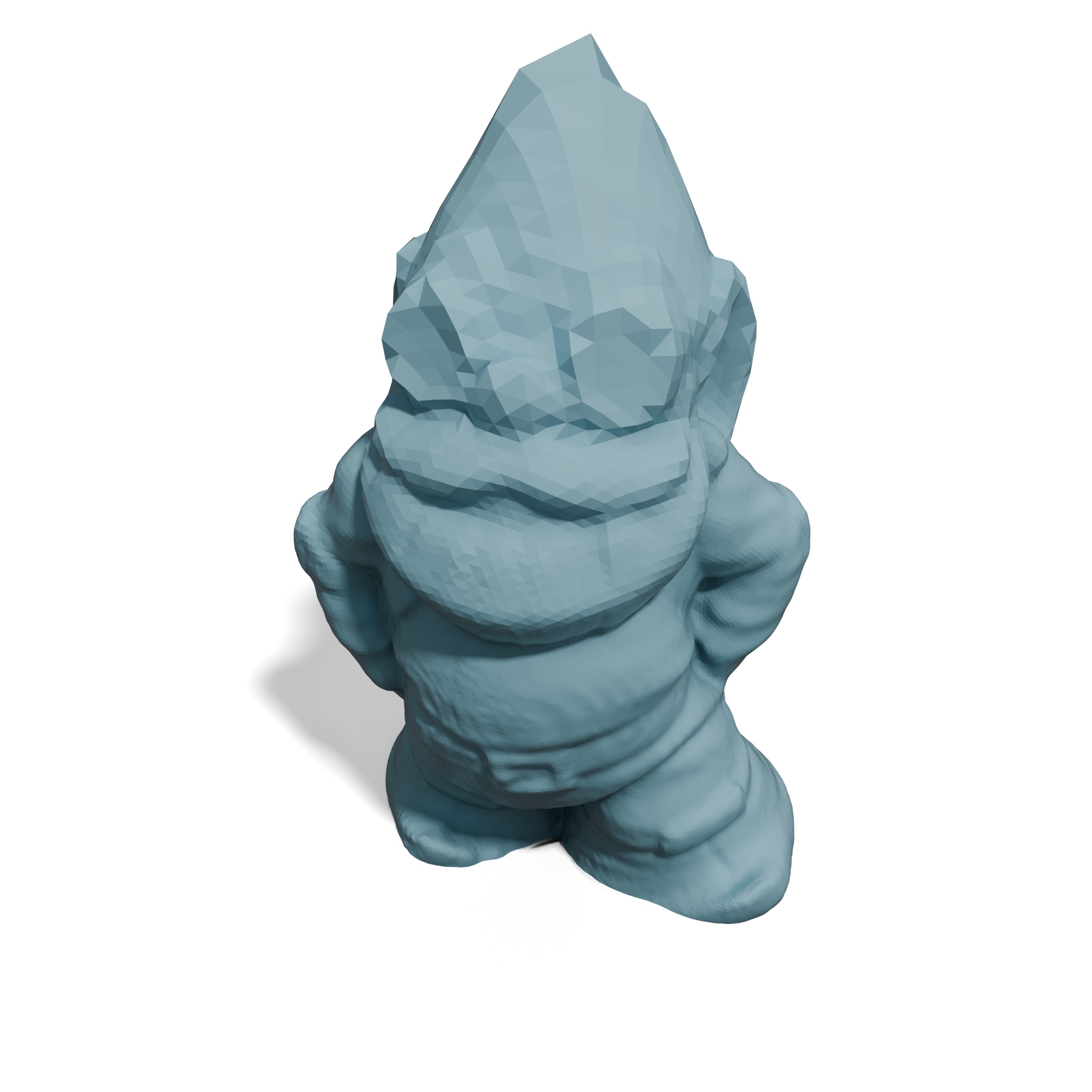}} &
    \multicolumn{2}{c}{\includegraphics[trim={300 0 150 0},clip,width=\resLen]{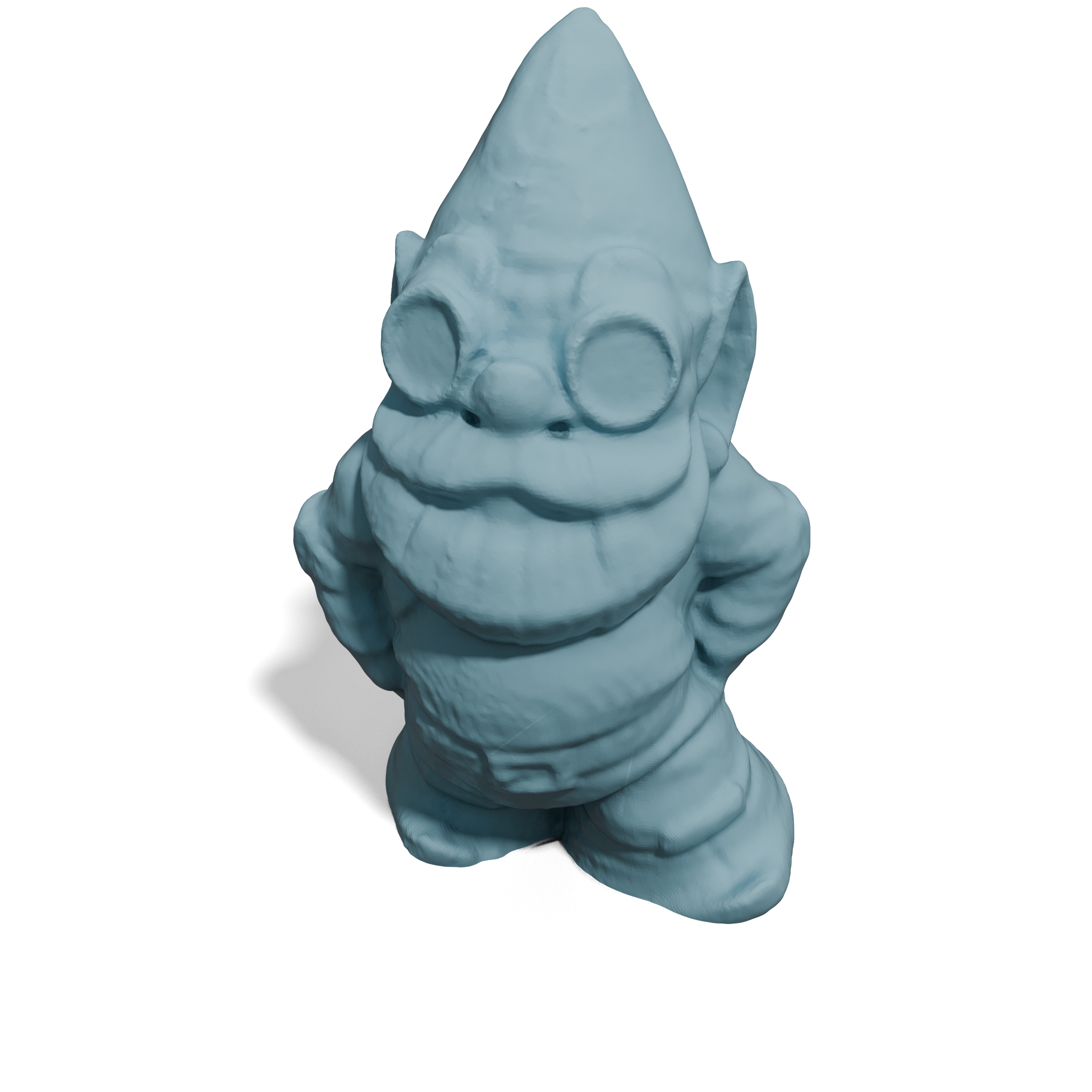}} \\[-15pt]

    \includegraphics[width=0.45\resLen]{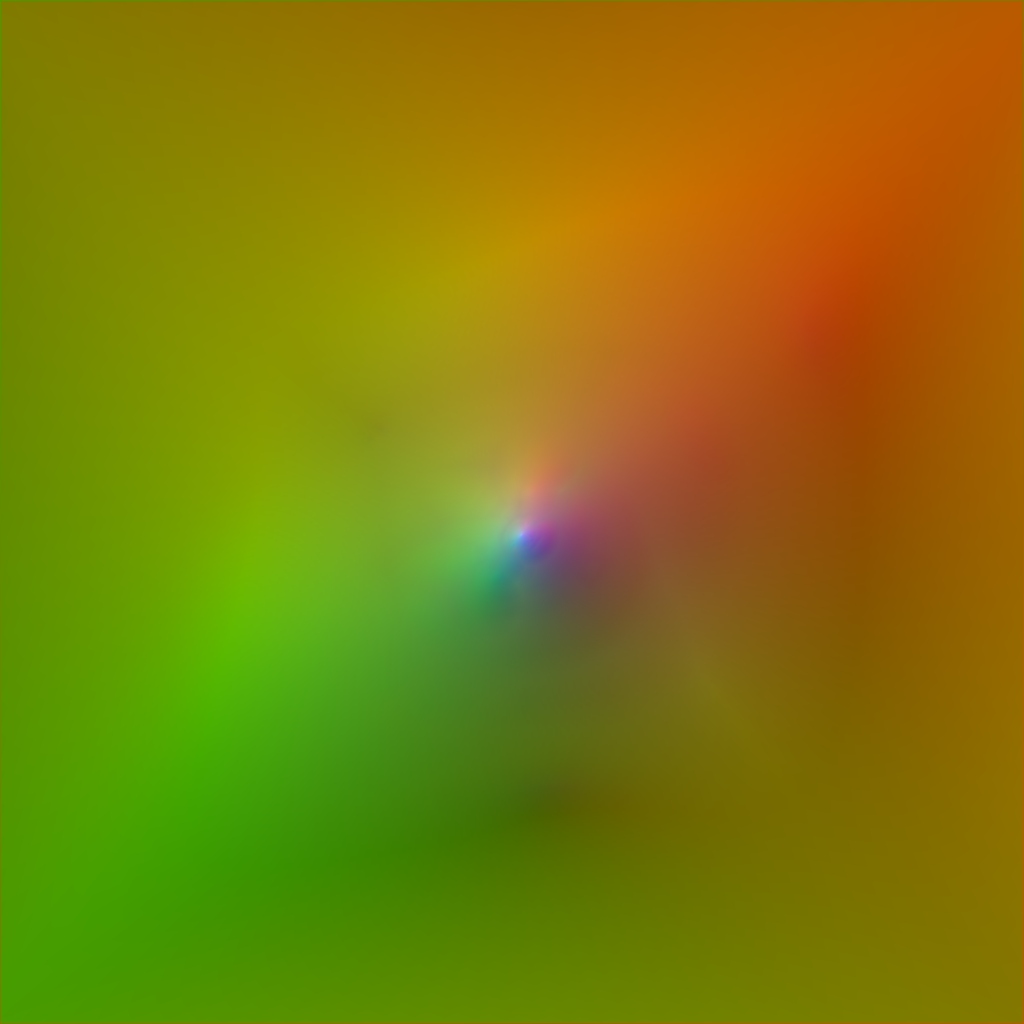} &
    \includegraphics[width=0.45\resLen]{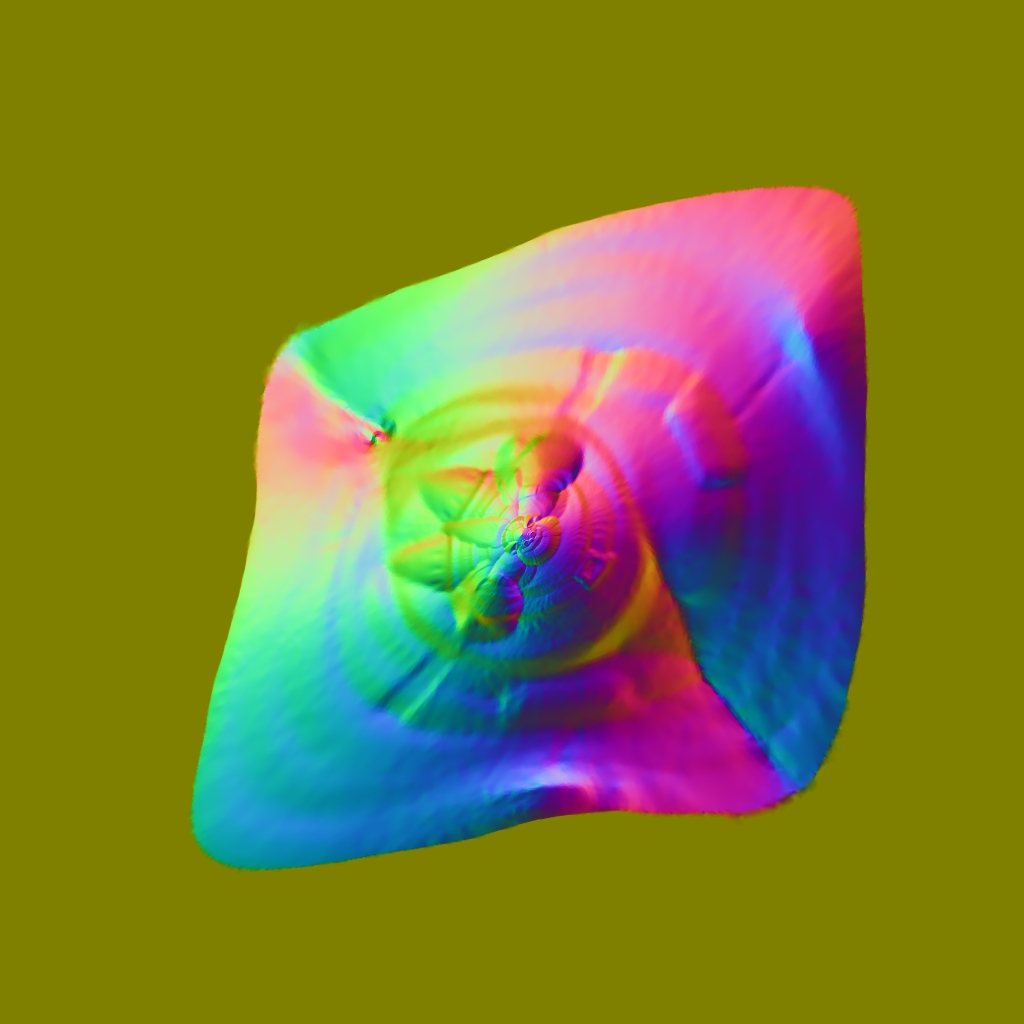} &
    \includegraphics[width=0.45\resLen]{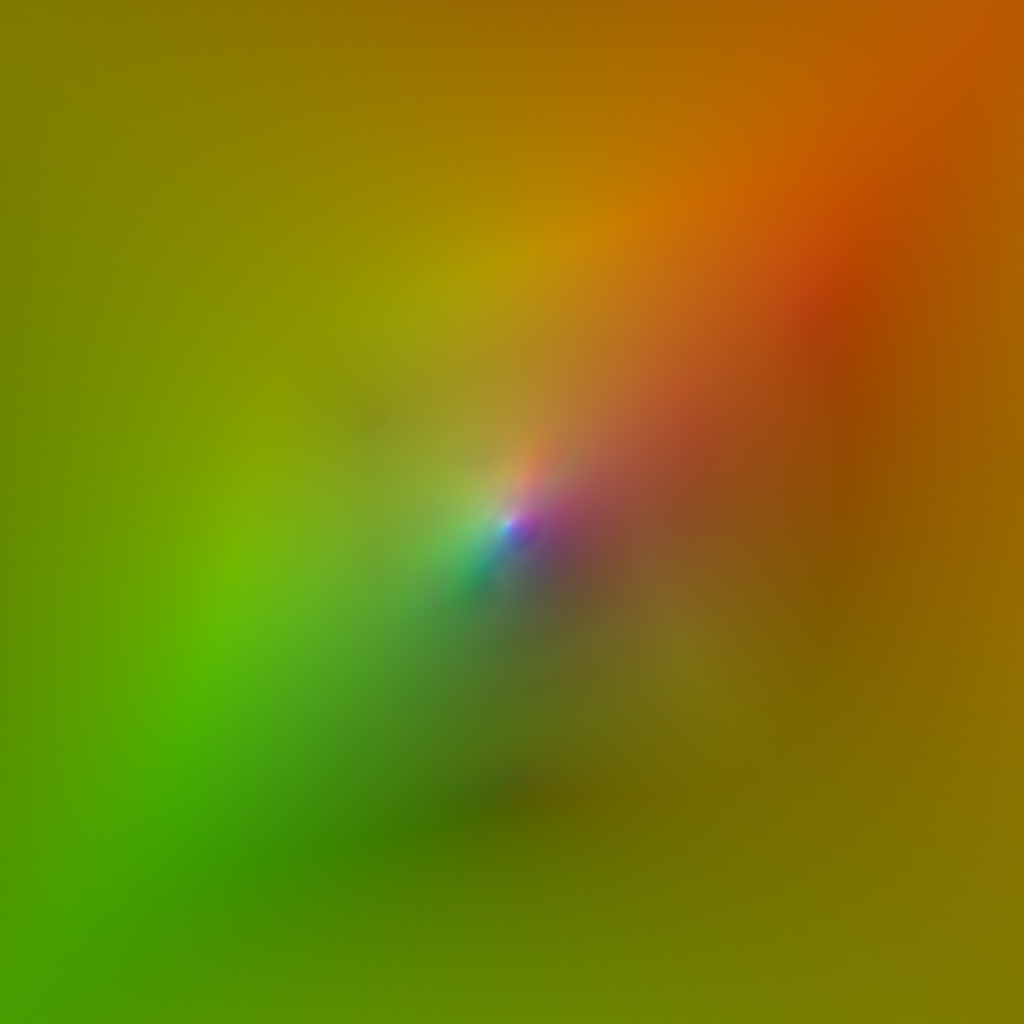} &
    \includegraphics[width=0.45\resLen]{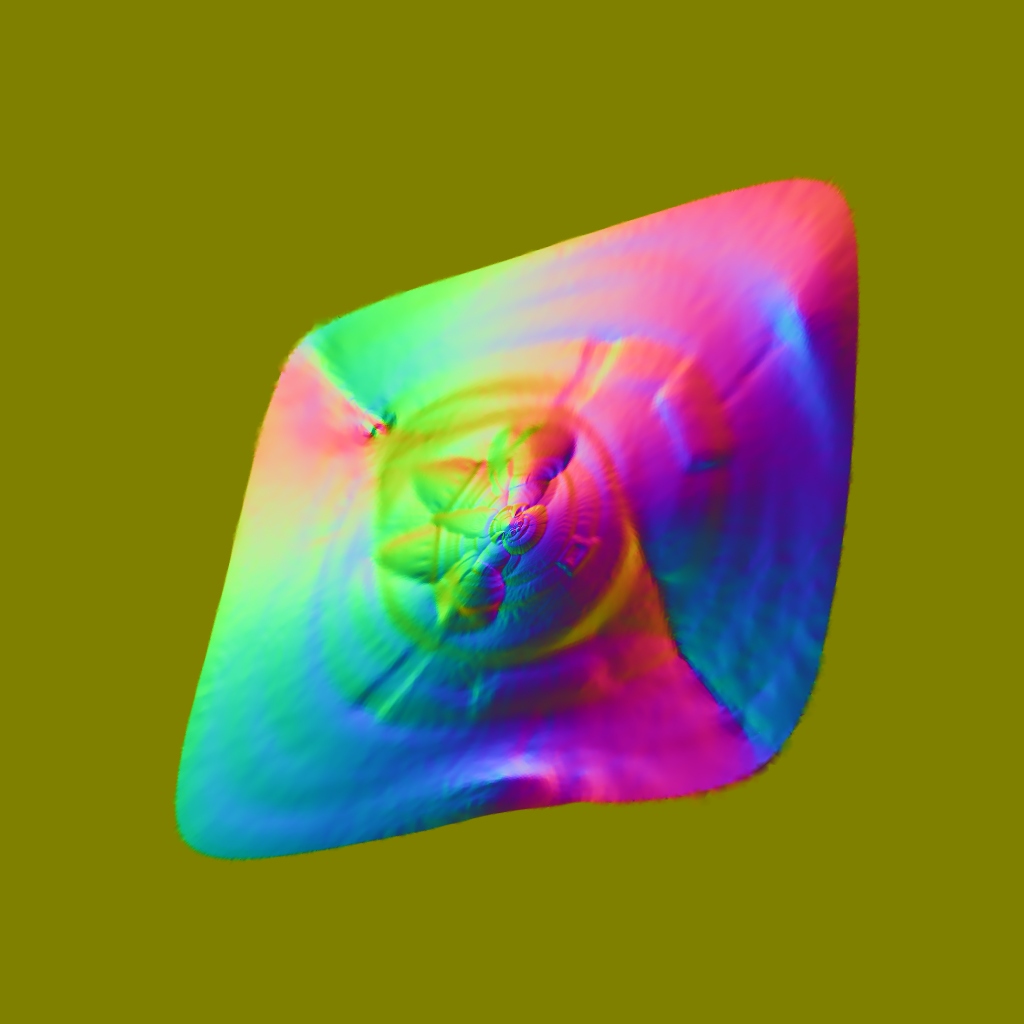} &
    \includegraphics[width=0.45\resLen]{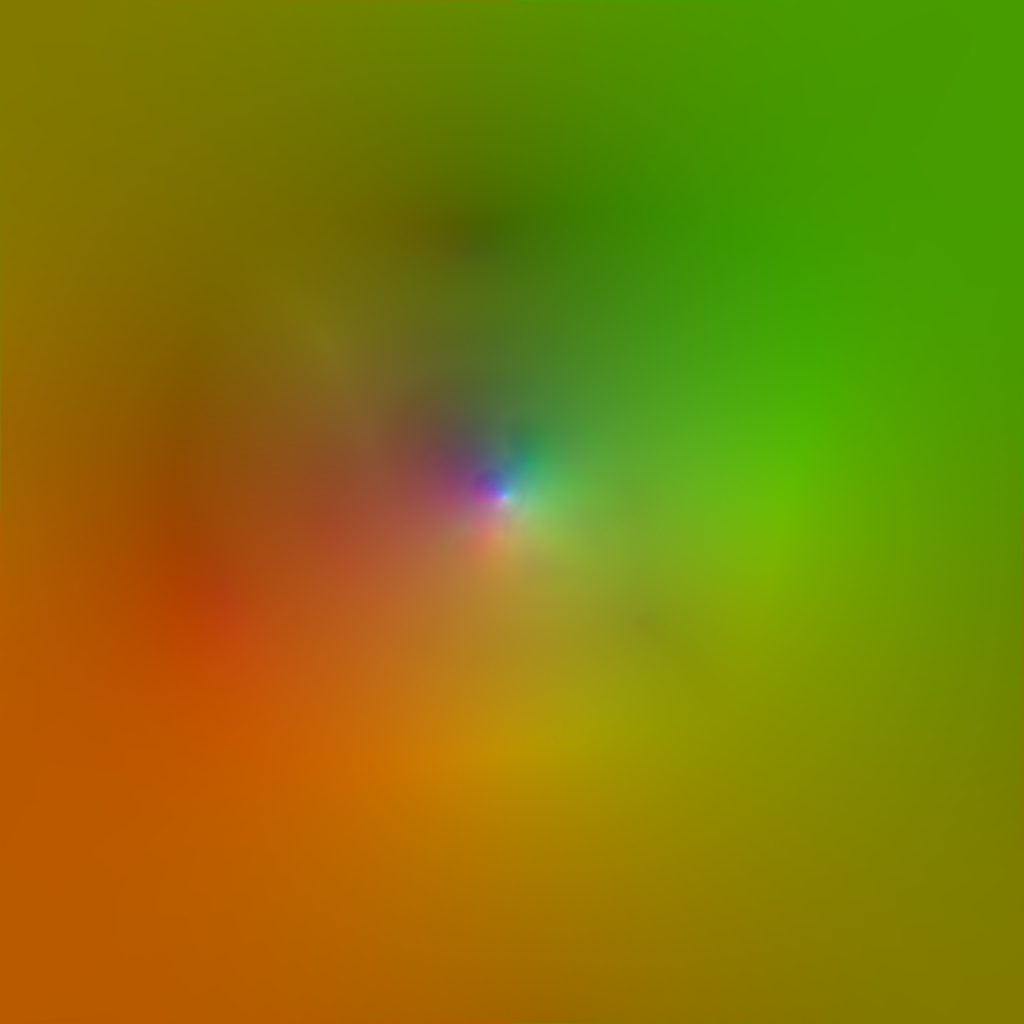} &
    \includegraphics[width=0.45\resLen]{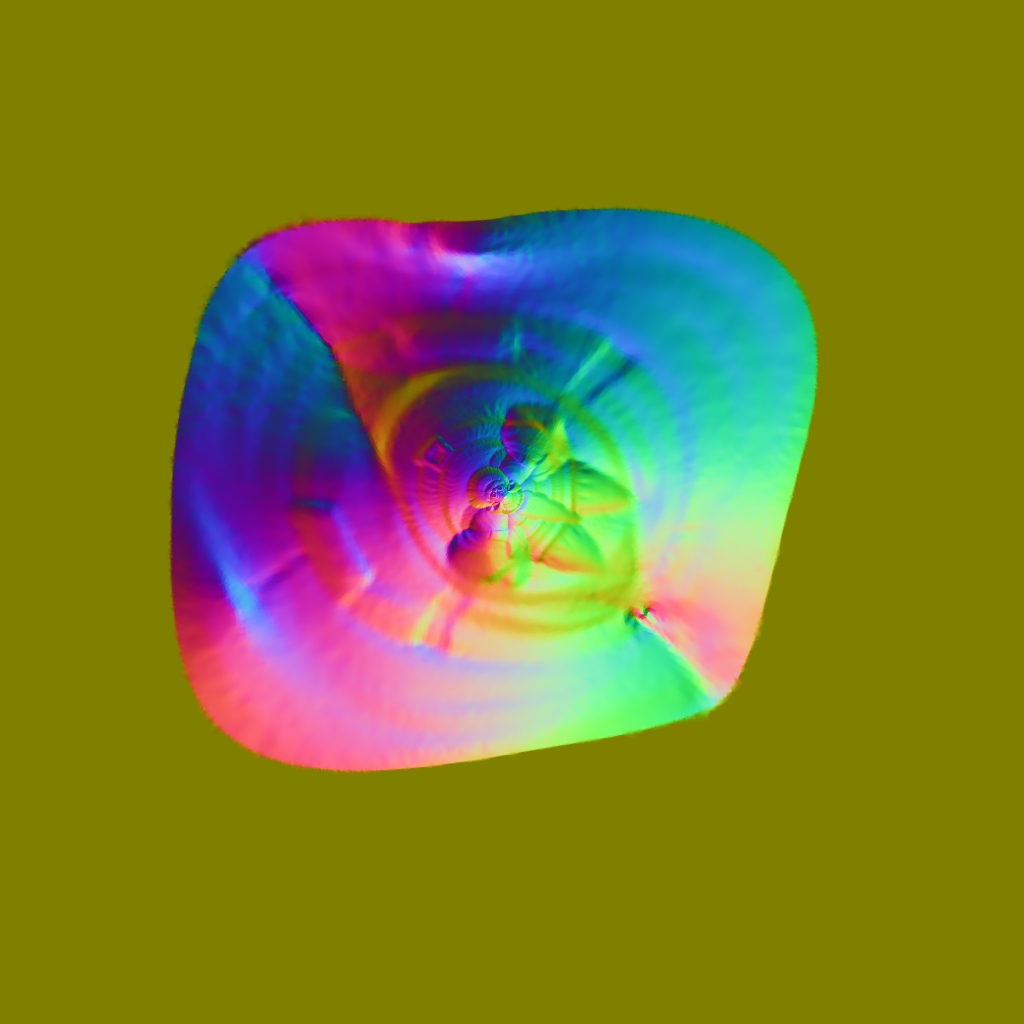} &
    \includegraphics[width=0.45\resLen]{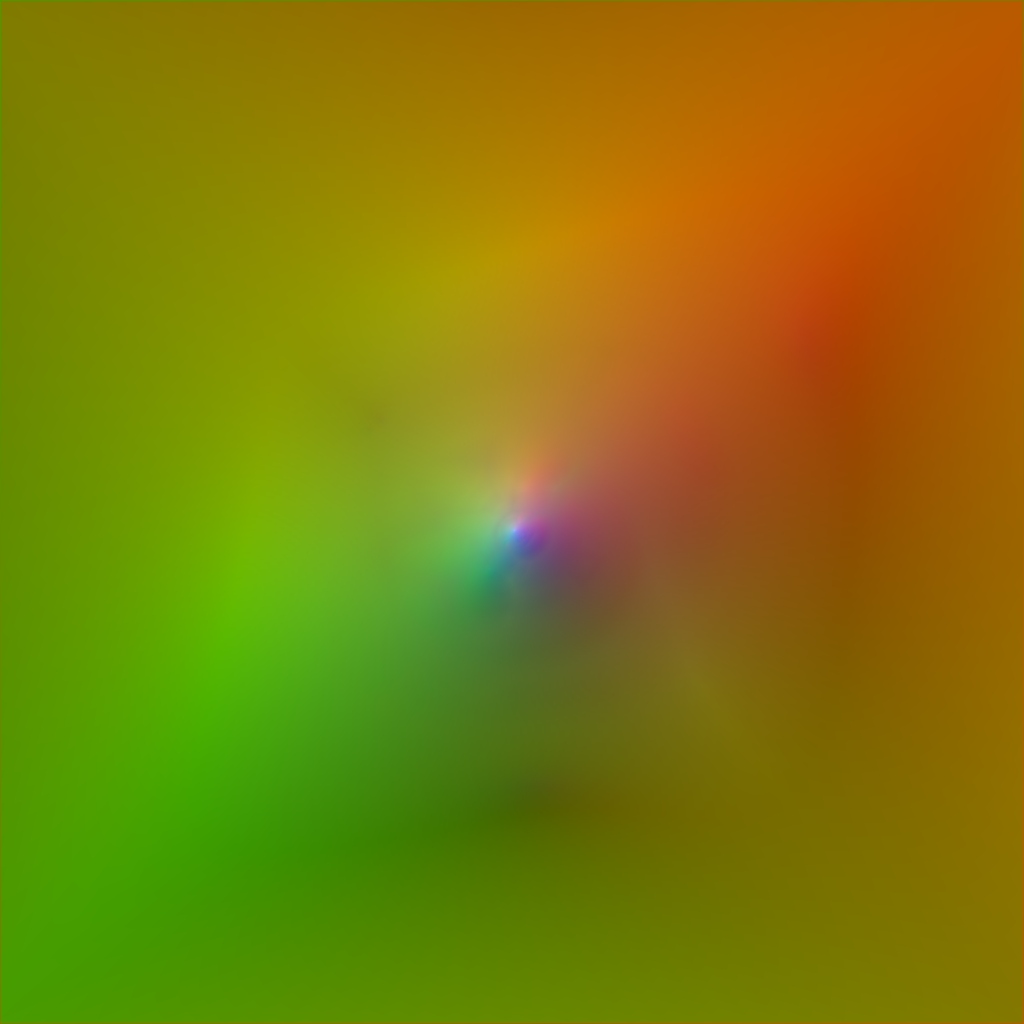} &
    \includegraphics[width=0.45\resLen]{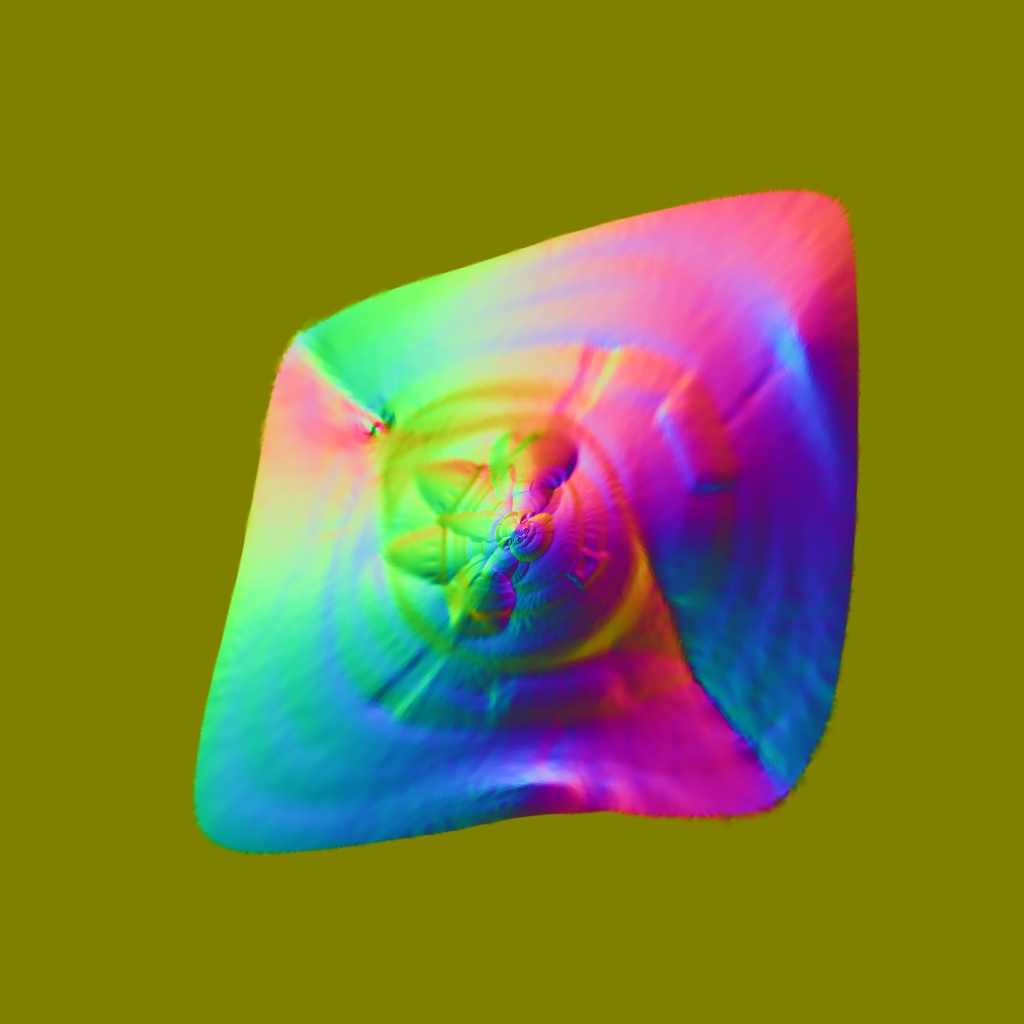} &
    \includegraphics[width=0.45\resLen]{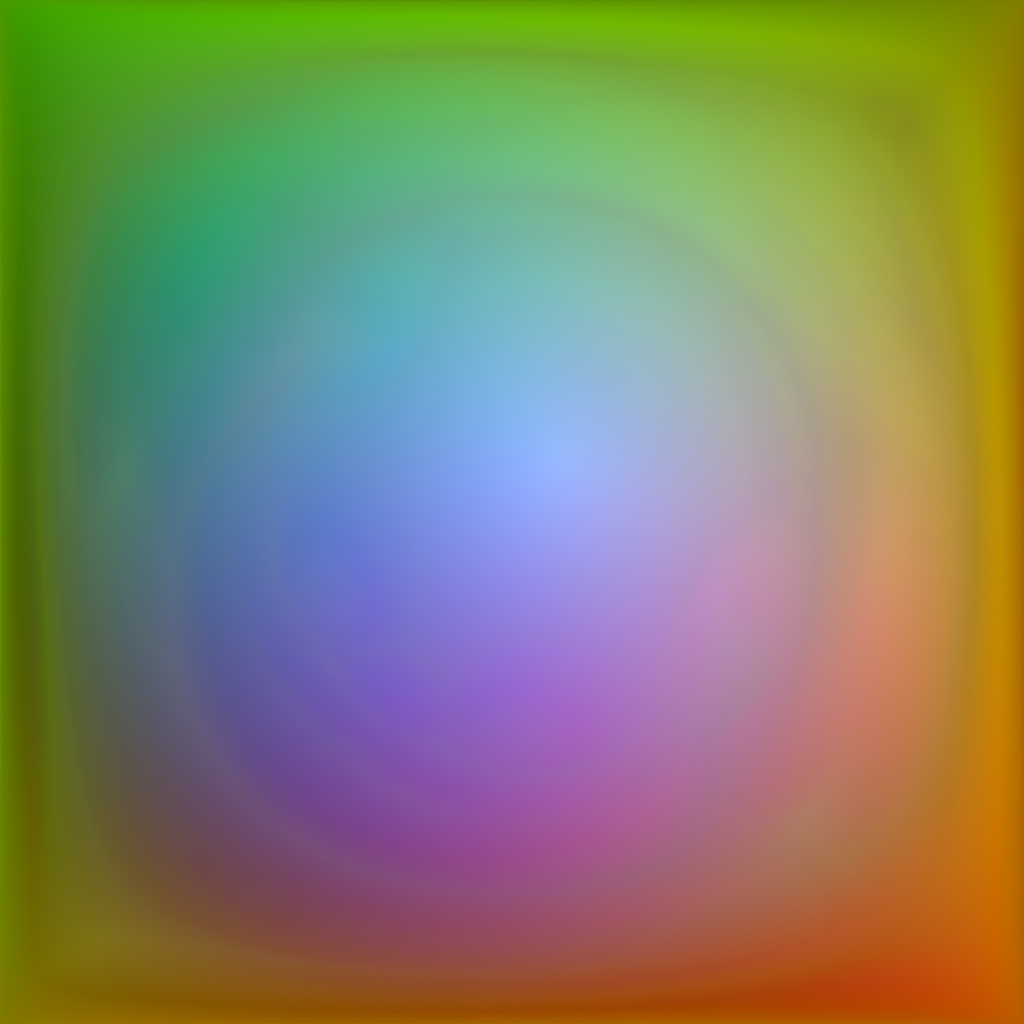} &
    \includegraphics[width=0.45\resLen]{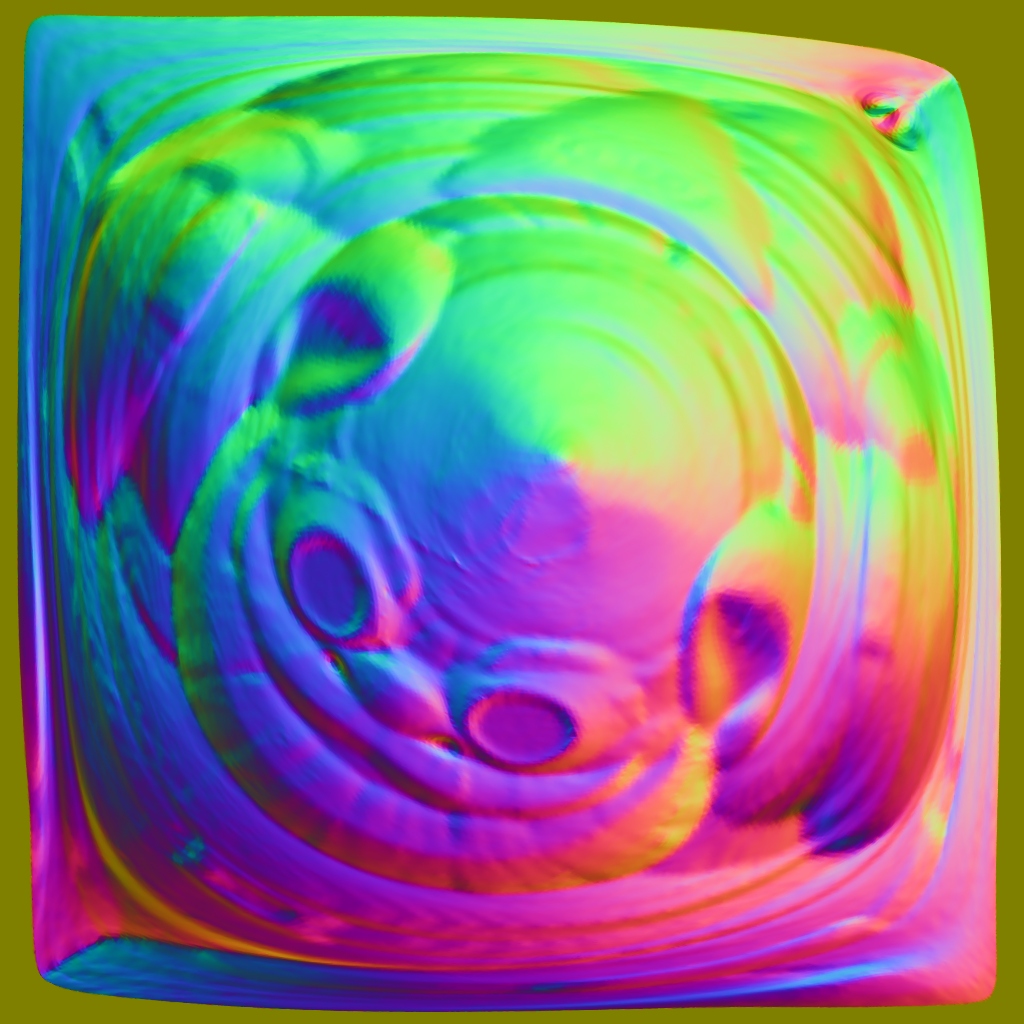} \\

    \multicolumn{2}{c}{6.5164 / 28.9256} &
    \multicolumn{2}{c}{5.9503 / 28.5869} &
    \multicolumn{2}{c}{6.7908 / 29.0235} &
    \multicolumn{2}{c}{8.0836 / 29.1703} &
    \multicolumn{2}{c}{\textbf{0.4356} / \textbf{28.2733}} \\[1pt]

    \multicolumn{2}{c}{Uniform~\cite{Tutte1963HowTD}} & 
    \multicolumn{2}{c}{Harmonic~\cite{Eck1995Harmonic}} & 
    \multicolumn{2}{c}{Ricci FLow~\cite{4483509}} & 
    \multicolumn{2}{c}{Authalic~\cite{Desbrun2002Intrinsic}} & 
    \multicolumn{2}{c}{\textbf{Ours /w OT}} \\[-10pt]
  \end{tabular}
\caption{\textbf{Qualitative comparison with state-of-the-art methods under imbalanced sampling during mesh triangulation and extraction.} 
Numbers below each method denote Chamfer Distance (CD, $\times 10^{-4}$) / Hausdorff Distance (HD, $\times 10^{-2}$), where lower values indicate better reconstruction quality. 
Our method achieves better preservation of surface details and greater robustness to sampling imbalance by applying OT to construct strictly area-preserving geometry images. 
Each 3D surface mesh is extracted from its corresponding pair of position and normal geometry images shown below it.}

  \label{fig:parameterization_compare_metrics}

\end{figure*}

\begin{table*}[t]
    \centering
    \caption{Quantitative comparison with baselines on 3D surface meshes from Thingi10k~\protect\cite{Thingi10K}. 
    We indicate whether each method is area-preserving, angle-preserving, or mixed. 
    Our method achieves strict area preservation via OT’s measure-preserving property, enabling continuous LoD representation without requiring multiple decoders through our neural geometry image-based representation.} 

    \vspace{1mm}
    \begin{tabular}{|c|c|c|c|c|c|}
        \hline
        \textbf{Method} & 
        \begin{tabular}[c]{@{}c@{}} \textbf{Chamfer Dist (CD) $\downarrow$} \\ ($\times 10^{-4}$) \end{tabular} & 
        \begin{tabular}[c]{@{}c@{}} \textbf{Hausdorff Dist (HD) $\downarrow$} \\ ($\times 10^{-2}$) \end{tabular} & 
        \textbf{Area-Preserving} & 
        \textbf{Angle-Preserving} & 
        \textbf{Mixed} \\
        \hline
        Uniform~\cite{Tutte1963HowTD}     &  11.9534  &  19.8963 & \xmark & \xmark & \xmark \\
        Harmonic~\cite{Eck1995Harmonic}   &  12.1267  &  19.4869 & \xmark & \xmark & \xmark \\
        Ricci Flow~\cite{4483509}         &  13.5256  &  20.5909 & \xmark & \cmark & \xmark \\
        Authalic~\cite{Desbrun2002Intrinsic} &  13.0667  &  20.2105 & \xmark & \xmark & \cmark \\
        \hline
        \textbf{Ours w/ OT}                     &  \textbf{1.5476}  &  \textbf{18.1094} & \cmark  & \xmark & \xmark \\
        \hline
    \end{tabular}
    \label{tab:parameterization_metrics}
    \vspace{-5mm}
\end{table*}

\vspace{-4mm}
\paragraph{Area Preservation via Optimal Transport.}  
To correct the area distortion from conformal maps, we apply Optimal Transport (OT) refinement. The OT formulation ensures that each surface patch in 3D is assigned a proportional pixel area in 2D, eliminating oversampling in flat regions and undersampling in curved ones. The goal is to find a measure-preserving map $T:\Omega \to P$ minimizing
\begin{equation}\label{formula:ot_cost}
    \int_\Omega \|x - T(x)\|^2 \mu(x)\, dx.
\end{equation}
Following Brenier’s theorem~\cite{brenier1991polar}, the solution is the gradient of a convex potential $u_{\mathbf{h}}$ parameterized by a height vector $\mathbf{h}=(h_1,\dots,h_k)$. The corresponding variational energy~\cite{gu2016variational} is
\begin{equation}\label{formula:otenergy}
    E(\mathbf{h})=\int_\Omega u_{\mathbf{h}}(x)\mu(x)\, dx-\sum_{i=1}^k \nu_i h_i,
\end{equation}
with gradient and Hessian
\begin{align}\label{ot:grad}
    \nabla E(\mathbf{h}) &= (w_1(\mathbf{h})-\nu_1,\dots,w_k(\mathbf{h})-\nu_k)^T, \\
    \frac{\partial^2 E}{\partial h_i \partial h_j} &=
    \begin{cases}
      \tfrac{1}{\|p_j-p_i\|}\int_{e_{ij}} \mu, & e_{ij}=W_i\cap W_j \cap \Omega, \\
      0, & \text{otherwise}.
    \end{cases}\label{ot:hessian}
\end{align}
Since $E(\mathbf{h})$ is strictly convex, Newton’s method converges to a unique minimizer $\mathbf{h}_0$, and the induced gradient map $\nabla u_{\mathbf{h}_0}$ yields the final area-preserving parameterization. The overall mapping is expressed as $\psi^{-1} \circ \phi$, where $\phi$ is the conformal initialization and $\psi$ the OT correction.

\subsection{Neural Geometry Images Network Training}
We train the network end-to-end in geometry image-space using reconstruction losses that aim to minimize pixel-wise errors while preserving local structural similarity. The architecture is composed of standard \textit{ResNet blocks}~\cite{Kaiming2006ResNet} for feature extraction and \textit{PixelShuffle layers}~\cite{Shi2016PixelShuffle} for upsampling, which are widely used in image restoration tasks.

\vspace{-5mm}
\subsubsection{Reconstruction Losses}  
We combine a pixel-wise $\ell_1$ loss with a structural loss based on Multi-Scale Structural Similarity (MS-SSIM)~\cite{Wang2003MS_SSIM}, to balance low-level accuracy and perceptual quality. The $\ell_1$ loss is defined as:
\begin{equation}\label{loss:l1}
\mathcal{L}_{\ell_1} = \frac{1}{N} \sum_{i=1}^{N} \| \widetilde{I}_i - I_i \|_1, 
\end{equation}
where \( \widetilde{I}_i \) and \( I_i \) denote the predicted and ground-truth geometry images, respectively. The structural loss is defined as:
\begin{equation}\label{loss:struct}
\mathcal{L}_{\text{struct}} = \frac{1}{N} \sum_{i=1}^{N} \left(1 - \text{MS-SSIM}(\widetilde{I}_i, I_i)\right), 
\end{equation}
where MS-SSIM~\cite{Wang2003MS_SSIM} measures structural similarity across multiple spatial scales using local statistics such as luminance, contrast, and structure. Specifically, it is computed as:
\begin{equation}\label{loss:msssim}
\text{MS-SSIM}(x, y) = [l_M(x, y)]^{\alpha_M} \cdot \prod_{j=1}^{M} [c_j(x, y)]^{\beta_j} [s_j(x, y)]^{\gamma_j}, 
\end{equation}
where \( l_M, c_j, s_j \) denote the luminance, contrast, and structure comparisons at scale \( j \), and \( \alpha_M, \beta_j, \gamma_j \) are fixed weights. The total reconstruction loss is given by:
\begin{equation}\label{loss:total}
\mathcal{L}_{\text{recon}} = \eta_{1} \mathcal{L}_{\ell_1} + \eta_{2} \mathcal{L}_{\text{struct}}, 
\end{equation}
With $\eta_{1} = 0.5$, $\eta_{2} = 0.5$, and optimization performed using the Adam optimizer with a learning rate $lr = 5 \times 10^{-5}$. These weights balance pixel accuracy ($\ell_1$) and structural similarity (MS-SSIM), reducing reconstruction errors while preserving both local and global details.

\vspace{-3mm}
\subsubsection{Input Precondition}
Given an input triangle mesh, we first perform adaptive isotropic remeshing~\cite{Jacques2020Remeshing} to obtain a dense, high-quality mesh and cut it open. We then compute a unique area-preserving parametrization using our proposed Optimal Mass Transport (OMT)-based method, \cref{algorithm:ot}, to obtain a bijective mapping to the unit square. This UV map is used to render vertex-based positions into a single $1024 \times 1024$ geometry image, stored as a 16-bit RGB PNG. All values are normalized to the range $[0,1]$, and boundary pixels are filled via nearest-boundary extrapolation to ensure continuity at seams. Mipmaps are generated on the GPU down to $1 \times 1$, and a specific lower-resolution level from the hierarchy is chosen as the network input based on the desired geometry image compression ratio (CR).

\begin{figure*}[htbh]
    \centering
    \setlength{\resLen}{0.185\textwidth}
    \begin{tabular}{ccccc}
        \includegraphics[trim={450 0 200 0},clip,width=\resLen]{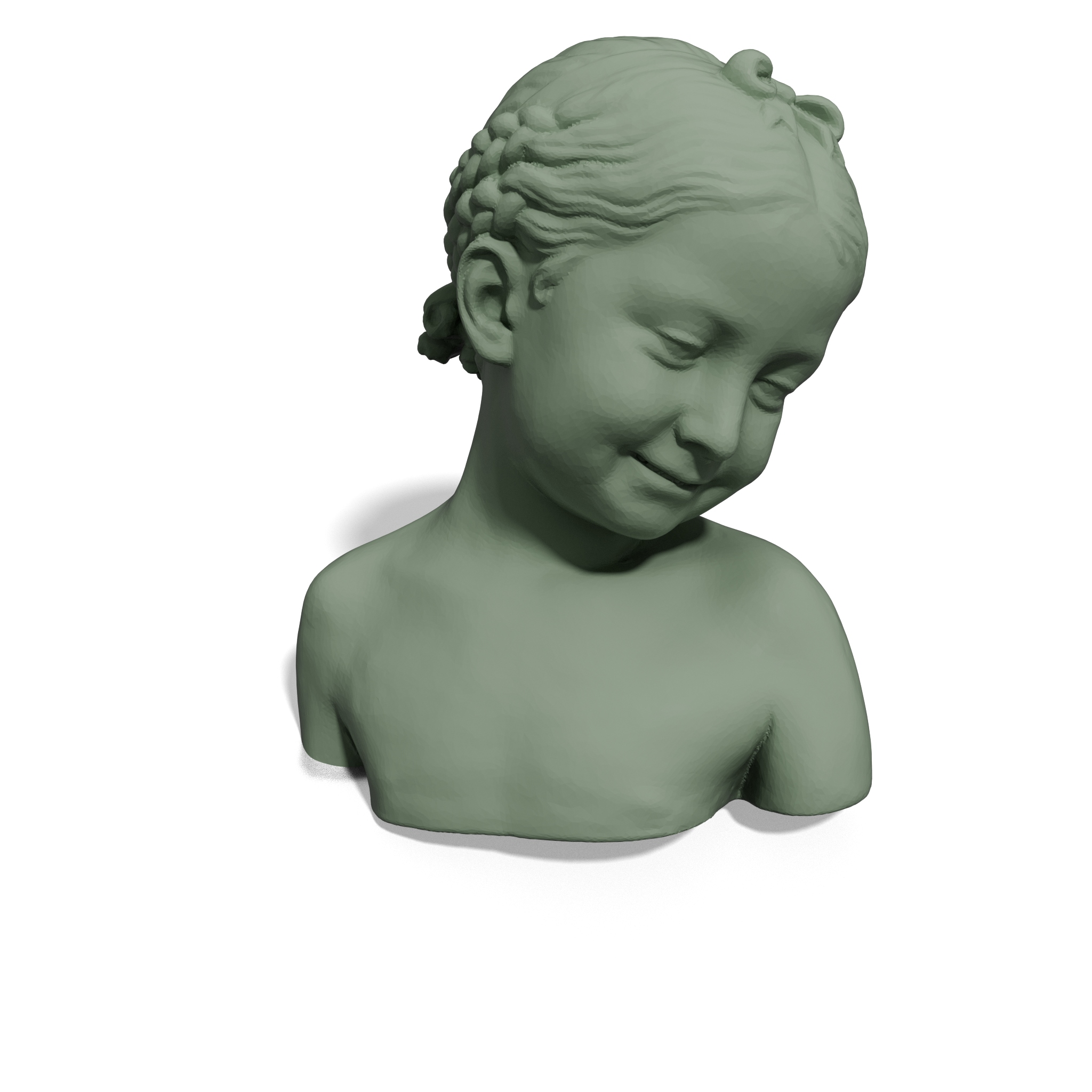} &
        \includegraphics[trim={450 0 200 0},clip,width=\resLen]{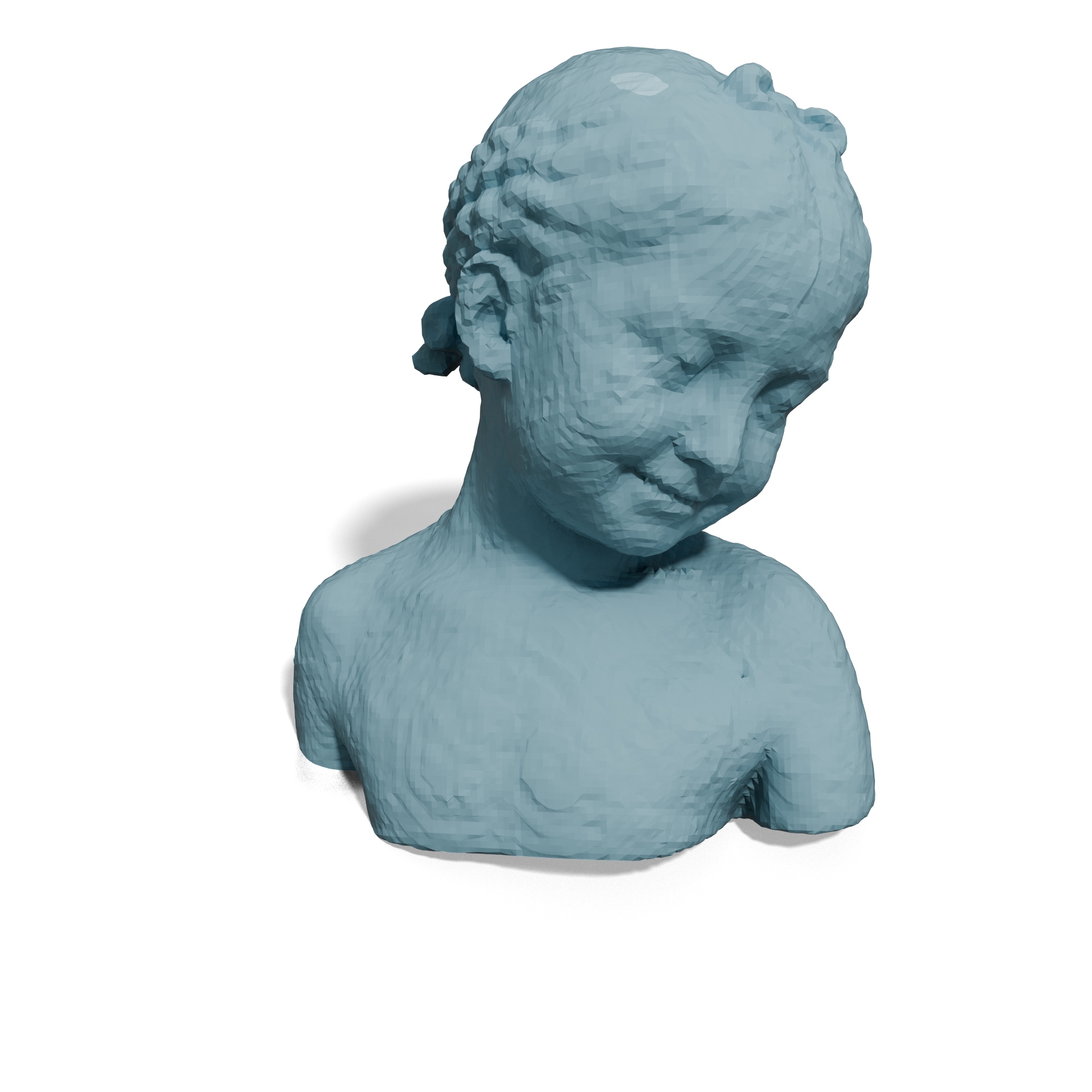} &
        \includegraphics[trim={450 0 200 0},clip,width=\resLen]{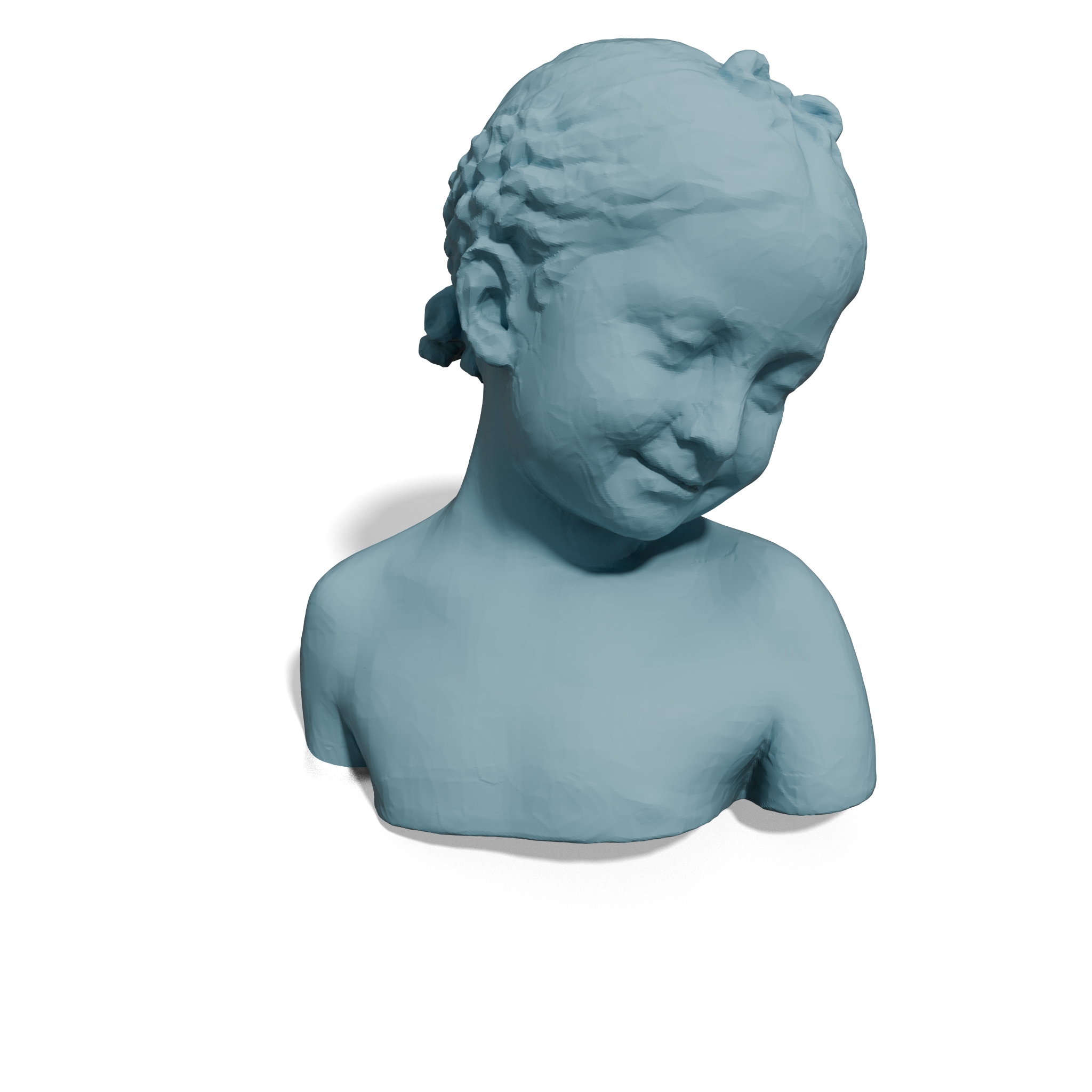} &
        \includegraphics[trim={450 0 200 0},clip,width=\resLen]{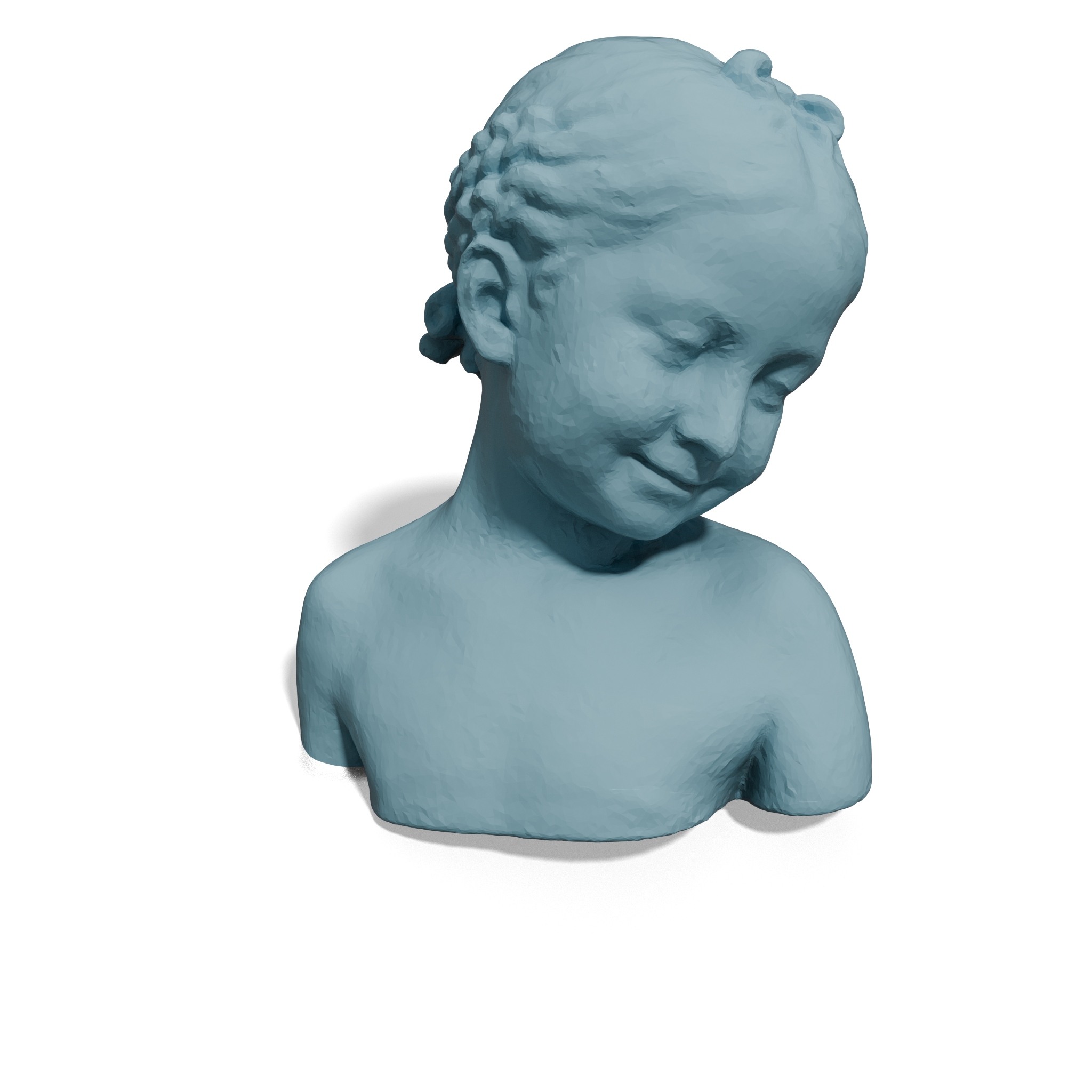} &
        \includegraphics[trim={450 0 200 0},clip,width=\resLen]{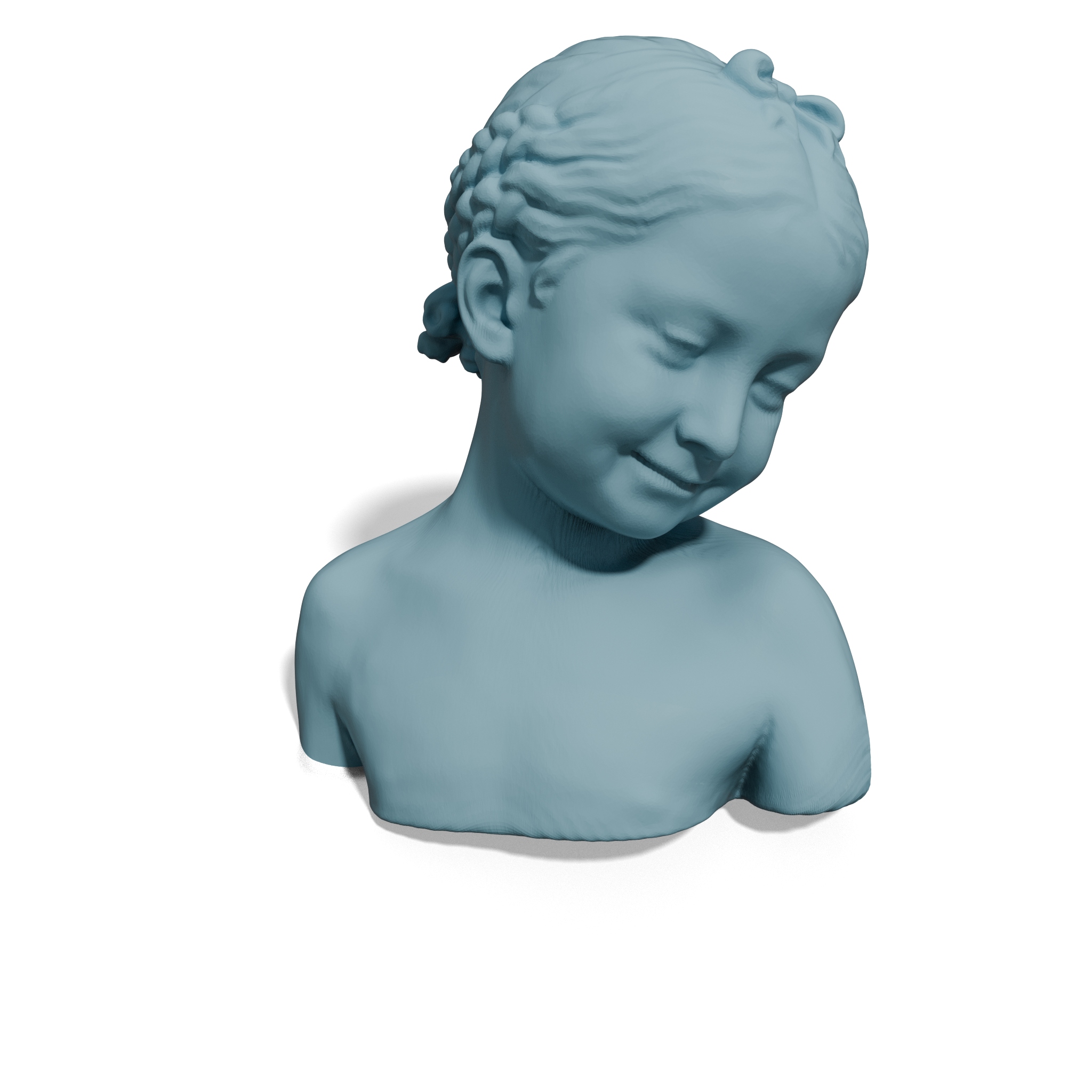} \\[-30pt] 

        \small CD ($\times10^{-4}$) / HD ($\times10^{-2}$) &
        \small 1.279 / 4.651 &
        \small 0.495 / 2.472 &
        \small 0.078 / 1.529 &
        \small 0.057 / 1.474 \\[5pt]

        Ground Truth &
        ACORN~\cite{martel2021acorn} &
        NGLOD~\cite{takikawa2021nglod} &
        NCS~\cite{Morreale2022NCS} &
        Ours
    \end{tabular}
    \caption{\textbf{Visual comparisons with neural overfitting methods.} 
    Our method achieves superior reconstruction quality in terms of Chamfer Distance (CD) and Hausdorff Distance (HD), 
    while maintaining a higher compression ratio (CR = 64).}

    \label{fig:neural_overfitting}
    \vspace{-3mm}
\end{figure*}

\begin{figure*}[htbh]
  \centering
  \setlength{\tabcolsep}{0pt}      % no intercolumn padding
  \renewcommand{\arraystretch}{0.9}
  \setlength{\resLen}{0.153\textwidth}  % 6×0.153 = 0.918 < 1.0

  \begin{tabular}{@{}cccccc@{}}
      % --- Row 1: Genome ---
      \includegraphics[trim={400 0 300 0},clip,width=\resLen]{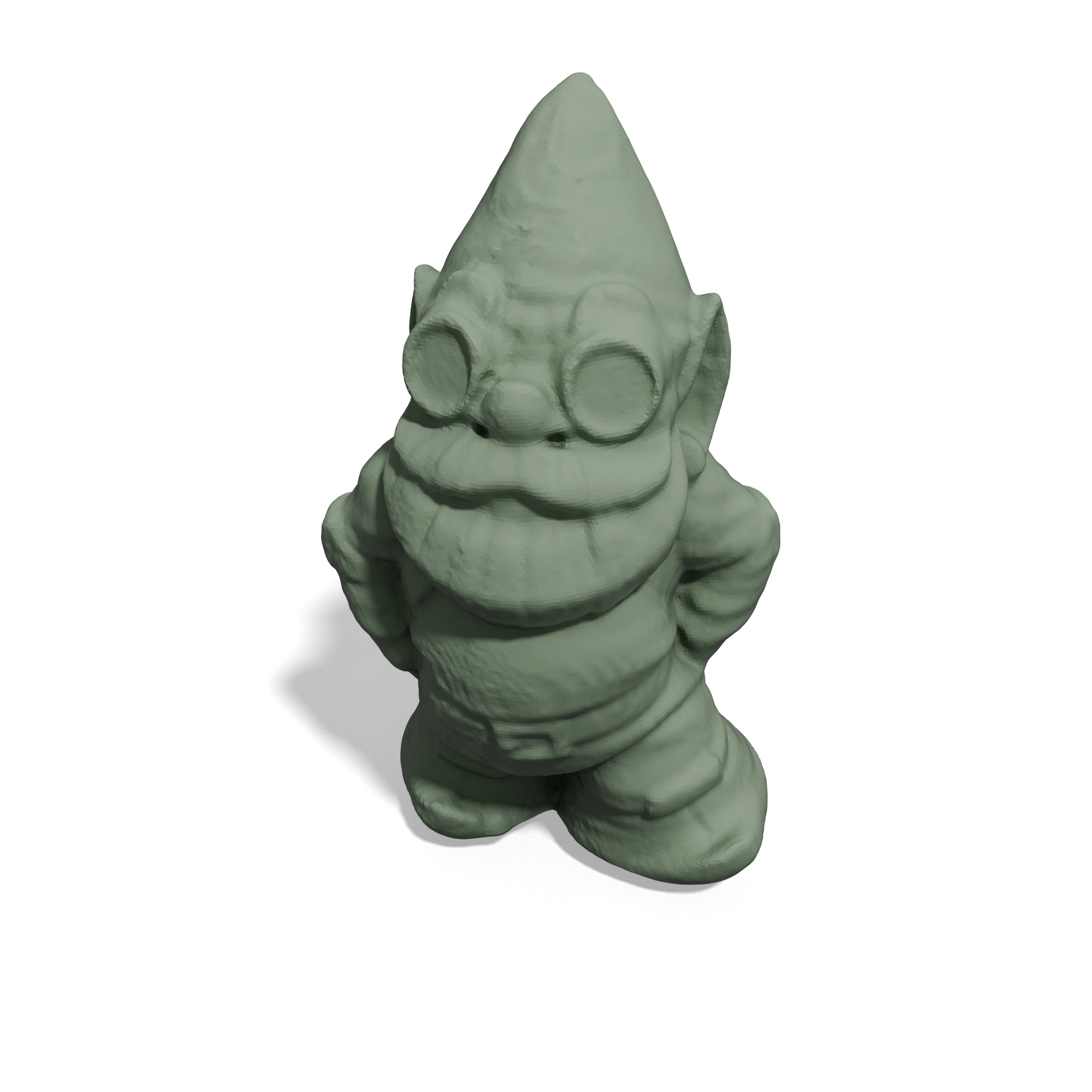} &
      \includegraphics[trim={400 0 300 0},clip,width=\resLen]{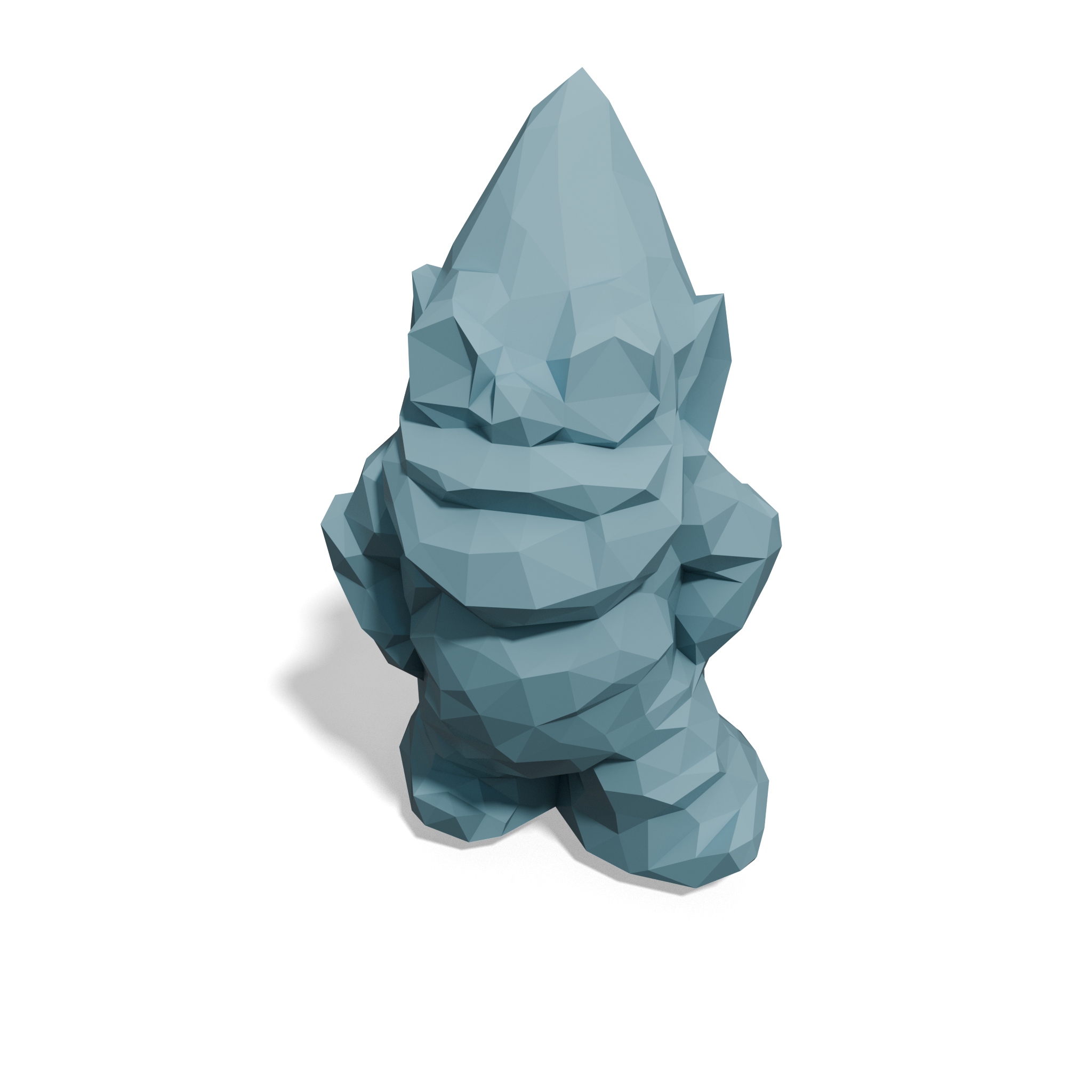} &
      \includegraphics[trim={400 0 300 0},clip,width=\resLen]{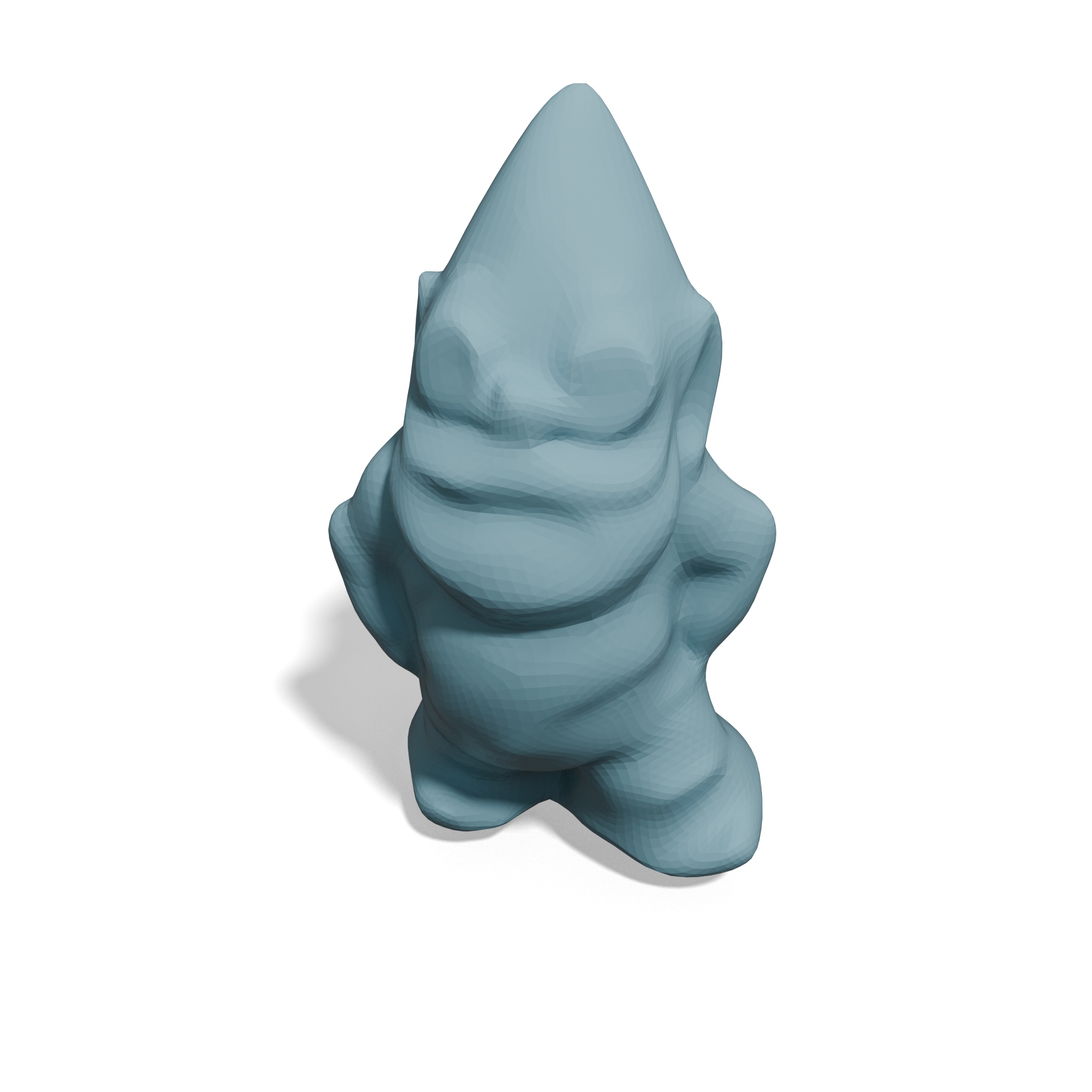} &
      \includegraphics[trim={400 0 300 0},clip,width=\resLen]{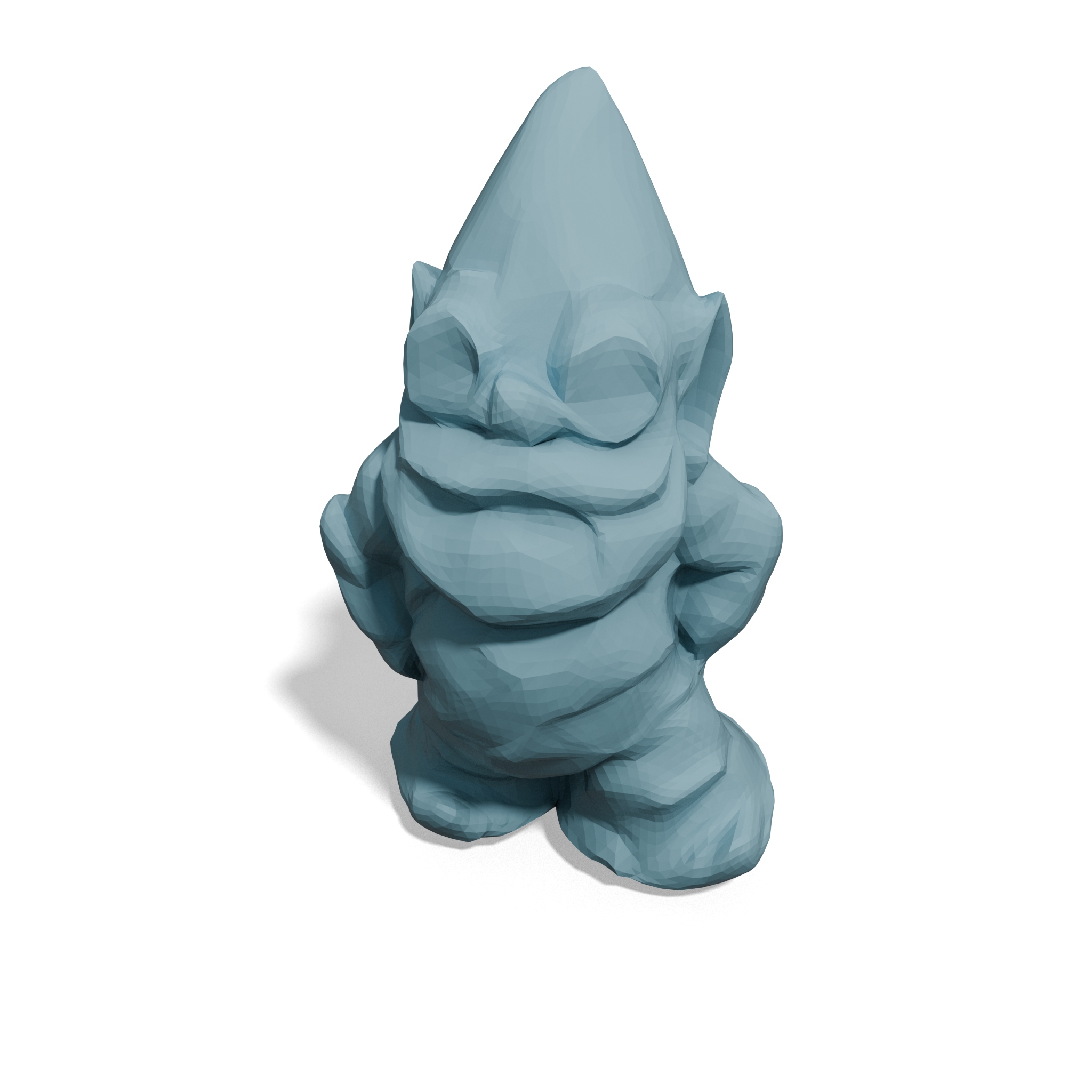} &
      \includegraphics[trim={400 0 300 0},clip,width=\resLen]{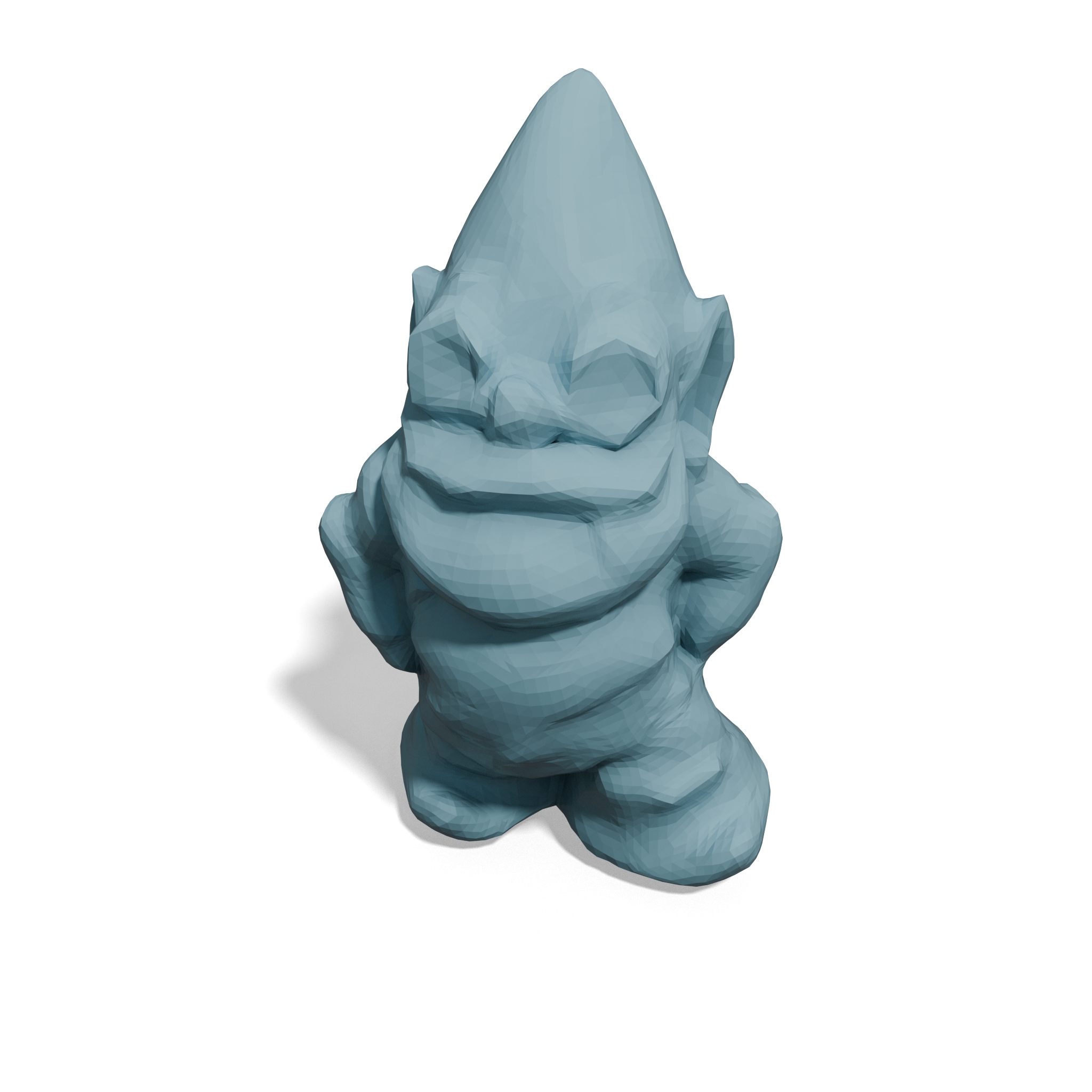} &
      \includegraphics[trim={400 0 300 0},clip,width=\resLen]{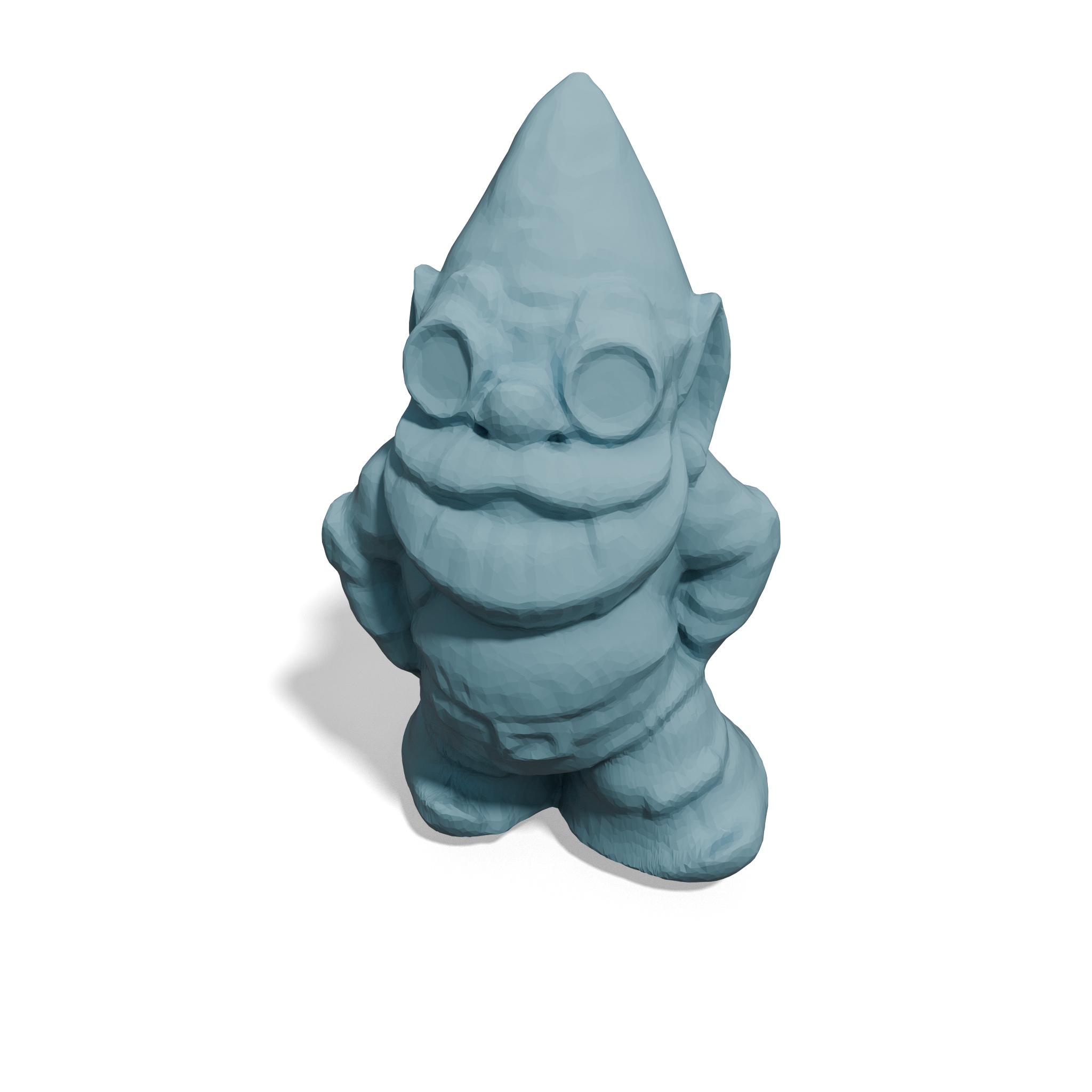}\\[-6pt]

      \small CD ($\times10^{-4}$) / HD ($\times10^{-2}$) &
      \small 4.5383 / 4.5468 &
      \small 5.1749 / 5.4676 &
      \small 5.4839 / 4.9798 &
      \small 5.0458 / 4.7055 &
      \small 4.2449 / 4.6707 \\[4pt]

      % --- Row 2: Building ---
      \includegraphics[width=\resLen]{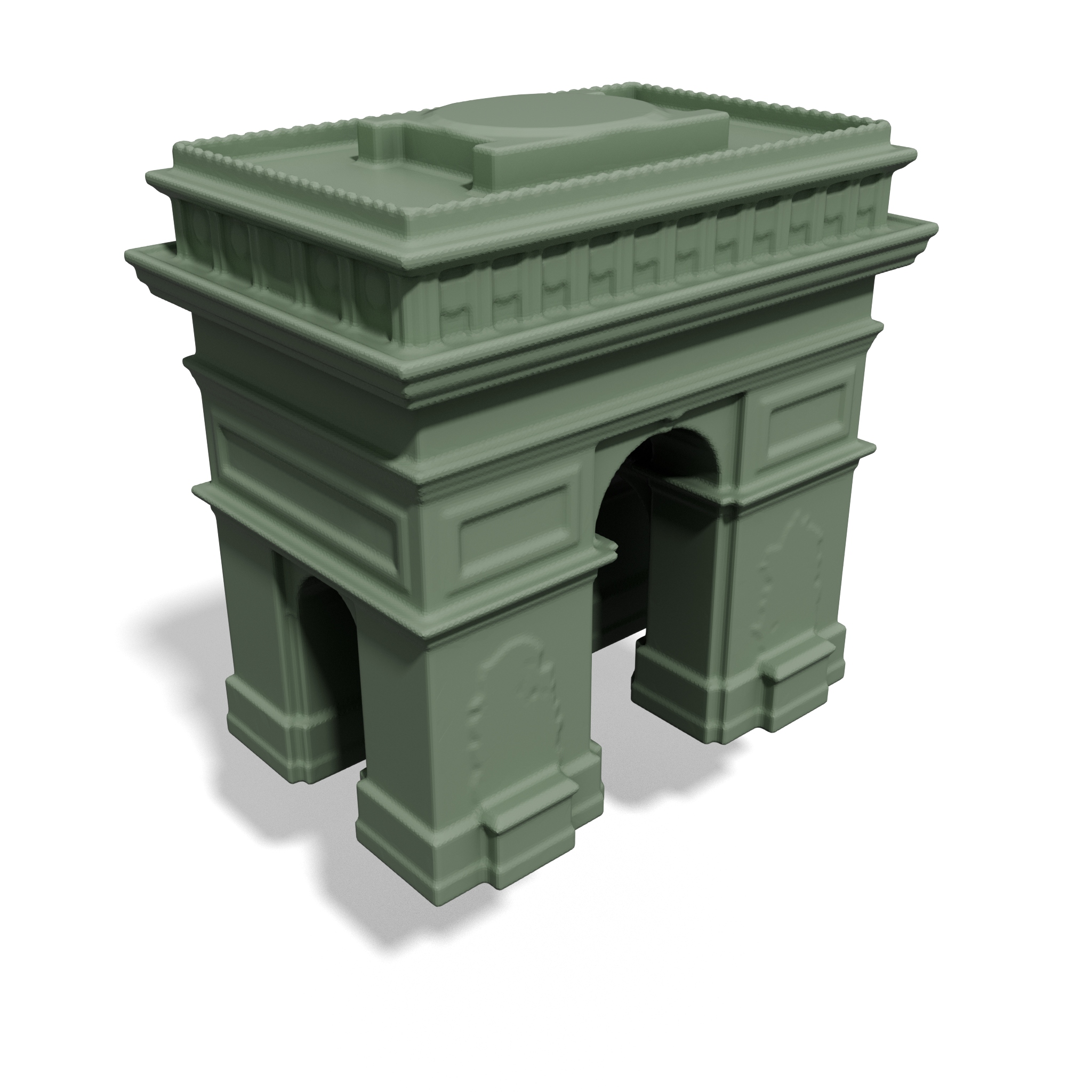} &
      \includegraphics[width=\resLen]{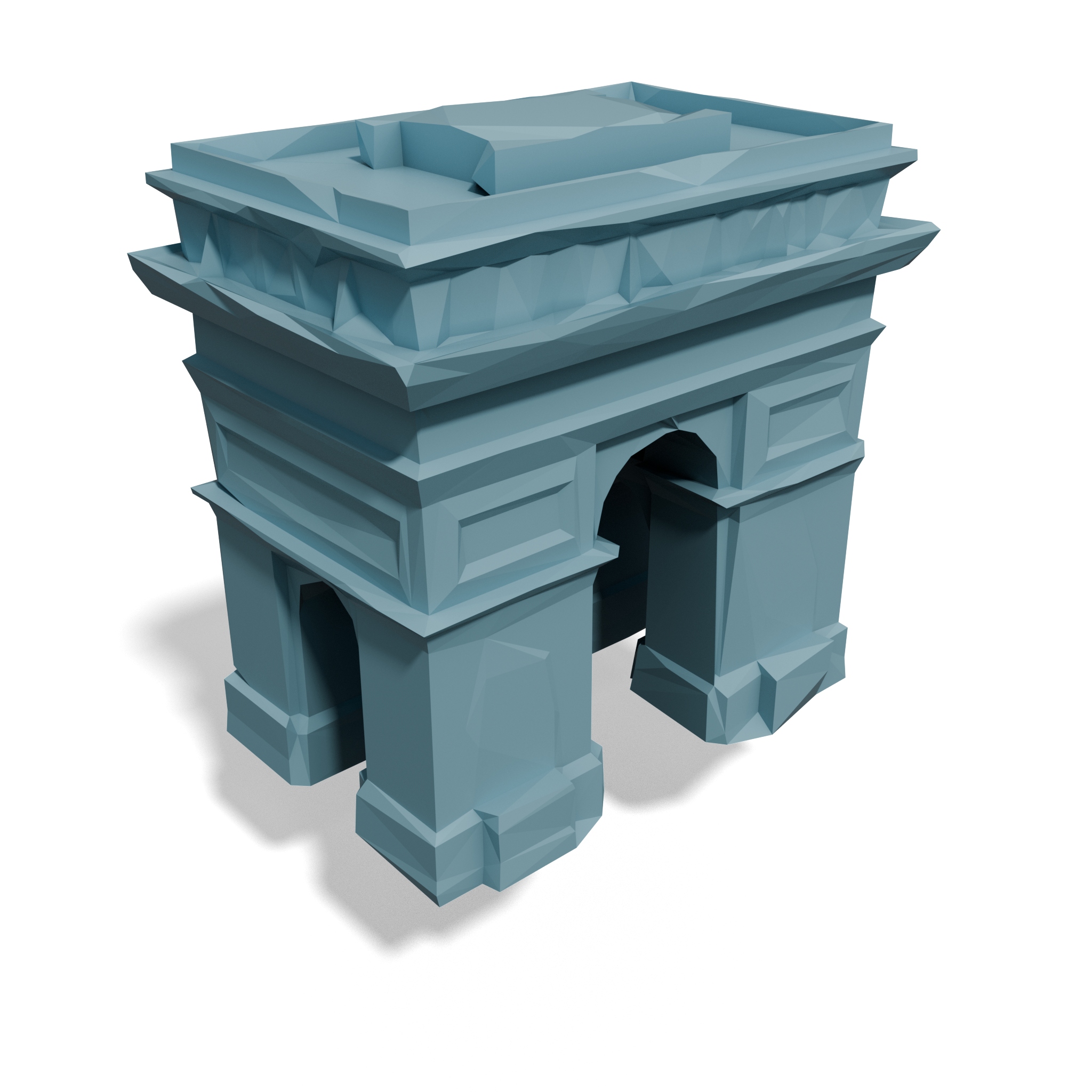} &
      \includegraphics[width=\resLen]{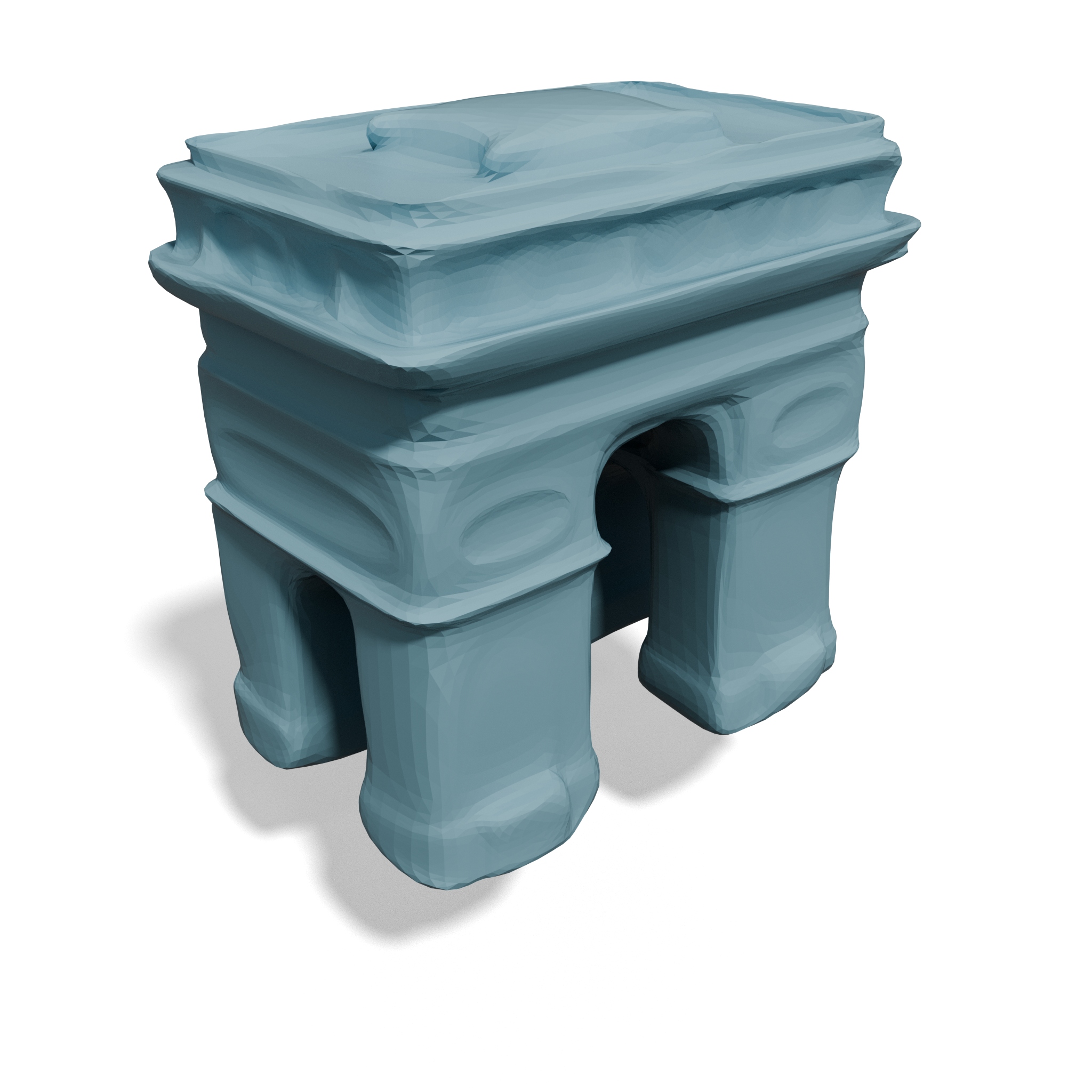} &
      \includegraphics[width=\resLen]{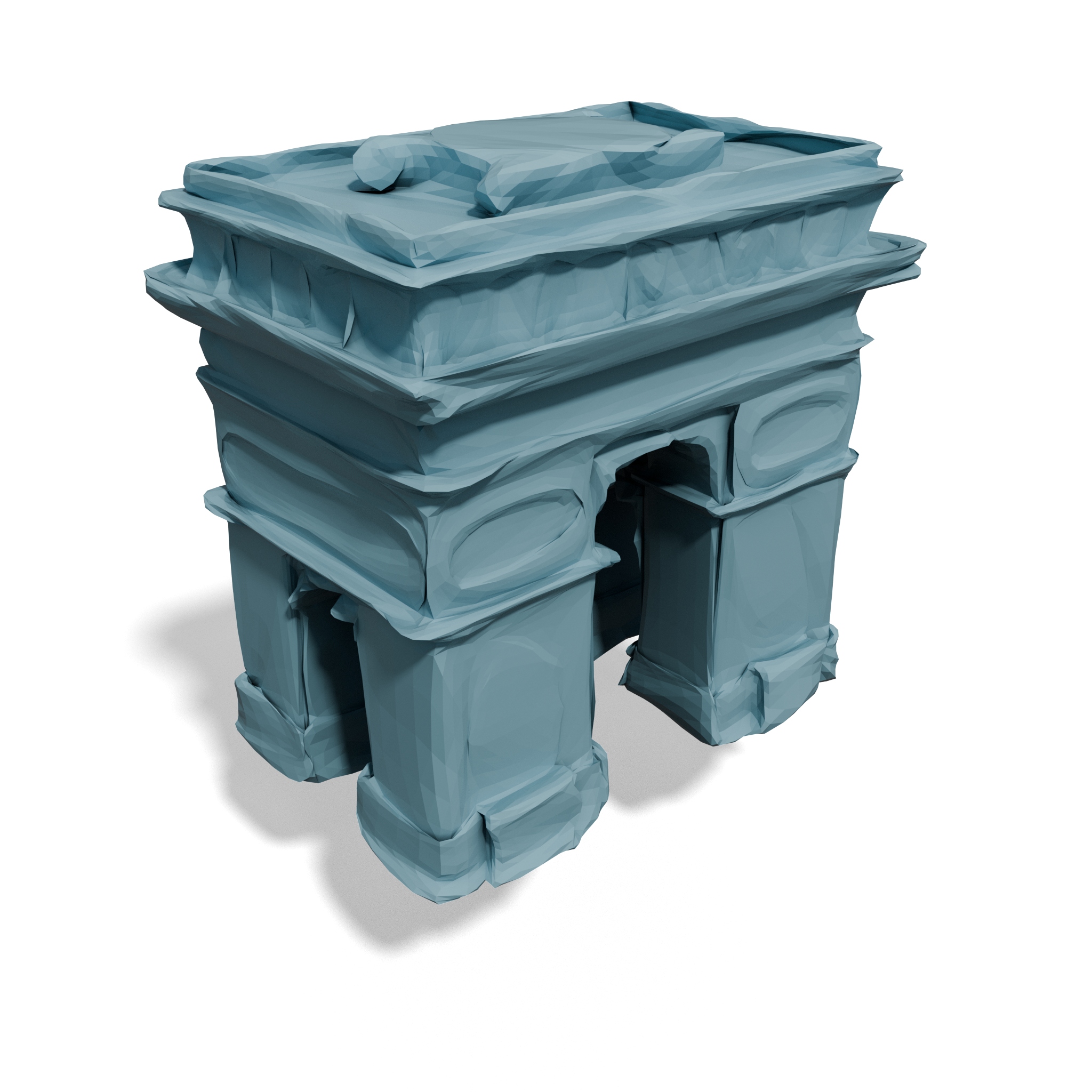} &
      \includegraphics[width=\resLen]{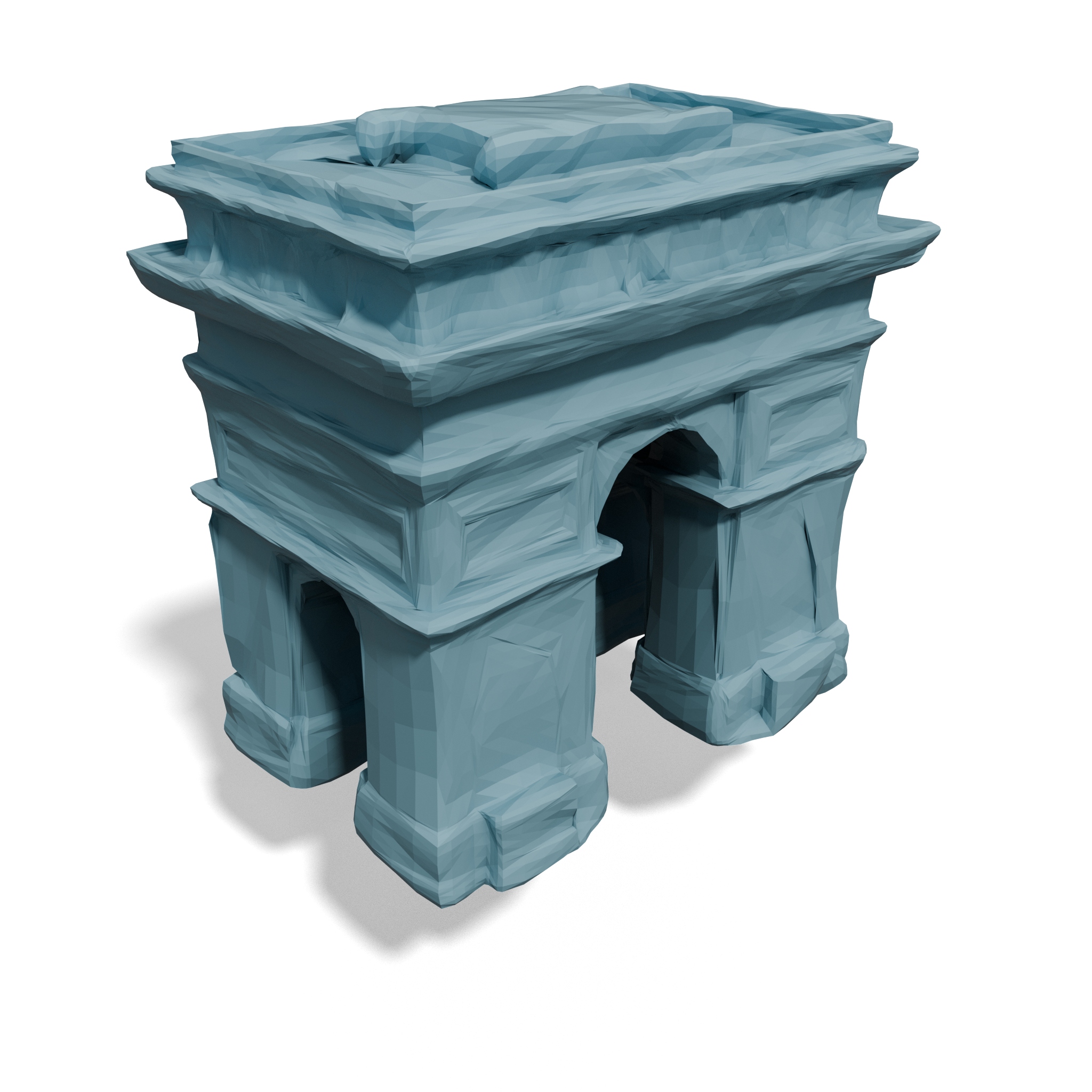} &
      \includegraphics[width=\resLen]{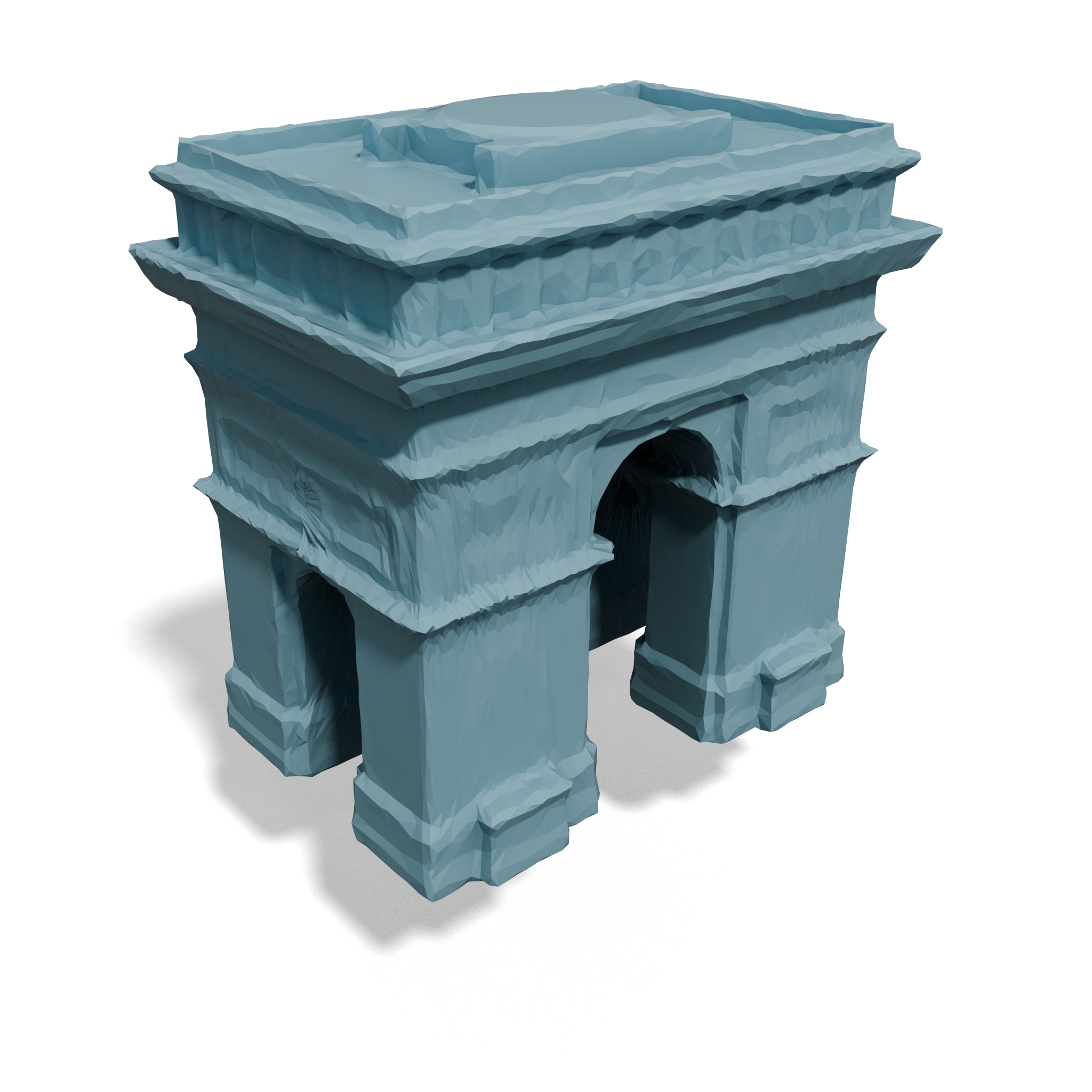}\\[-6pt]

      \small CD ($\times10^{-4}$) / HD ($\times10^{-2}$) &
      \small 12.4223 / 7.7670 &
      \small 14.0599 / 8.2054 &
      \small 18.6925 / 7.8472 &
      \small 14.0140 / 8.3563 &
      \small 12.3844 / 7.4082 \\[4pt]

      % --- Row 3: Colonel ---
      \includegraphics[trim={350 0 100 0},clip,width=\resLen]{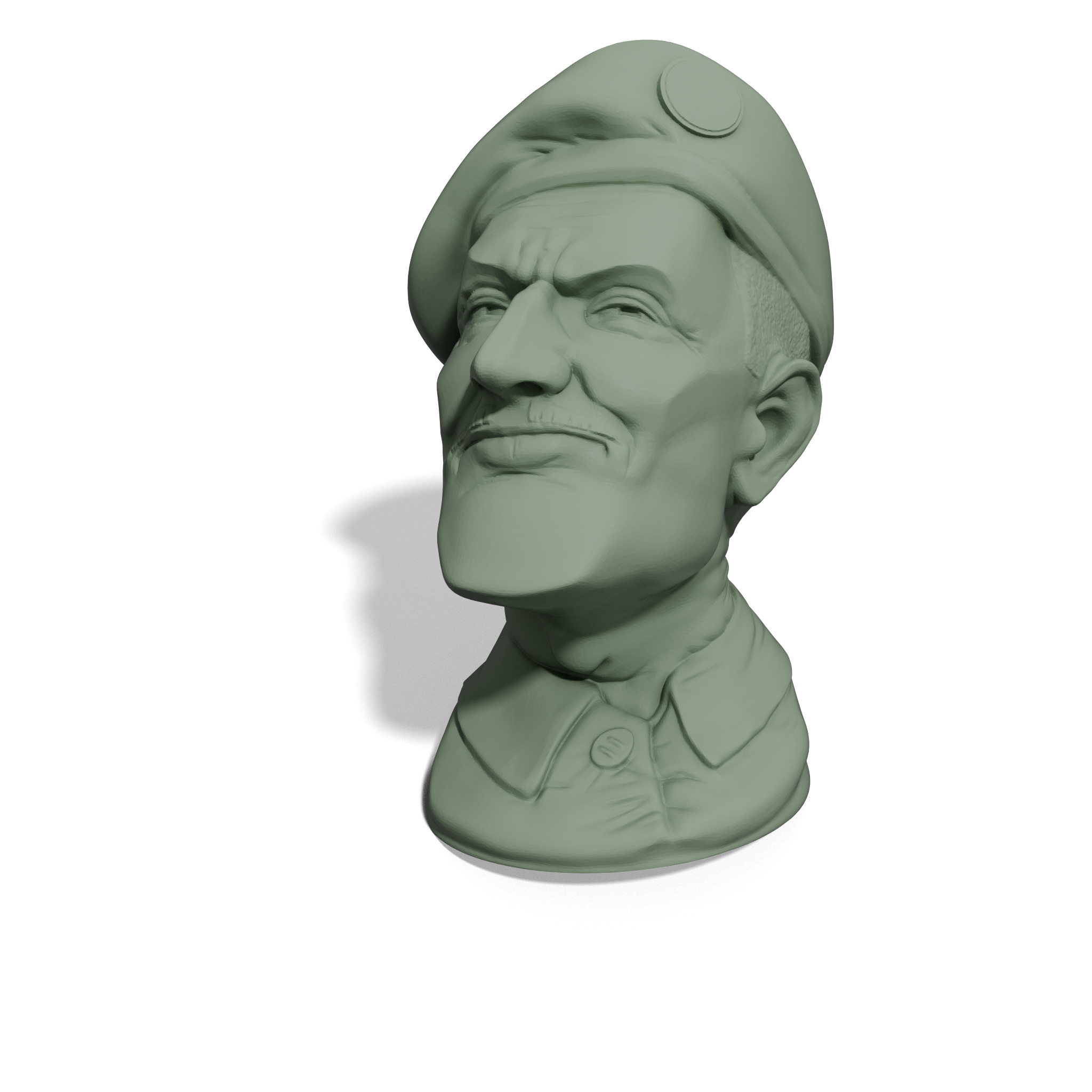} &
      \includegraphics[trim={350 0 100 0},clip,width=\resLen]{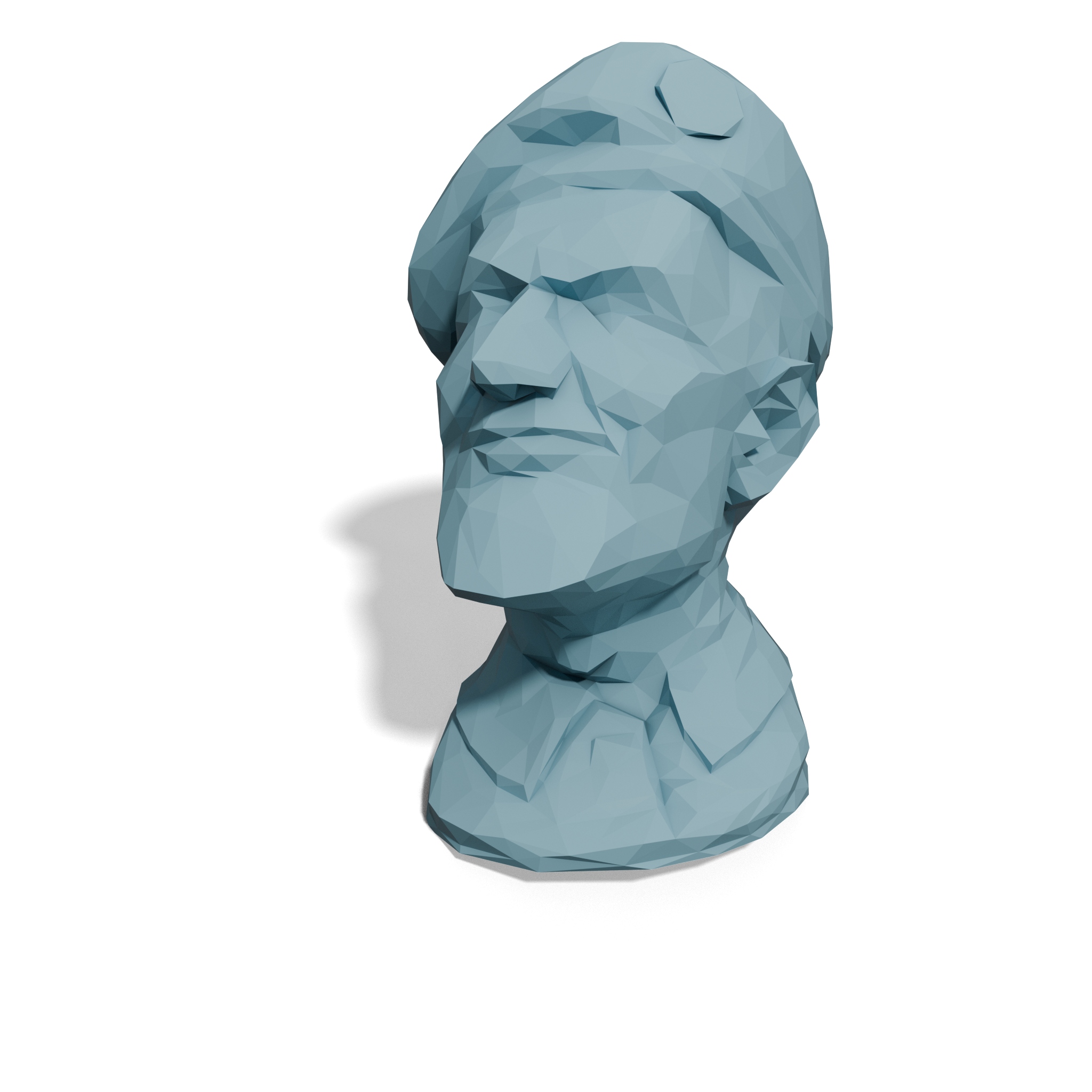} &
      \includegraphics[trim={350 0 100 0},clip,width=\resLen]{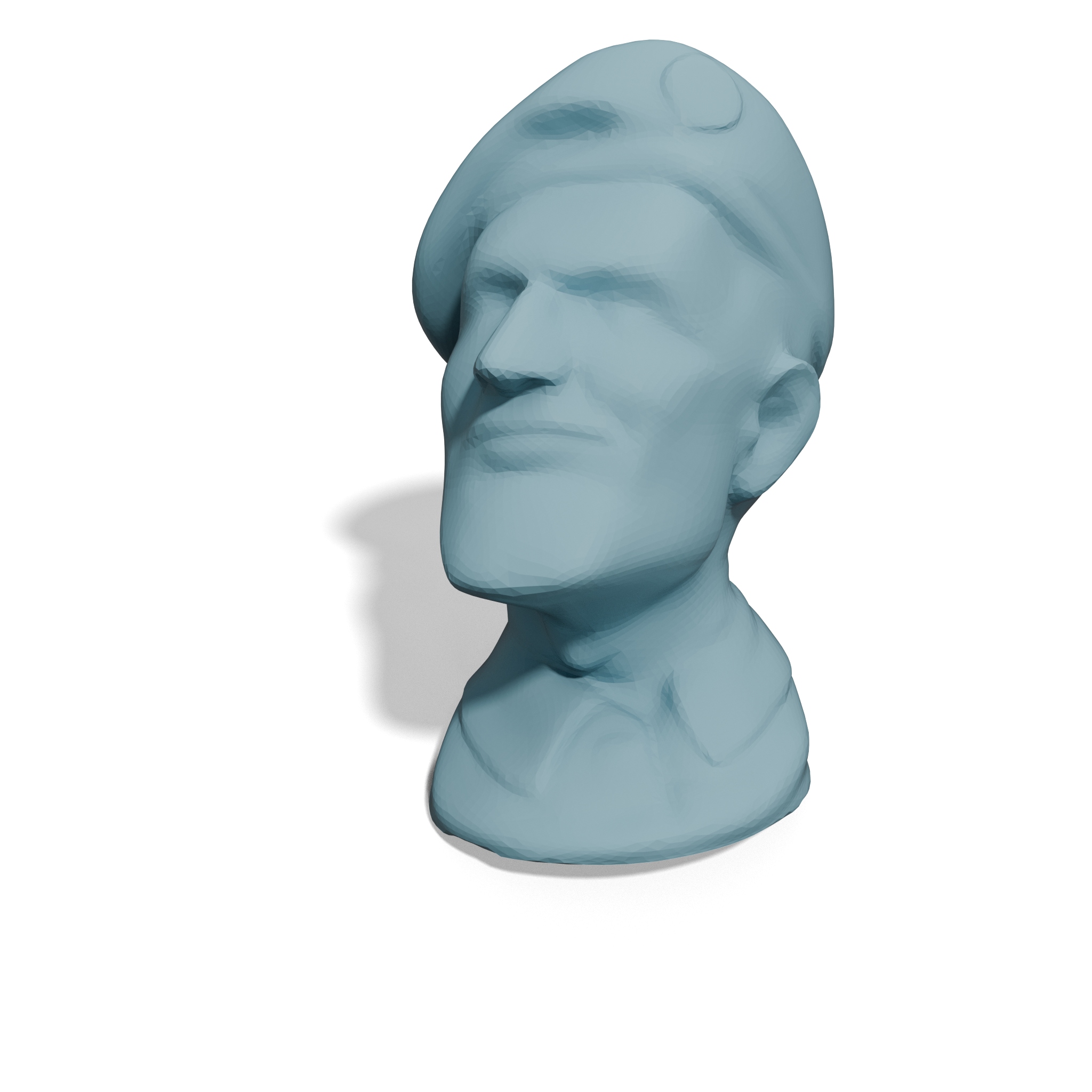} &
      \includegraphics[trim={350 0 100 0},clip,width=\resLen]{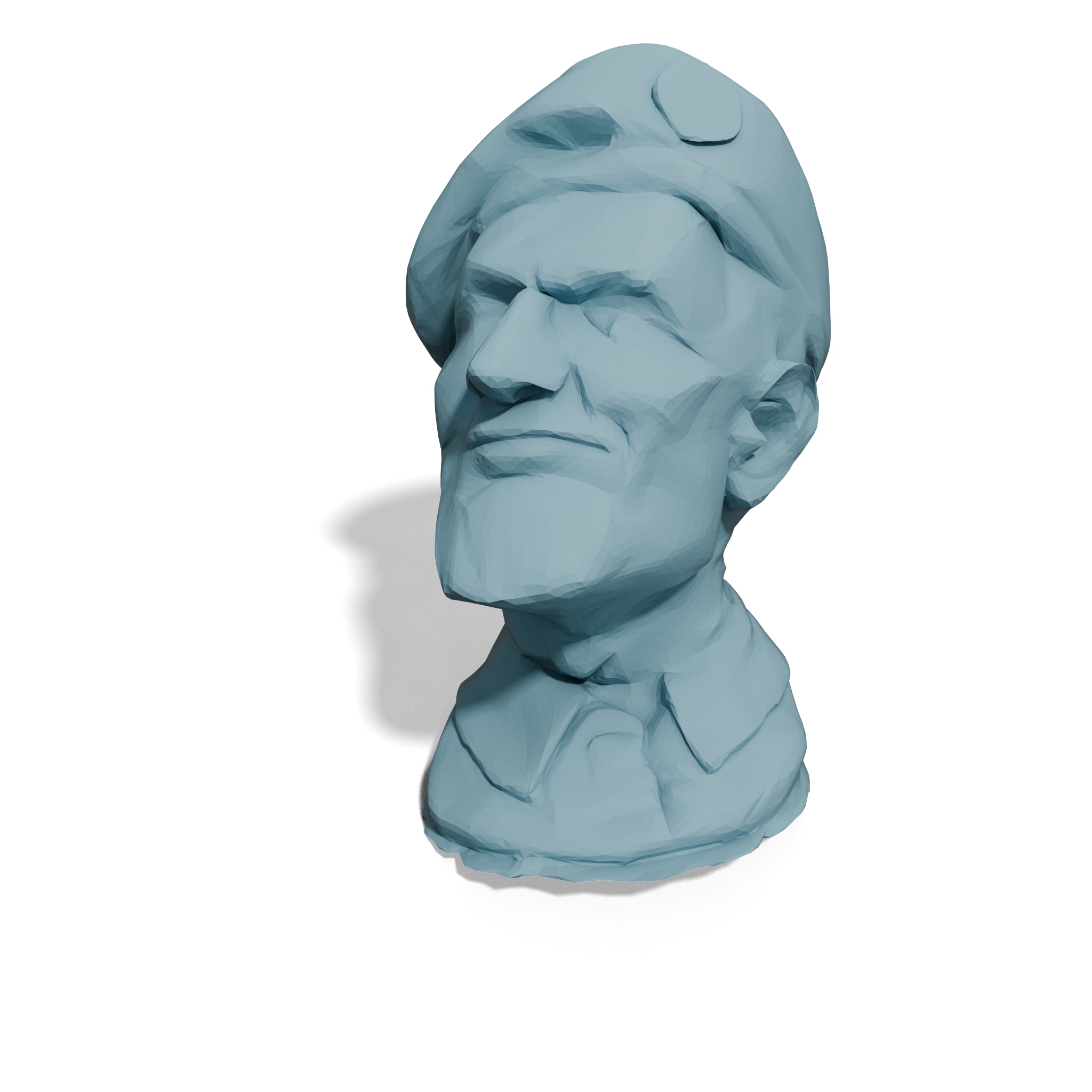} &
      \includegraphics[trim={350 0 100 0},clip,width=\resLen]{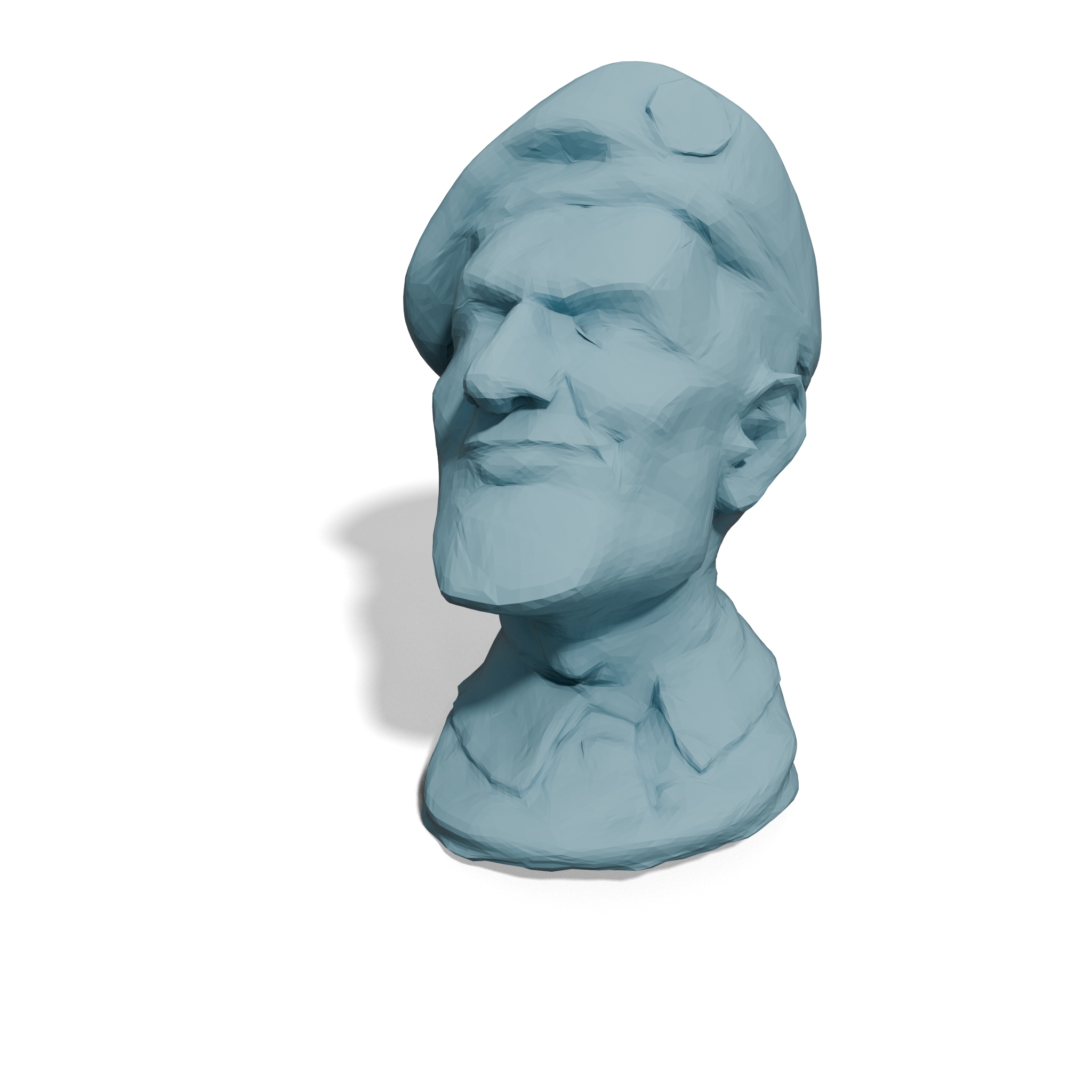} &
      \includegraphics[trim={350 0 100 0},clip,width=\resLen]{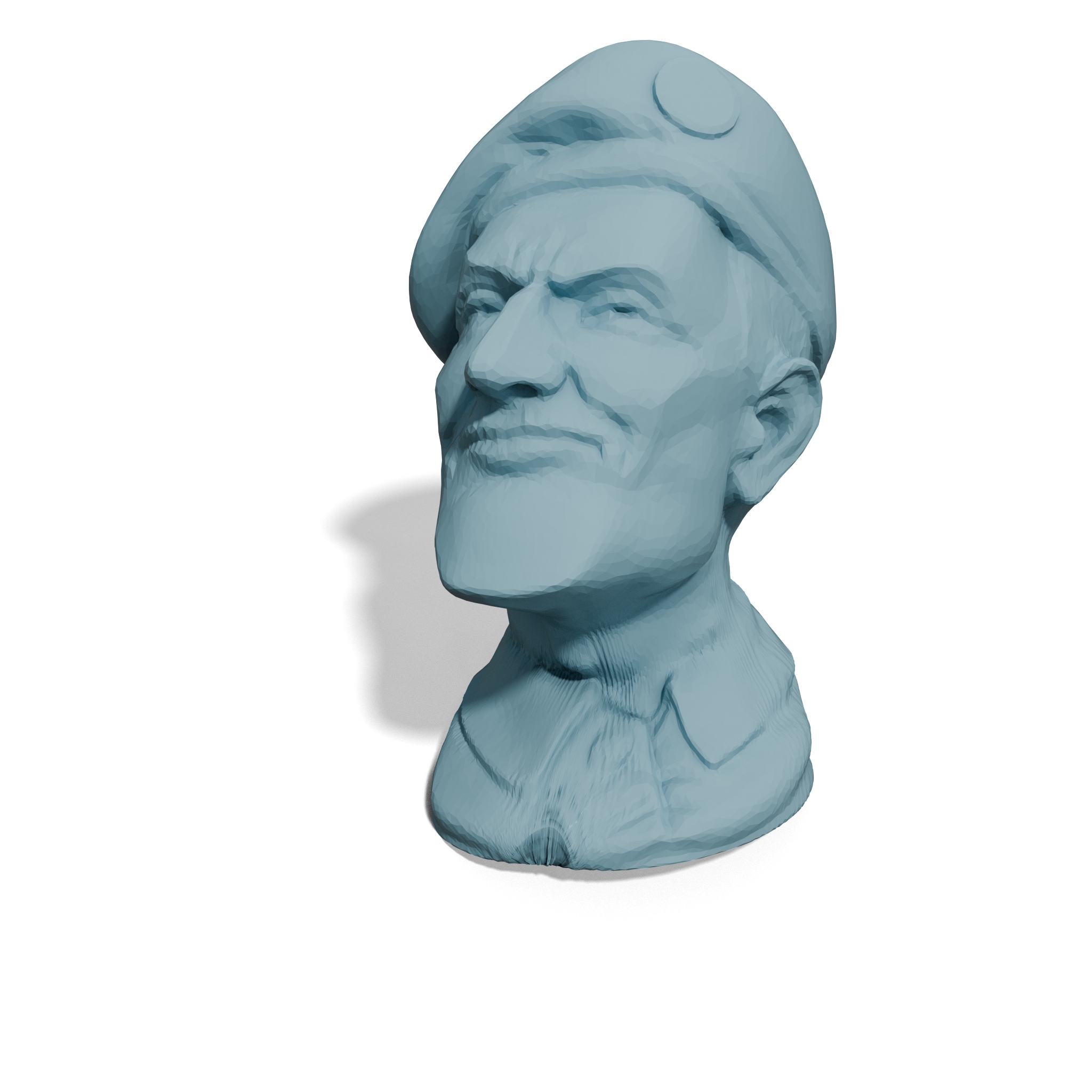}\\[-6pt]
      
      \small CD ($\times10^{-4}$) / HD ($\times10^{-2}$) &
      \small 1.4598 / 2.8164 &
      \small 1.6018 / 3.0431 &
      \small 1.7986 / 7.1388 &
      \small 4.9518 / 5.3569 &
      \small 1.5816 / 4.4674 \\[4pt]

      Ground Truth & QSlim~\cite{QEM1997} & Loop~\cite{LOOP1987Loop} & Butterfly~\cite{ZORIN1996Butterfly} & Neural Subdivision~\cite{LIU2020NeuralSubdivision} & Ours
  \end{tabular}

\caption{Quantitative comparison with baselines on 3D surface meshes randomly sampled from the Thingi10k dataset~\protect\cite{Thingi10K}. 
Our method enforces strict area preservation via OT’s measure-preserving property, enabling neural processing of geometry images 
without decoders and supporting continuous LoD representations.}
  \label{fig:subdivision_comparison}
  \vspace{-5mm}
\end{figure*}

\begin{table*}[htbh]
    \centering

    \caption{Quantitative comparison with state-of-the-art non-mesh representation approaches on 3D surface meshes from Thingi10k~\protect\cite{Thingi10K}. 
    All results are reported at CR = 64. 
    We report Chamfer Distance (CD, $\times 10^{-4}$) and Hausdorff Distance (HD, $\times 10^{-2}$), and indicate whether each method supports single-pass reconstruction, requires successive decoding for LoD, or enables decoder-free LoD.}

    \begin{tabular}{|c|c|c|c|c|c|c|}
        \hline
        \textbf{Method} & 
        \textbf{Shape Representation} &
        \textbf{CD $\downarrow$} & 
        \textbf{HD $\downarrow$} &
        \textbf{Single-Pass} &
        \textbf{Multi-Pass} &
        \textbf{Decoder-Free LoD} \\
        \hline
        ACORN~\protect\cite{martel2021acorn}   & Neural Octree SDF              & 1.300 & 4.693 & \xmark & \cmark & \xmark \\
        NGLOD~\protect\cite{takikawa2021nglod} & Multi-Resolution Feature Grid  & 0.510 & 2.499 & \xmark & \cmark & \xmark \\
        NCS~\protect\cite{Morreale2022NCS}     & Convolutional Surface Features & 0.083 & 1.538 & \xmark & \cmark & \xmark \\
        \hline
        \textbf{Ours}                          & \textbf{Neural Geometry Images} & \textbf{0.059} & \textbf{1.482} & \cmark & \xmark & \cmark \\
        \hline
    \end{tabular}
    \label{tab:representation_metrics}
    
\end{table*}

\begin{table*}[htbh]
    \centering
    \caption{Quantitative comparison with mesh-based subdivision approaches on 3D surface meshes from Thingi10k~\protect\cite{Thingi10K}. 
    All results are reported at CR = 64. 
    We report Chamfer Distance (CD, $\times 10^{-4}$) and Hausdorff Distance (HD, $\times 10^{-2}$), and indicate whether each method supports single-pass reconstruction, requires successive decoding for LoD, or enables decoder-free LoD. 
    Traditional methods (QSlim, Loop, Butterfly) rely on progressive refinement but lack neural processing.}
    \setlength{\tabcolsep}{5pt} % overall spacing
    \begin{tabular}{|c|c|c|c|c|c|}
        \hline
        \textbf{Method} & 
        \textbf{Chamfer Dist (CD) $\downarrow$} & 
        \textbf{Hausdorff Dist (HD) $\downarrow$} &
        \begin{tabular}[c]{@{}c@{}}\textbf{Single}\\\textbf{Pass}\end{tabular} &
        \begin{tabular}[c]{@{}c@{}}\textbf{Multiple}\\\textbf{Passes}\end{tabular} &
        \begin{tabular}[c]{@{}c@{}}\textbf{Decoder-}\\\textbf{Free LoD}\end{tabular} \\
        \hline
        QSlim~\protect\cite{QEM1997}            & 2.9340 & 5.0445 & \xmark & \cmark & \cmark \\
        Loop~\protect\cite{LOOP1987Loop}        & 3.6250 & 5.4401 & \xmark & \cmark & \cmark \\
        Butterfly~\protect\cite{ZORIN1996Butterfly} & 4.3275 & 7.0944 & \xmark & \cmark & \cmark \\
        Neural Subdivision~\protect\cite{LIU2020NeuralSubdivision} & 2.7152 & 4.8234 & \xmark & \cmark & \xmark \\
        \hline
        \textbf{Ours (Neural Geometry Images)} & \textbf{2.4637} & \textbf{4.5148} & \cmark & \xmark & \cmark \\
        \hline
    \end{tabular}
    \label{tab:neural_subdivision_metrics}
\end{table*}

\begin{figure}[htb]
  \centering
  \setlength{\resLen}{0.3\linewidth} % 3 images per row

  % --- First row (3 images) ---
  \begin{tabular}{@{}ccc@{}}
    \includegraphics[width=\resLen]{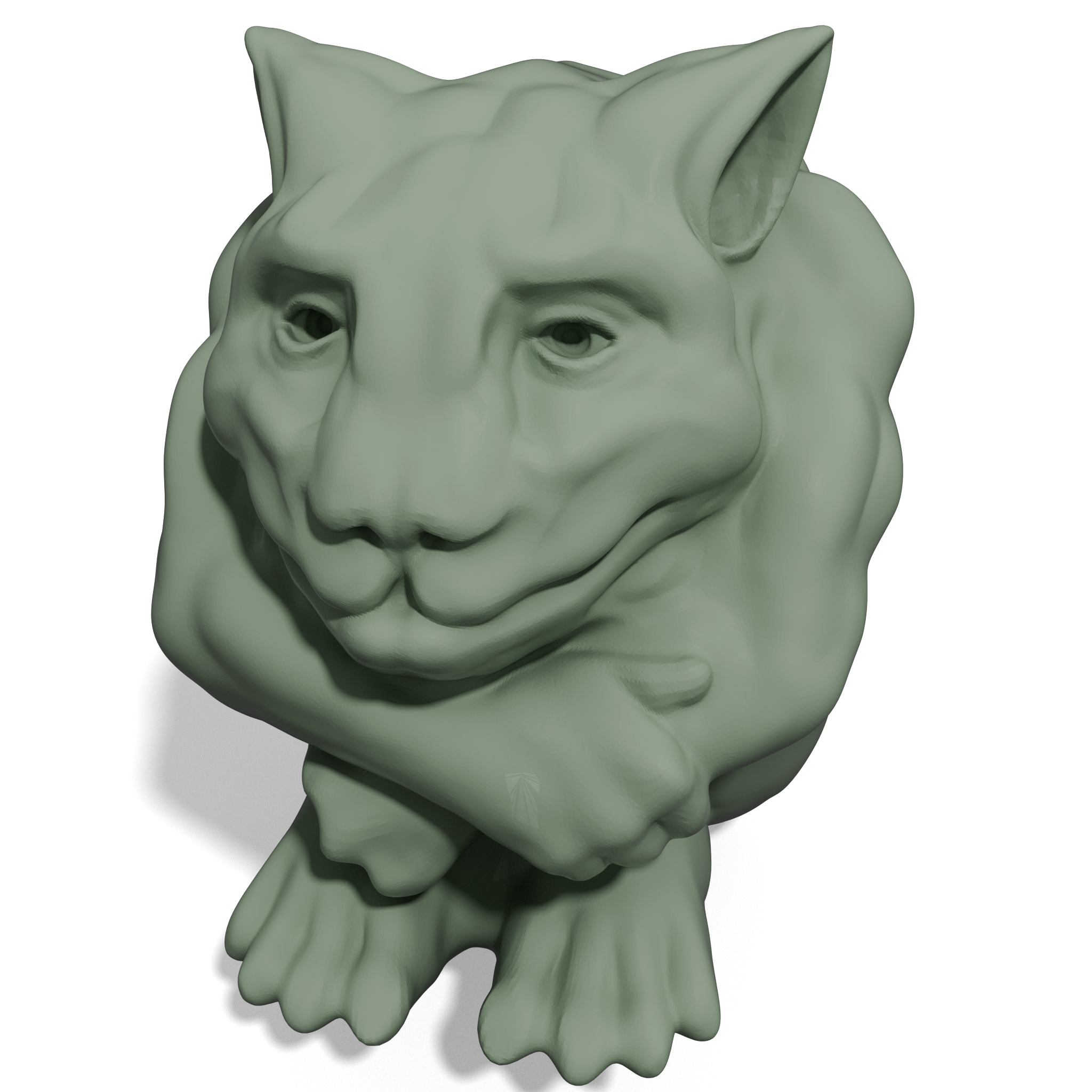} &
    \includegraphics[width=\resLen]{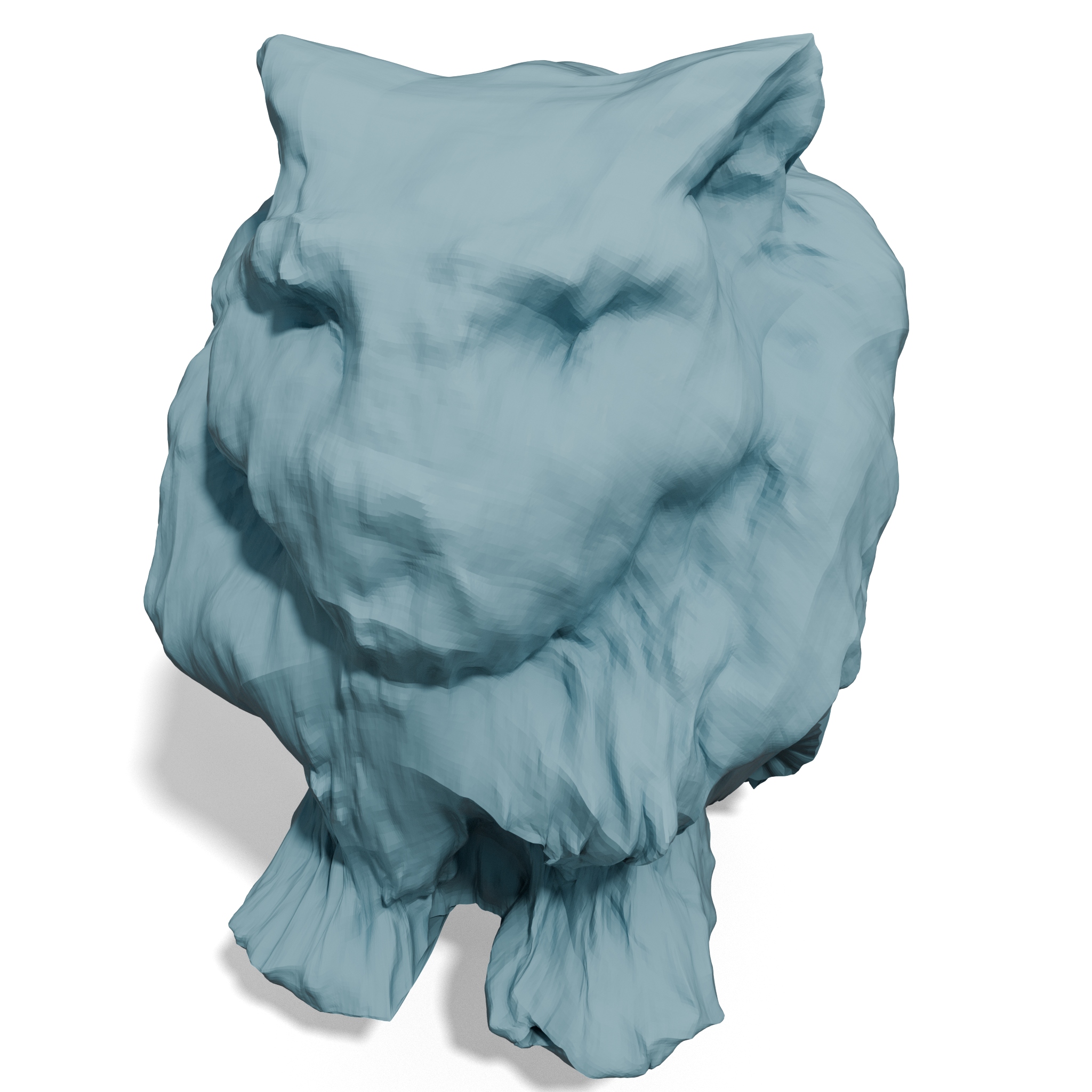} &
    \includegraphics[width=\resLen]{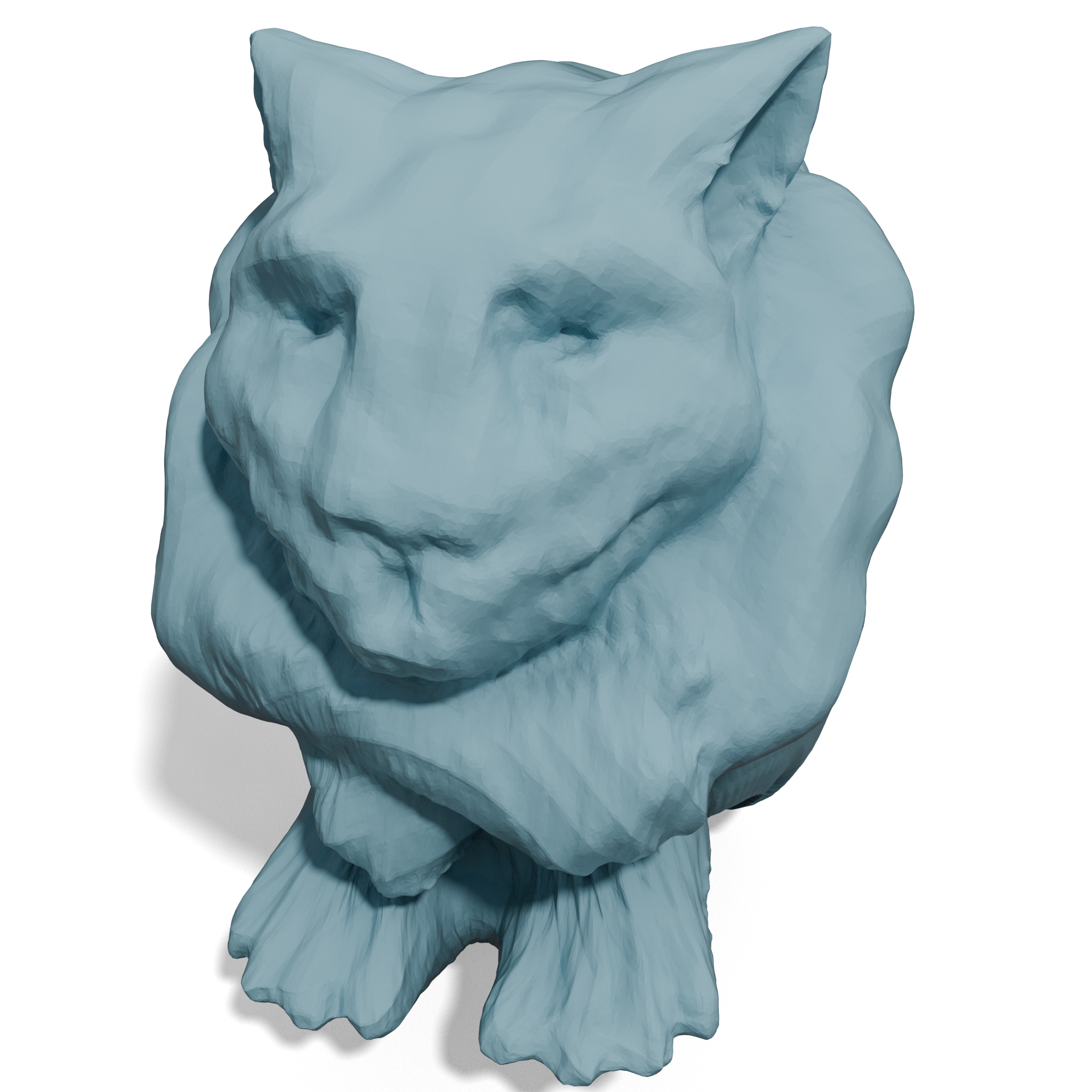} \\
    
    \tiny CR / CD ($\times10^{-4}$) / HD ($\times10^{-2}$) &
    \small 1024 / 44.98 / 23.83 &
    \small 256 / 19.05 / 17.77 \\[1pt]

    Ground Truth &
    Ours ($32 \times 32$) &
    Ours ($64 \times 64$) \\
  \end{tabular}

  \vspace{6pt} % spacing between the two rows

  % --- Second row (3 images) ---
  \begin{tabular}{@{}ccc@{}}
    \includegraphics[width=\resLen]{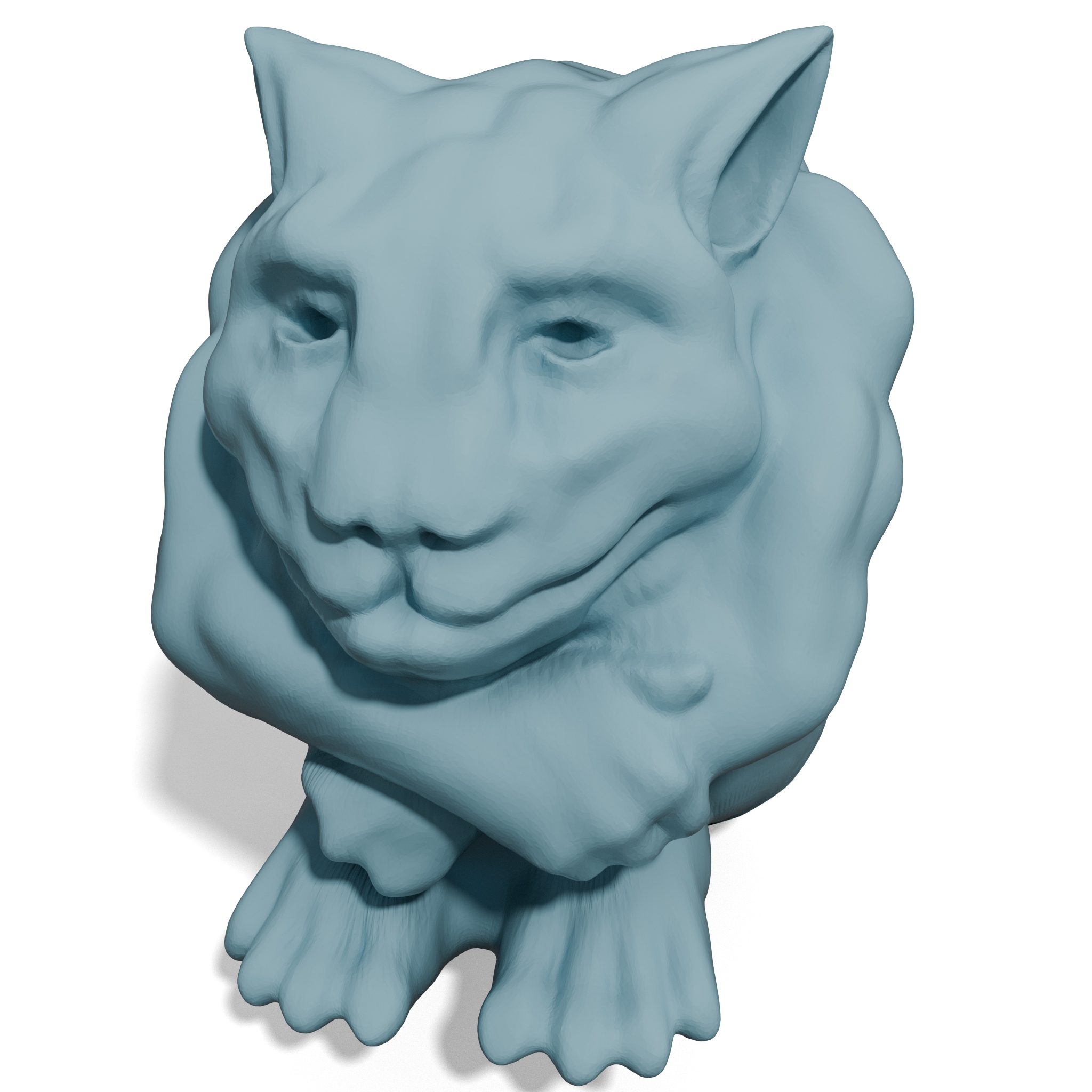} &
    \includegraphics[width=\resLen]{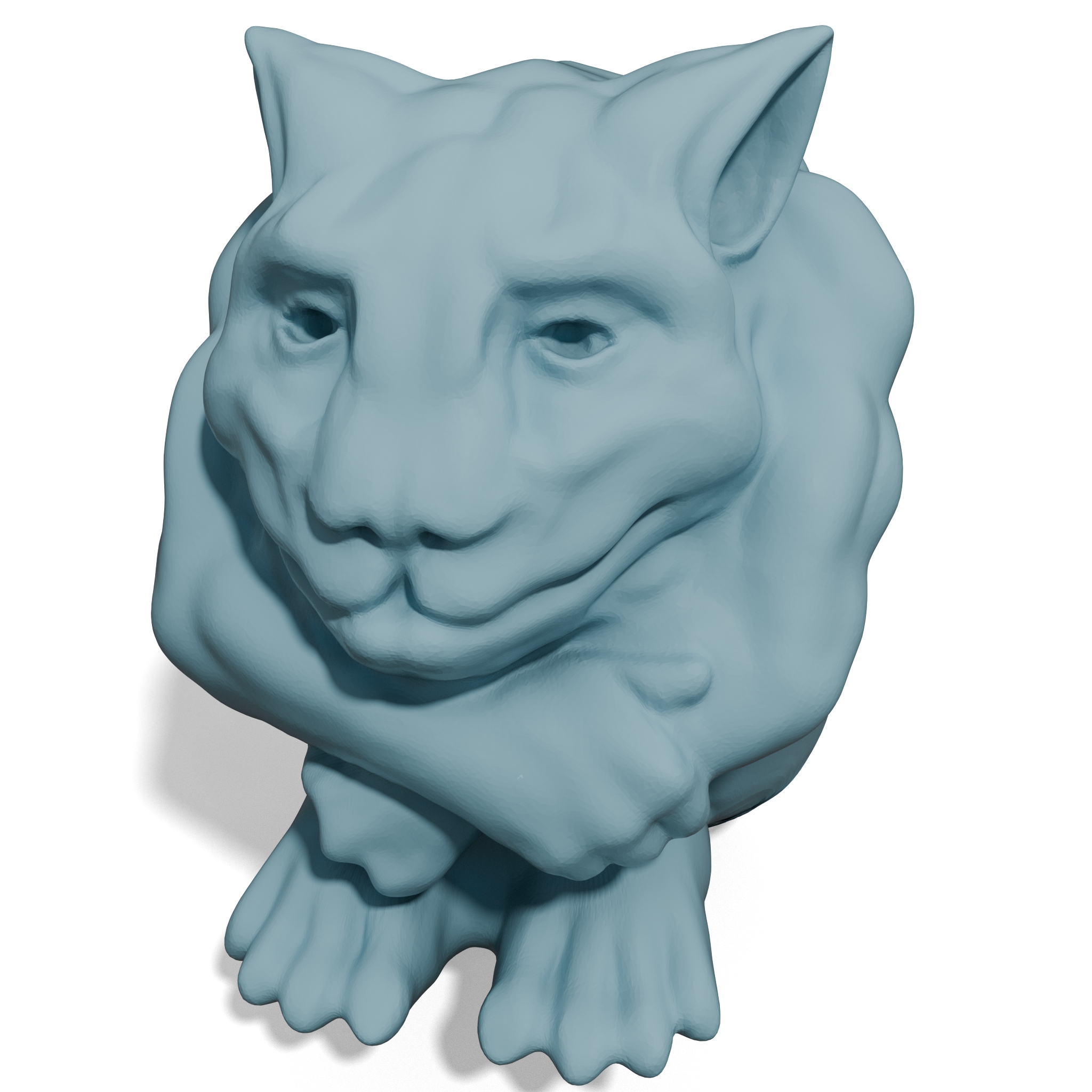} &
    \includegraphics[width=\resLen]{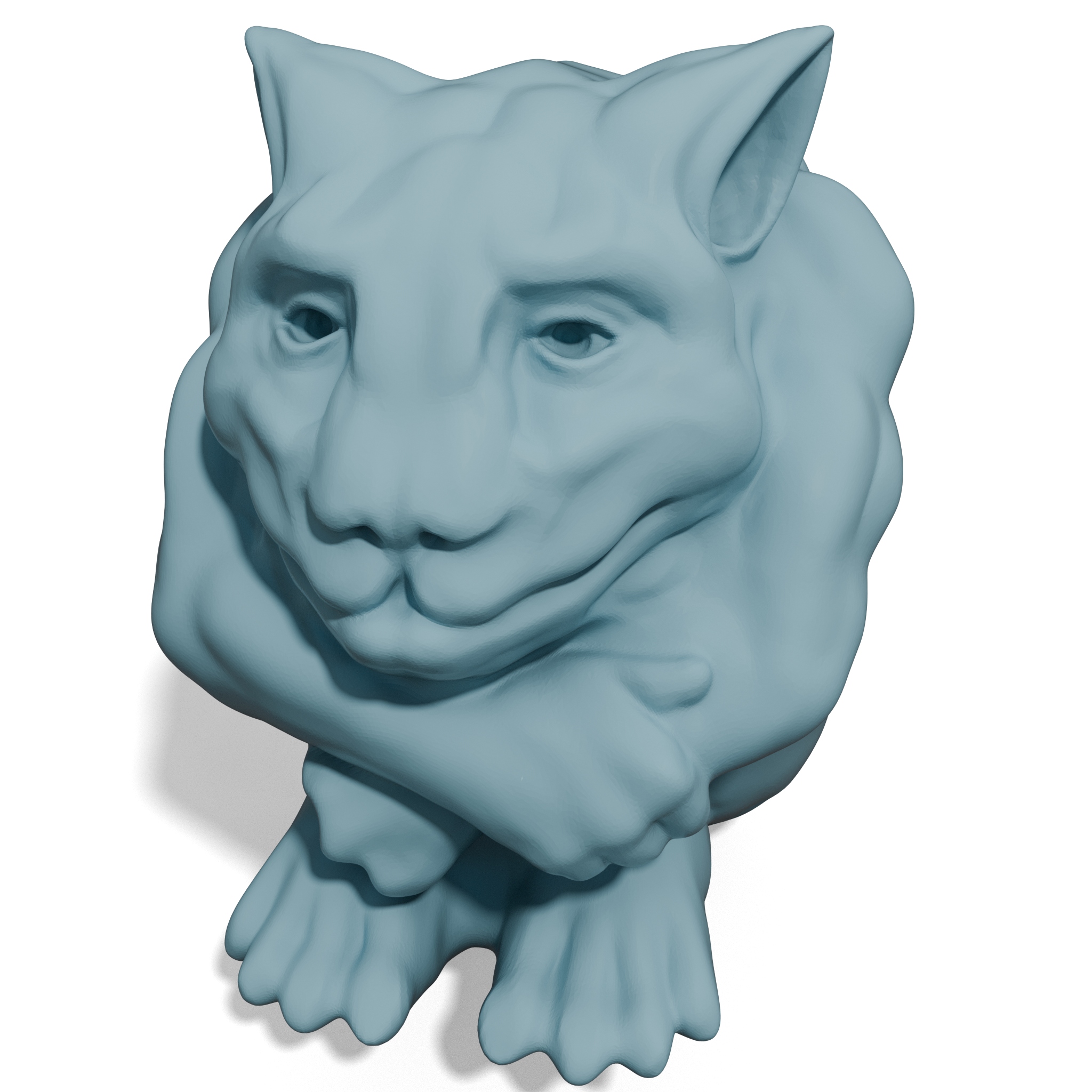} \\

    \small 64 / 3.62 / 13.72 &
    \small 16 / 1.71 / 12.79 &
    \small 4 / 0.52 / 9.17 \\[1pt]

    Ours ($128 \times 128$) &
    Ours ($256 \times 256$) &
    Ours ($512 \times 512$) \\
  \end{tabular}

  \caption{\textbf{Ablation study across stored geometry image mipmap resolutions.}
  Lower resolutions give higher compression but larger errors, while CR = 64 ($128 \times 128$) is the sweet spot between storage and accuracy.}
  \label{fig:varying_resolution_goyle}
\end{figure}

\vspace{-3mm}
\section{Experiments and Results}
We evaluate our method on OMT-based geometry images constructed from diverse objects in the Thingi10K dataset~\cite{Thingi10K}, and compare against baseline parameterization and subdivision methods, as well as recent non-mesh representation approaches, all evaluated on the same dataset.

\subsection{Evaluation Metrics}  
We report Chamfer Distance (CD) and Hausdorff Distance (HD) to measure reconstruction error, and the geometry image compression ratio (CR) to assess storage efficiency. CR is defined as the ratio of the stored image size to that of a $1024 \times 1024$ geometry image, with higher values indicating greater compression and storage savings.

\subsection{Implementation Details}  
The geometry image super-resolution network takes as input a pair of low-resolution mipmapped geometry images: a position image cross-attended with a normal image. Each input is processed by separate convolutional branches, with cross-attention fusing normal features into the position stream. The fused features are refined by residual blocks~\cite{Kaiming2006ResNet} and progressively upsampled via pixel shuffle layers~\cite{Shi2016PixelShuffle}, followed by a final convolution to produce the high-resolution output. To retain low-frequency information, we add a skip connection from the bilinearly upsampled input. Training uses the ADAM optimizer~\cite{kingma2017ADAM} in PyTorch~\cite{paszke2019PyTorch} with a fixed learning rate of $5 \times 10^{-5}$, and data augmentation with $90^\circ$ rotations and flips.

\begin{table}[htbh]
  \centering
  \renewcommand{\arraystretch}{1.2} % Taller rows
  {\normalsize
    \caption{Ablation study across stored geometry image mipmap resolutions. 
    Results show that lower resolutions yield higher compression but larger errors, while CR = 64 ($128 \times 128$) is the sweet spot between storage efficiency and reconstruction accuracy.}
    \label{tab:lod_ablation}
      \vspace{-3mm}
    \begin{tabular}{|c|c|c|c|}
      \hline
      \textbf{Stored Mipmap Resolution} & \textbf{CR} & \textbf{CD $\downarrow$} & \textbf{HD $\downarrow$} \\ 
      \hline
      $32 \times 32$   & 1024 & 0.264 & 1.572 \\ 
      $64 \times 64$   & 256  & 0.074 & 1.393 \\ 
      $128 \times 128$ & 64   & 0.009 & 1.085 \\ 
      $256 \times 256$ & 16   & 0.004 & 0.893 \\ 
      $512 \times 512$ & 4    & \textbf{0.002} & \textbf{0.628} \\
      \hline
    \end{tabular}
  }
\end{table}

\subsection{Geometry Image Sampling}  
We compare our geometry image construction by applying OT with four parameterization methods: uniform~\cite{Tutte1963HowTD}, harmonic~\cite{Eck1995Harmonic}, conformal (Ricci Flow)~\cite{4483509}, and authalic~\cite{Desbrun2002Intrinsic}. Sampling quality is measured by Chamfer Distance (CD) and Hausdorff Distance (HD). As shown in \Cref{fig:parameterization_compare_metrics}, applying OT yields geometry images that qualitatively preserve surface details more faithfully. Quantitative results in \Cref{tab:parameterization_metrics} confirm that OT achieves strictly area-preserving sampling, producing balanced geometry images and consistently lower reconstruction errors. In contrast, traditional parameterizations often suffer from imbalanced sampling, leading to oversampling in low-curvature regions and undersampling in high-curvature regions, which degrades reconstruction quality. By enforcing strict area preservation while remaining compatible with neural processing, applying OT avoids this imbalance and enables continuous LoD without multiple decoders.

\subsection{Neural Overfitting Mesh Compression}  
We compare against recent non-mesh–based neural representations that compress geometry by overfitting implicit functions or feature grids to high-resolution meshes, including NCS~\cite{Morreale2022NCS}, ACORN~\cite{martel2021acorn}, and NGLOD~\cite{takikawa2021nglod}. These methods encode geometry indirectly as network weights or grids and require multiple successive decoder applications to realize levels of detail (LoD), which limits efficiency and scalability. In contrast, our \textbf{neural geometry image-based representation with OT} stores geometry directly as strictly area-preserving images and reconstructs the full-resolution surface in a \textbf{single pass}, eliminating the need for progressive decoding. As shown in \Cref{fig:neural_overfitting} and \Cref{tab:representation_metrics}, at a compression ratio of CR = 64 our representation achieves superior Chamfer Distance (CD) and Hausdorff Distance (HD) while supporting decoder-free continuous LoD, demonstrating the effectiveness of image-based representations over non-mesh neural overfitting approaches.

\subsection{Neural Subdivision Mesh Compression}  
Subdivision-based methods compress geometry by starting from a coarse mesh and progressively refining it to recover a high-resolution surface. Classical approaches such as QSlim~\cite{Garland1997QEM}, Loop~\cite{LOOP1987Loop}, and Butterfly~\cite{ZORIN1996Butterfly} apply hand-crafted refinement rules, while Neural Subdivision~\cite{LIU2020NeuralSubdivision} uses learned weights to perform refinement and reduce memory cost. However, all of these techniques rely on mesh-based representations, which are storage-inefficient and require successive decoding passes to realize levels of detail (LoD). In contrast, our \textbf{neural geometry image-based representation with OT} enforces strictly area-preserving sampling and reconstructs the full-resolution surface in a \textbf{single pass}, eliminating the need for progressive refinement. As shown in \Cref{tab:neural_subdivision_metrics}, at CR = 64 our representation achieves superior Chamfer Distance (CD) and Hausdorff Distance (HD) while supporting decoder-free continuous LoD.

\subsection{Ablation Study}  
We conduct an ablation study to assess the effectiveness of our neural geometry image–based representation with OT under different compression ratios. As summarized in \Cref{tab:lod_ablation}, lower-resolution mipmaps yield higher compression but larger reconstruction errors, while higher resolutions achieve near-lossless recovery. At CR = 64 ($128 \times 128$), our representation provides an optimal trade-off, delivering low Chamfer Distance (CD) and Hausdorff Distance (HD) while preserving strict area balance and enabling decoder-free reconstruction. These results confirm the scalability and efficiency of our approach across varying storage budgets.

\subsection{Limitations and Failure Cases}  
Our proposed neural geometry image–based representation with OT may produce seams along mesh boundaries, which can be reduced by post-processing at additional cost.

\section{Conclusions}
We presented a neural geometry image-based representation for compact storage and accurate reconstruction of 3D surface meshes. Conformal initialization combined with optimal transport refinement yields strictly area-preserving geometry images with balanced sampling. These low-resolution images are restored to full resolution in a single CNN forward pass, eliminating the need for complex decoders and naturally supporting continuous levels of detail on GPUs. Future work includes extending the framework to dynamic mesh sequences and integrating with physics-informed neural networks for simulation.
\clearpage
%%%%%%%%% REFERENCES
{\small
\bibliographystyle{ieee_fullname}
\bibliography{egbib.bib}

@String(CVPR= {IEEE Conf. Comput. Vis. Pattern Recog.})

@String(ECCV= {Eur. Conf. Comput. Vis.})

@String(BMVC= {Brit. Mach. Vis. Conf.})

@String(TOG= {ACM Trans. Graph.})

@String(CVPR  = {CVPR})

@String(ECCV  = {ECCV})

@String(BMVC  =	{BMVC})

@String(TOG   = {ACM TOG})

@article{Gu2002GeometryImages,
  author = {Gu, Xianfeng and Gortler, Steven J. and Hoppe, Hugues},
  title = {Geometry Images},
  journal = {ACM Trans. Graph.},
  volume = {21},
  number = {3},
  pages = {355--361},
  year = {2002},
  doi = {10.1145/566654.566589}
}

@inproceedings{CHEN2023NeuralCompression,
  author = {Chen, Yun-Chun and Kim, Vladimir and Aigerman, Noam and Jacobson, Alec},
  title = {Neural Progressive Meshes},
  booktitle = {ACM SIGGRAPH Conference Proceedings},
  year = {2023},
  articleno = {84},
  numpages = {9},
  doi = {10.1145/3588432.3591531}
}

@article{LIU2020NeuralSubdivision, 
  author = {Liu, Hsueh-Ti Derek and Kim, Vladimir G. and Chaudhuri, Siddhartha and Aigerman, Noam and Jacobson, Alec},
  title = {Neural Subdivision},
  journal = {ACM Trans. Graph.},
  volume = {39},
  number = {4},
  articleno = {124},
  numpages = {16},
  year = {2020},
  doi = {10.1145/3386569.3392418}
}

@inproceedings{ZORIN1996Butterfly, 
  author = {Zorin, Denis and Schröder, Peter and Sweldens, Wim},
  title = {Interpolating Subdivision for Meshes with Arbitrary Topology},
  booktitle = {Proc. SIGGRAPH},
  year = {1996},
  pages = {189--192},
  doi = {10.1145/237170.237254}
}

@inproceedings{LOOP1987Loop, 
  title={Smooth Subdivision Surfaces Based on Triangles},
  author={Charles T. Loop},
  year={1987},
  url={https://api.semanticscholar.org/CorpusID:116150707}
}

@inproceedings{GU2003GlobalConformalParameterization,
  booktitle = {Eurographics Symposium on Geometry Processing},
  editor = {Leif Kobbelt and Peter Schroeder and Hugues Hoppe},
  title = {Global Conformal Surface Parameterization},
  author = {Gu, Xianfeng and Yau, Shing-Tung},
  year = {2003},
  publisher = {The Eurographics Association},
  ISSN = {1727-8384},
  ISBN = {3-905673-06-1},
  DOI = {10.2312/SGP/SGP03/127-137}
}

@article{takikawa2021nglod,
    title = {Neural Geometric Level of Detail: Real-time Rendering with Implicit {3D} Shapes}, 
    author = {Towaki Takikawa and
              Joey Litalien and 
              Kangxue Yin and 
              Karsten Kreis and 
              Charles Loop and 
              Derek Nowrouzezahrai and 
              Alec Jacobson and 
              Morgan McGuire and 
              Sanja Fidler},
    year = {2021},
    journal = {arXiv preprint arXiv:2101.10994}
}

@ARTICLE{4483509,
  author={Jin, Miao and Kim, Junho and Luo, Feng and Gu, Xianfeng},
  journal={IEEE Transactions on Visualization and Computer Graphics}, 
  title={Discrete Surface Ricci Flow}, 
  year={2008},
  volume={14},
  number={5},
  pages={1030-1043},
  doi={10.1109/TVCG.2008.57}}

@inproceedings{Carr2006Fast,
  author    = {Nathan A. Carr and Jared Hoberock and Keenan Crane and John C. Hart},
  title     = {Fast GPU Ray Tracing of Dynamic Meshes Using Geometry Images},
  booktitle = {Proc. Graphics Interface (GI)},
  year      = {2006},
  pages     = {203--209}
}

@article{brenier1991polar,
  title={Polar factorization and monotone rearrangement of vector-valued functions},
  author={Brenier, Yann},
  journal={Communications on pure and applied mathematics},
  volume={44},
  number={4},
  pages={375--417},
  year={1991},
  publisher={Wiley Online Library}
}

@article{gu2016variational,
  title={Variational principles for Minkowski type problems, discrete optimal transport, and discrete Monge-Ampere equations},
  author={Gu, Xianfeng and Luo, Feng and Sun, Jian and Yau, Shing Tung},
  journal={Asian Journal of Mathematics},
  volume={20},
  number={2},
  pages={383--398},
  year={2016},
  publisher={International Press of Boston, Inc.}
}

@article{Levy2002LSCM,
  author    = {Bruno L\'{e}vy and Sylvain Petitjean and Nicolas Ray and J\'{e}r\^{o}me Maillot},
  title     = {Least Squares Conformal Maps for Automatic Texture Atlas Generation},
  journal   = {ACM Transactions on Graphics (TOG)},
  volume    = {21},
  number    = {3},
  pages     = {362--371},
  year      = {2002},
  month     = {July},
  doi       = {10.1145/566654.566590},
  url       = {https://doi.org/10.1145/566654.566590},
  publisher = {Association for Computing Machinery}
}

@article{Desbrun2002Intrinsic,
journal = {Computer Graphics Forum},
title = {{Intrinsic Parameterizations of Surface Meshes}},
author = {Desbrun, Mathieu and Meyer, Mark and Alliez, Pierre},
year = {2002},
publisher = {Blackwell Publishers, Inc and the Eurographics Association},
ISSN = {1467-8659},
DOI = {10.1111/1467-8659.00580}
}

@article{Tutte1963HowTD,
  title={How to Draw a Graph},
  author={William T. Tutte},
  journal={Proceedings of The London Mathematical Society},
  year={1963},
  volume={13},
  pages={743-767},
  url={https://api.semanticscholar.org/CorpusID:13517317}
}

@inproceedings{Eck1995Harmonic,
  author    = {Matthias Eck and Tony DeRose and Tom Duchamp and Hugues Hoppe and Michael Lounsbery and Werner Stuetzle},
  title     = {Multiresolution Analysis of Arbitrary Meshes},
  booktitle = {Proc. SIGGRAPH},
  year      = {1995},
  pages     = {173--182},
  doi       = {10.1145/218380.218440}
}

@inproceedings{Kraevoy2003Matchmaker,
  author    = {Vladislav Kraevoy and Alla Sheffer and Craig Gotsman},
  title     = {Matchmaker: Constructing Constrained Texture Maps},
  booktitle = {ACM SIGGRAPH 2003 Papers},
  pages     = {326--333},
  year      = {2003},
  doi       = {10.1145/1201775.882271}
}

@INPROCEEDINGS{Kaiming2006ResNet,
  author={He, Kaiming and Zhang, Xiangyu and Ren, Shaoqing and Sun, Jian},
  booktitle={2016 IEEE Conference on Computer Vision and Pattern Recognition (CVPR)}, 
  title={Deep Residual Learning for Image Recognition}, 
  year={2016},
  volume={},
  number={},
  pages={770-778},
  doi={10.1109/CVPR.2016.90}}

@INPROCEEDINGS{Shi2016PixelShuffle,
  author={Shi, Wenzhe and Caballero, Jose and Huszár, Ferenc and Totz, Johannes and Aitken, Andrew P. and Bishop, Rob and Rueckert, Daniel and Wang, Zehan},
  booktitle={2016 IEEE Conference on Computer Vision and Pattern Recognition (CVPR)}, 
  title={Real-Time Single Image and Video Super-Resolution Using an Efficient Sub-Pixel Convolutional Neural Network}, 
  year={2016},
  volume={},
  number={},
  pages={1874-1883},
  doi={10.1109/CVPR.2016.207}}

@INPROCEEDINGS{Wang2003MS_SSIM,
  author       = {Zhou Wang and Eero P. Simoncelli and Alan C. Bovik},
  title        = {Multiscale Structural Similarity for Image Quality Assessment},
  booktitle    = {Proceedings of the 37th Asilomar Conference on Signals, Systems and Computers},
  year         = {2003},
  volume       = {2},
  pages        = {1398--1402},
  doi          = {10.1109/ACSSC.2003.1292216},
  organization = {IEEE}
}

@article{Jacques2020Remeshing,
  author  =    {Jacques-Olivier Lachaud, Pascal Romon, Boris Thibert and David
Coeurjolly},
  title   =    {Interpolated corrected curvature measures for polygonal surfaces},
  journal =    {Computer Graphics Forum (Proceedings of Symposium on Geometry Processing 2020)},
  year    =    {2020},
  volume  =    {39},
  number  =    {5},
}

@article{Thingi10K,
  title={Thingi10K: A Dataset of 10,000 3D-Printing Models},
  author={Zhou, Qingnan and Jacobson, Alec},
  journal={arXiv preprint arXiv:1605.04797},
  year={2016}
}

@misc{kingma2017ADAM,
      title={Adam: A Method for Stochastic Optimization}, 
      author={Diederik P. Kingma and Jimmy Ba},
      year={2017},
      eprint={1412.6980},
      archivePrefix={arXiv},
      primaryClass={cs.LG},
      url={https://arxiv.org/abs/1412.6980}, 
}

@misc{paszke2019PyTorch,
      title={PyTorch: An Imperative Style, High-Performance Deep Learning Library}, 
      author={Adam Paszke and Sam Gross and Francisco Massa and Adam Lerer and James Bradbury and Gregory Chanan and Trevor Killeen and Zeming Lin and Natalia Gimelshein and Luca Antiga and Alban Desmaison and Andreas Köpf and Edward Yang and Zach DeVito and Martin Raison and Alykhan Tejani and Sasank Chilamkurthy and Benoit Steiner and Lu Fang and Junjie Bai and Soumith Chintala},
      year={2019},
      eprint={1912.01703},
      archivePrefix={arXiv},
      primaryClass={cs.LG},
      url={https://arxiv.org/abs/1912.01703}, 
}

@INPROCEEDINGS{Morreale2022NCS,
  author={Morreale, Luca and Aigerman, Noam and Guerrero, Paul and Kim, Vladimir G. and Mitra, Niloy J.},
  booktitle={2022 IEEE/CVF Conference on Computer Vision and Pattern Recognition (CVPR)}, 
  title={Neural Convolutional Surfaces}, 
  year={2022},
  volume={},
  number={},
  pages={19311-19320},
  doi={10.1109/CVPR52688.2022.01873}}

@article{martel2021acorn,
  author    = {Martel, Julien N. P. and Lindell, David B. and Lin, Connor Z. and Chan, Eric R. and Monteiro, Marco and Wetzstein, Gordon},
  title     = {ACORN: Adaptive Coordinate Networks for Neural Scene Representation},
  journal   = {ACM Transactions on Graphics (TOG)},
  volume    = {40},
  number    = {4},
  pages     = {58:1--58:13},
  year      = {2021},
  publisher = {Association for Computing Machinery},
  doi       = {10.1145/3450626.3459785}
}

@inproceedings{Garland1997QEM,
  author    = {Michael Garland and Paul S. Heckbert},
  title     = {Surface Simplification Using Quadric Error Metrics},
  booktitle = {Proc. SIGGRAPH},
  year      = {1997},
  pages     = {209--216},
  doi       = {10.1145/258734.258849}
}

@article{Nicolet2021Large,
  author = {Baptiste Nicolet and Alec Jacobson and Wenzel Jakob},
  title = {Large Steps in Inverse Rendering of Geometry},
  journal = {ACM Transactions on Graphics (TOG)},
  volume = {40},
  number = {6},
  article = {248},
  year = {2021},
  month = {December},
  pages = {13},
  doi = {10.1145/3478513.3480501}
}

@misc{liu2023zero1to3,
      title={Zero-1-to-3: Zero-shot One Image to 3D Object}, 
      author={Ruoshi Liu and Rundi Wu and Basile Van Hoorick and Pavel Tokmakov and Sergey Zakharov and Carl Vondrick},
      year={2023},
      eprint={2303.11328},
      archivePrefix={arXiv},
      primaryClass={cs.CV}
}

@article{liu2023syncdreamer,
  title={SyncDreamer: Generating Multiview-consistent Images from a Single-view Image},
  author={Liu, Yuan and Lin, Cheng and Zeng, Zijiao and Long, Xiaoxiao and Liu, Lingjie and Komura, Taku and Wang, Wenping},
  journal={arXiv preprint arXiv:2309.03453},
  year={2023}
}

@inproceedings{mildenhall2020nerf,
  title={NeRF: Representing Scenes as Neural Radiance Fields for View Synthesis},
  author={Ben Mildenhall and Pratul P. Srinivasan and Matthew Tancik and Jonathan T. Barron and Ravi Ramamoorthi and Ren Ng},
  year={2020},
  booktitle={ECCV},
}

@book{botsch2010polygon,
  author    = {Mario Botsch and Pierre Alliez and Bruno Lévy and Leif Kobbelt and Mark Pauly},
  title     = {Polygon Mesh Processing},
  year      = {2010},
  publisher = {CRC Press},
  doi       = {10.1201/b10688},
}

@article{guo2024tetsphere,
  title={TetSphere Splatting: Representing High-Quality Geometry with Lagrangian Volumetric Meshes},
  author={Guo, Minghao and Wang, Bohan and He, Kaiming and Matusik, Wojciech},
  journal={arXiv preprint arXiv:2405.20283},
  year={2024}
}

@article{long2023wonder3d,
  title={Wonder3D: Single Image to 3D using Cross-Domain Diffusion},
  author={Long, Xiaoxiao and Guo, Yuan-Chen and Lin, Cheng and Liu, Yuan and Dou, Zhiyang and Liu, Lingjie and Ma, Yuexin and Zhang, Song-Hai and Habermann, Marc and Theobalt, Christian and others},
  journal={arXiv preprint arXiv:2310.15008},
  year={2023}
}

@inproceedings{mandikal20183dlmnet,
  title = {{3D-LMNet}: Latent Embedding Matching for Accurate and Diverse 3D Point Cloud Reconstruction from a Single Image},
  author = {Mandikal, Priyanka and Navaneet, K L and Agarwal, Mayank and Babu, R Venkatesh},
  booktitle = {Proceedings of the British Machine Vision Conference ({BMVC})},
  year = {2018}
}

@InProceedings{Groueix_2018_CVPR,
author = {Groueix, Thibault and Fisher, Matthew and Kim, Vladimir G. and Russell, Bryan C. and Aubry, Mathieu},
title = {A Papier-Mâché Approach to Learning 3D Surface Generation},
booktitle = {Proceedings of the IEEE Conference on Computer Vision and Pattern Recognition (CVPR)},
month = {June},
year = {2018}
}

@BOOK{CCG08,
  AUTHOR =       {Xianfeng David Gu and Shing-Tung Yau},
  TITLE =        {Computational Conforaml Geometry},
  PUBLISHER =    {High Education Press and International Press},
  YEAR =         {2008},
  series =       {Advanced Lectures in Mathematics},
}

@inproceedings{Crane:2013:DGP,
                  author = {Keenan Crane and Fernando de Goes and Mathieu Desbrun and Peter Schröder},
                  title = {Digital Geometry Processing with Discrete Exterior Calculus},
                  booktitle = {ACM SIGGRAPH 2013 courses},
                  series = {SIGGRAPH '13},
                  year = {2013},
                  location = {Anaheim, California},
                  numpages = {126},
                  publisher = {ACM},
                  address = {New York, NY, USA},
                  }

@inproceedings{Neus,
 author = {Wang, Peng and Liu, Lingjie and Liu, Yuan and Theobalt, Christian and Komura, Taku and Wang, Wenping},
 booktitle = {Advances in Neural Information Processing Systems},
 editor = {M. Ranzato and A. Beygelzimer and Y. Dauphin and P.S. Liang and J. Wortman Vaughan},
 pages = {27171--27183},
 publisher = {Curran Associates, Inc.},
 title = {NeuS: Learning Neural Implicit Surfaces by Volume Rendering for Multi-view Reconstruction},
 url = {https://proceedings.neurips.cc/paper_files/paper/2021/file/e41e164f7485ec4a28741a2d0ea41c74-Paper.pdf},
 volume = {34},
 year = {2021}}

@article{dreamFusion22,
  author = {Poole, Ben and Jain, Ajay and Barron, Jonathan T. and Mildenhall, Ben},
  title = {DreamFusion: Text-to-3D using 2D Diffusion},
  journal = {arXiv},
  year = {2022},
}

@inproceedings{mehta2022level,
  author = {Mehta, Ishit and Chandraker, Manmohan and Ramamoorthi, Ravi},
  title = {A Level Set Theory for Neural Implicit Evolution Under Explicit Flows},
  booktitle = {Proceedings of the 17th European Conference on Computer Vision (ECCV)},
  year = {2022},
  pages = {711--729},
  publisher = {Springer},
  doi = {10.1007/978-3-031-20086-1_41},
  url = {https://doi.org/10.1007/978-3-031-20086-1_41}
}

@article{nichol2022pointe,
  title={Point-E: A System for Generating 3D Point Clouds from Complex Prompts},
  author={Alex Nichol and Heewoo Jun and Prafulla Dhariwal and Pamela Mishkin and Mark Chen},
  journal={arXiv preprint arXiv:2212.08751},
  year={2022}
}

@article{pointflow,
 title={PointFlow: 3D Point Cloud Generation with Continuous Normalizing Flows},
 author={Yang, Guandao and Huang, Xun and Hao, Zekun and Liu, Ming-Yu and Belongie, Serge and Hariharan, Bharath},
 journal={arXiv},
 year={2019}
}

@inproceedings{Sifakis2012FEM,
  author = {Sifakis, Eftychios and Barbic, Jernej},
  title = {FEM simulation of 3D deformable solids: A practitioner's guide to theory, discretization and model reduction},
  booktitle = {ACM SIGGRAPH 2012 Courses},
  year = {2012},
  pages = {20:1--20:50},
  doi = {10.1145/2343483.2343501}
}

@book{BrennerScott1994,
  author = {Brenner, Susanne C. and Scott, L. Ridgway},
  title = {The Mathematical Theory of Finite Element Methods},
  publisher = {Springer},
  year = {1994},
  series = {Texts in Applied Mathematics},
  volume = {15},
  doi = {10.1007/978-1-4757-3655-3}
}

@inproceedings{Jung2023,
  author = {Yucheol Jung and Hyomin Kim and Gyeongha Hwang and Seung-Hwan Baek and Seungyong Lee},
  title = {Mesh Density Adaptation for Template-based Shape Reconstruction},
  year = {2023},
  doi = {10.1145/3588432.3591498},
  booktitle = {ACM SIGGRAPH 2023 Conference Proceedings},
  publisher = {ACM},
  url = {https://doi.org/10.1145/3588432.3591498},
  location = {Los Angeles, CA, USA}
}

@misc{neuralflow,
title={Neural Mesh Flow: 3D Manifold Mesh Generationvia Diffeomorphic Flows},
author={Kunal Gupta and Manmohan Chandraker},
year={2020},
eprint={2007.10973},
archivePrefix={arXiv},
primaryClass={cs.CV} }

@inproceedings{2dgs24,
    title={2D Gaussian Splatting for Geometrically Accurate Radiance Fields},
    author={Huang, Binbin and Yu, Zehao and Chen, Anpei and Geiger, Andreas and Gao, Shenghua},
    publisher = {Association for Computing Machinery},
    booktitle = {SIGGRAPH 2024 Conference Papers},
    year      = {2024},
    doi       = {10.1145/3641519.3657428}
}

@article{MvDream23,
  author = {Tang, Shitao and Zhang, Fuyang and Chen, Jiacheng and Wang, Peng and Furukawa, Yasutaka},
  title = {MVDiffusion: Enabling Holistic Multi-view Image Generation with Correspondence-Aware Diffusion},
  journal = {arXiv},
  year = {2023},
}

@inproceedings{voxels,
 author = {Wu, Jiajun and Zhang, Chengkai and Xue, Tianfan and Freeman, Bill and Tenenbaum, Josh},
 booktitle = {Advances in Neural Information Processing Systems},
 editor = {D. Lee and M. Sugiyama and U. Luxburg and I. Guyon and R. Garnett},
 pages = {},
 publisher = {Curran Associates, Inc.},
 title = {Learning a Probabilistic Latent Space of Object Shapes via 3D Generative-Adversarial Modeling},
 url = {https://proceedings.neurips.cc/paper_files/paper/2016/file/44f683a84163b3523afe57c2e008bc8c-Paper.pdf},
 volume = {29},
 year = {2016}
}

@inproceedings{VoxNet15,
  title={VoxNet: A 3D Convolutional Neural Network for real-time object recognition},
  author={Maturana, Daniel and Scherer, Sebastian},
  booktitle={Ieee/rsj International Conference on Intelligent Robots and Systems},
  pages={922-928},
  year={2015},
}

@article {ocnn17,
title 		= {{O-CNN: Octree-based Convolutional Neural Networks for 3D Shape Analysis}},
author 		= {Wang, Peng-Shuai and Liu, Yang and Guo, Yu-Xiao and Sun, Chun-Yu and Tong, Xin},
journal 	= {ACM Transactions on Graphics (SIGGRAPH)},
volume 		= {36},
number		= {4},
year 		= {2017},
}

@article{SparseConvNet17,
  title={Submanifold Sparse Convolutional Networks},
  author={Graham, Benjamin and van der Maaten, Laurens},
  journal={arXiv preprint arXiv:1706.01307},
  year={2017}
}

@inproceedings{WangSDF2024,
author = {Wang, Zichen and Deng, Xi and Zhang, Ziyi and Jakob, Wenzel and Marschner, Steve},
title = {A Simple Approach to Differentiable Rendering of SDFs},
booktitle = {SIGGRAPH Asia 2024 Conference Papers},
year = {2024},
publisher = {Association for Computing Machinery},
address = {New York, NY, USA},
doi = {10.1145/3680528.3687573},
url = {https://doi.org/10.1145/3680528.3687573},
articleno = {119},
numpages = {11}
}

@InProceedings{deepSDF19,
author = {Park, Jeong Joon and Florence, Peter and Straub, Julian and Newcombe, Richard and Lovegrove, Steven},
title = {DeepSDF: Learning Continuous Signed Distance Functions for Shape Representation},
booktitle = {Proceedings of the IEEE/CVF Conference on Computer Vision and Pattern Recognition (CVPR)},
month = {June},
year = {2019}
}

@InProceedings{jiang2020sdfdiff,
    author = {Jiang, Yue and Ji, Dantong and Han, Zhizhong and Zwicker, Matthias},
    title = {SDFDiff: Differentiable Rendering of Signed Distance Fields for 3D Shape Optimization},
    booktitle = {The IEEE/CVF Conference on Computer Vision and Pattern Recognition (CVPR)},
    month = {June},
    year = {2020} 
}

@inproceedings{QEM1997,
  author    = {Garland, Michael and Heckbert, Paul S.},
  title     = {Surface simplification using quadric error metrics},
  booktitle = {Proceedings of the 24th Annual Conference on Computer Graphics and Interactive Techniques},
  series    = {SIGGRAPH '97},
  year      = {1997},
  pages     = {209–216},
  publisher = {ACM Press/Addison-Wesley Publishing Co.},
  address   = {USA},
  doi       = {10.1145/258734.258849},
  url       = {https://doi.org/10.1145/258734.258849},
  isbn      = {0897918967},
  numpages  = {8}
}

@article{Gao2025GraphCutUnwrapping,
  author    = {Gao, Xiang and Wang, Xinmu and Zhao, Zhou and Huang, Junqi and Gu, Xianfeng David},
  title     = {Hierarchical GraphCut Phase Unwrapping Based on Invariance of Diffeomorphisms Framework},
  journal   = {IEEE Open Journal of Signal Processing},
  year      = {2025},
  volume    = {6},
  pages     = {546--554},
  doi       = {10.1109/OJSP.2025.3568757}
}

@inproceedings{Wang2025OTTALK,
  author    = {Wang, Xinmu and Gao, Xiang and Song, Xiyun and Yu, Heather and Lin, Zongfang and Peng, Liang and Gu, Xianfeng},
  title     = {OT-Talk: Animating 3D Talking Head with Optimal Transportation},
  booktitle = {Proceedings of the 2025 International Conference on Multimedia Retrieval (ICMR '25)},
  year      = {2025},
  pages     = {1340--1349},
  publisher = {Association for Computing Machinery},
  address   = {New York, NY, USA},
  doi       = {10.1145/3731715.3733411},
  url       = {https://doi.org/10.1145/3731715.3733411},
  isbn      = {9798400718779},
  location  = {Chicago, IL, USA},
  numpages  = {10}
}

@InProceedings{Shi_2025_CVPR,
    author    = {Shi, Jingyu and Luthra, Achleshwar and Li, Jiazhi and Gao, Xiang and Song, Xiyun and Lin, Zongfang and Gu, David and Yu, Heather},
    title     = {OccludeNeRF: Geometry-aware 3D Scene Inpainting with Collaborative Score Distillation in NeRF},
    booktitle = {Proceedings of the Computer Vision and Pattern Recognition Conference (CVPR) Workshops},
    month     = {June},
    year      = {2025},
    pages     = {284-294}
}

@article{chang2015shapenet,
  title={ShapeNet: An Information-Rich 3D Model Repository},
  author={Chang, Angel X. and Funkhouser, Thomas A. and Guibas, Leonidas J. and Hanrahan, Pat and Huang, Qi-Xing and Li, Zimo and Savarese, Silvio and Savva, Manolis and Song, Shuran and Su, Hao and others},
  journal={arXiv preprint arXiv:1512.03012},
  year={2015}
}

@article{objaverse,
  title={Objaverse: A Universe of Annotated 3D Objects},
  author={Matt Deitke and Dustin Schwenk and Jordi Salvador and Luca Weihs and
          Oscar Michel and Eli VanderBilt and Ludwig Schmidt and
          Kiana Ehsani and Aniruddha Kembhavi and Ali Farhadi},
  journal={arXiv preprint arXiv:2212.08051},
  year={2022}
}

@article{objaverseXL,
  title={Objaverse-XL: A Universe of 10M+ 3D Objects},
  author={Matt Deitke and Ruoshi Liu and Matthew Wallingford and Huong Ngo and
          Oscar Michel and Aditya Kusupati and Alan Fan and Christian Laforte and
          Vikram Voleti and Samir Yitzhak Gadre and Eli VanderBilt and
          Aniruddha Kembhavi and Carl Vondrick and Georgia Gkioxari and
          Kiana Ehsani and Ludwig Schmidt and Ali Farhadi},
  journal={arXiv preprint arXiv:2307.05663},
  year={2023}
}
}

\end{document}